%% file: egpaper_for_review.tex
\documentclass[10pt,twocolumn,letterpaper]{article}

\usepackage{iccv}
\usepackage{times}
\usepackage{epsfig}
\usepackage{graphicx}
\usepackage{amsmath}
\usepackage{amssymb}
\usepackage{xcolor}
\usepackage{nicefrac}
\usepackage{setspace}
\usepackage{booktabs}
\usepackage{multirow}
\usepackage{float}
\usepackage{subcaption}
\usepackage{listings}
\usepackage{wrapfig}
\usepackage{algorithm}
\usepackage{enumitem}
\usepackage[export]{adjustbox}
\usepackage[accsupp]{axessibility} %

\usepackage[pagebackref=true,breaklinks=true,letterpaper=true,colorlinks,bookmarks=false]{hyperref}

\iccvfinalcopy %

\def\iccvPaperID{8051} %

\newcommand{\ourmethod}{{FlowE}}
\newcommand{\ourdataset}{{UrbanCity}}

\newif\ifsubmit
\submitfalse

\ifsubmit
\newcommand{\yuwen}[1]{\textcolor{blue}{}}
\newcommand{\mengye}[1]{\textcolor{orange}{}}
\else

\newcommand{\yuwen}[1]{\textcolor{blue}{Yuwen: #1}}
\newcommand{\mengye}[1]{\textcolor{orange}{Mengye: #1}}

\fi

\usepackage{etoolbox}
\makeatletter
\AfterEndEnvironment{algorithm}{\let\@algcomment\relax}
\AtEndEnvironment{algorithm}{\kern2pt\hrule\relax\vskip3pt\@algcomment}
\let\@algcomment\relax
\newcommand\algcomment[1]{\def\@algcomment{\footnotesize#1}}
\renewcommand\fs@ruled{\def\@fs@cfont{\bfseries}\let\@fs@capt\floatc@ruled
  \def\@fs@pre{\hrule height.8pt depth0pt \kern2pt}%
  \def\@fs@post{}%
  \def\@fs@mid{\kern2pt\hrule\kern2pt}%
  \let\@fs@iftopcapt\iftrue}
\makeatother

\ificcvfinal\fi

\begin{document}

\title{Self-Supervised Representation Learning from Flow Equivariance}

\author{Yuwen Xiong\quad\quad Mengye Ren\quad\quad Wenyuan Zeng\quad\quad Raquel Urtasun\\
Waabi\thanks{This work was done by all authors while at Uber ATG},\quad  University of Toronto\\
{\tt\small \{yuwen, mren, wenyuan, urtasun\}@cs.toronto.edu}
}

\maketitle
\ificcvfinal\thispagestyle{empty}\fi

\input{section/abstract.tex}

\input{section/intro.tex}
\input{section/related.tex}
\section{Methodology}

\input{section/background.tex}
\input{section/model.tex}
\input{section/experiment.tex}
\input{section/conclusion.tex}

{\small
  \bibliographystyle{ieee_fullname}
  \bibliography{egbib}
}

\newpage
\section{Supplementary Material}
\input{section/supplementary.tex}

\end{document}

%% file: section/abstract.tex
\begin{abstract}
Self-supervised representation learning is able to learn semantically meaningful features; however,
much of its recent success relies on multiple crops of an image with very few objects. Instead of
learning view-invariant representation from simple images, humans learn representations in a complex
world with changing scenes by observing object movement, deformation, pose variation and ego motion.
Motivated by this ability, we present a new self-supervised learning representation framework that
can be directly deployed on a video stream of complex scenes with many moving objects. Our framework
features a simple flow equivariance objective that encourages the network to predict the features of
another frame by applying a flow transformation to the features of the current frame. Our
representations, learned from high-resolution raw video, can be readily used for downstream tasks on
static images. Readout experiments on challenging semantic segmentation, instance segmentation, and
object detection benchmarks show that we are able to outperform representations obtained from
previous state-of-the-art methods including SimCLR~\cite{chen2020simple} and
BYOL~\cite{grill2020bootstrap}.
\end{abstract}

%% file: section/intro.tex
\section{Introduction}
Rich and informative visual representations epitomize the revolution of deep
learning in computer vision in the past decade. Deep neural nets deliver
surprisingly competitive performance on tasks such as object detection
~\cite{girshick2015fast,ren2015faster,dai2016r} and semantic
segmentation~\cite{chen2017deeplab,zhao2017pyramid}. Until very recently,
visual representations have been learned by large scale supervised learning.
However, for more challenging tasks such as semantic or instance segmentation,
it is much more expensive to obtain labels compared to object classification.
On the other hand, the human brain learns generic visual representations from
raw video of the complex world without much explicit supervision. This is the
direction that we would like to get one step closer towards in this paper.

\begin{figure}
    \centering
    \includegraphics[width=\linewidth]{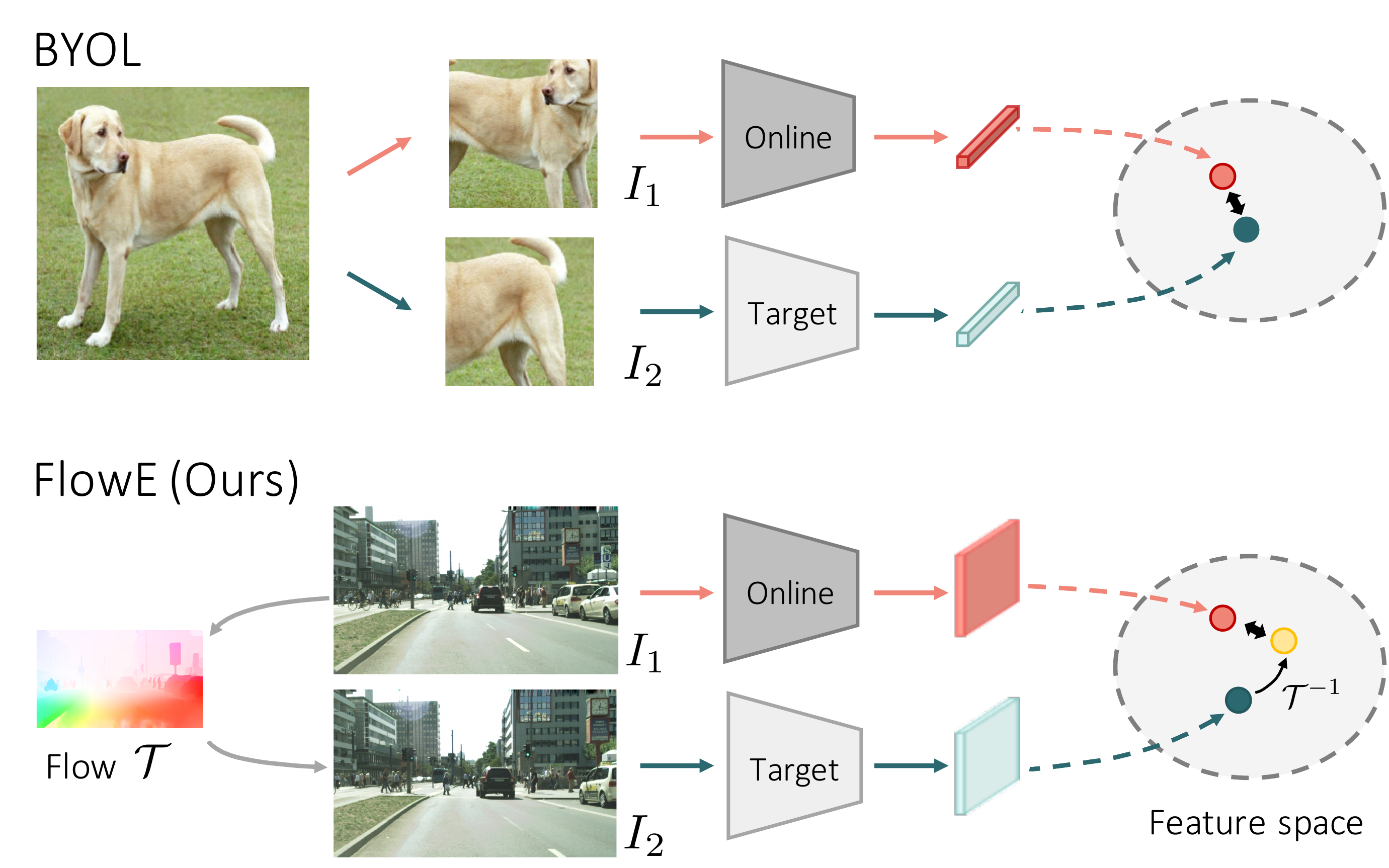}
    \caption{\textbf{Our proposed self-supervised representation learning from Flow
            Equivariance (\ourmethod{}).} Our method is based on
        BYOL~\cite{grill2020bootstrap}, a state of the art method for static image
        representation learning. We encourage the features to obey the same flow
        transformation as the input image pairs.}
    \vspace{-0.15in}
    \label{fig:teaser}
\end{figure}

Recent advances in self-supervised or unsupervised representation learning,
such as SimCLR~\cite{chen2020simple} and BYOL~\cite{grill2020bootstrap}, seem
to point us to a bright path forward: by simply minimizing the feature distance
between two different views of a single image, and performing linear readout at
test time on top, state-of-the-art approaches are now able to match the
classification performance of networks trained with full supervision
end-to-end~\cite{krizhevsky2017imagenet, simonyan2014very, he2016deep}. While
not using any class labels, these methods still rely on the dataset curation
process of carefully selecting clean and object-centered images with a balanced
class distribution. In contrast, videos in the wild feature crowded scenes and
severe data imbalance. As a result, different crops of the same frame can often
lead to either uninteresting regions or erroneous alignment of different
instances in crowded areas. Moreover, none of these methods leverage temporal
information, which contains a rich set of object movement, deformation, and
pose variations. While there has been a large body of literature on learning
representation from video ~\cite{wang2015unsupervised, li2019joint,
    wang2019learning, sermanet2018time, dwibedi2019temporal, dwibedi2020counting,
    wang2011action, wang2013action}, they typically focus on the predicting
correspondence across frames and have not shown better performance on generic
downstream tasks such as semantic and instance segmentation than pretrained
supervised representations from ImageNet~\cite{he2017mask}.

In this paper, we are interested in learning generic representation from raw
high-resolution videos that are directly useful for object detection as well as
semantic and instance segmentation. Whereas prior invariance-based learning
algorithms completely disregard ego-motion and flow transformations across
frames, we argue these are essential elements responsible for the learning of
visual representations in complex scenes~\cite{agrawal2015learning}. Instead of
enforcing multiple crops of the same image (or adjacent frames) to be close in
the feature space, as advocated in prior literature
\cite{he2020momentum,chen2020simple,grill2020bootstrap,gordon2020watching}, we
propose a simple flow equivariance objective that can be applied densely at
every pixel on the feature map, summarized in Figure~\ref{fig:teaser}. In
particular, given two consecutive video frames, we estimate an optical flow map
that denotes a pixel-wise transformation $\mathcal{T}$ between the two frames.
We then train the network to minimize the distance between the the first frame
$\mathbf{h}_1$ and the warped features of the second frame
$\mathcal{T}^{-1}(\mathbf{h}_2)$. Using optical flow ensures that crowded
regions are handled with precise instance alignment. It is also worth noting
that off-the-shelf flow estimators can be trained using either from graphics
simulation~\cite{butler2012naturalistic,dosovitskiy2015flownet,mayer2016large}
or from ego-motion and depth estimation~\cite{ranjan2019competitive}, without
any human labeling effort.

Experiments are carried out on two complex driving video datasets,
BDD100K~\cite{Yu_2020_CVPR} and our in-house dataset UrbanCity, which are
collected from a front camera on a moving car, just like seeing from a mobile
agent in the wild. Our approach, learning from raw videos, can achieve
competitive readout performance on semantic and instance segmentation tasks.
Surprisingly, we are also able to outperform pre-trained representations from
ImageNet~\cite{he2016deep}, likely because of the large domain gap between
ImageNet images and driving videos.

%% file: section/related.tex
\section{Related Work}

In past few years, there has been tremendous progress in learning visual
representations without class label supervision~\cite{goyal2019scaling}.
Typically, networks are trained to predict certain held-out information about
the inputs, such as
context~\cite{doersch2015unsupervised,noroozi2016unsupervised},
rotation~\cite{gidaris2018unsupervised}, colorization~\cite{zhang2016colorful}
and counting~\cite{noroozi2017representation}. Although they have shown to
learn interesting representations, they are still significantly behind
supervised representations on classification tasks.

More recently, contrastive
learning~\cite{oord2018representation,tian2019contrastive} has emerged as a
promising direction for representation learning, closing the gap with
supervised representation on ImageNet. The high level idea is to obtain
different views of the same image using random cropping and other data
augmentations to serve as positive labels, contrasting with other images that
serve as negative labels. MoCo~\cite{he2020momentum} proposed to perform
momentum averaging on the network that encodes negative samples, and
SimCLR~\cite{chen2020simple} proposed to add a non-linear projection head to
make the core representation more general.

Building along this line of work, BYOL~\cite{grill2020bootstrap} removed the
need for negative samples by simply using a slow network with weights getting
slowly updated from the fast network. BYOL proposed to simply minimize the
feature distance between a pair of views of the same image. It is currently one
of the state-of-the-art methods for representation learning on ImageNet.
However, all of the above methods rely on clean static images, which cannot be
easily obtained through raw videos.

Applying contrastive learning on videos seems like a direct extension.
\cite{wang2015unsupervised} proposed to perform unsupervised tracking first to
obtain positive and negative crops of images from different frames in a video
sequence. \cite{sermanet2018time} proposed a multi-view approach that tries to
learn by matching different views from multiple cameras. More recently,
\cite{orhan2020self} treated adjacent frames as positive pairs, whereas
\cite{pirk2020online} preprocessed videos by a class agnostic object detector.
\cite{gordon2020watching} proposed multi-label video contrastive learning.
While using video as input, these methods only considers the invariance
relation between frames and throw away transformations across frames.

This drawback could potentially be complemented by another class of
self-supervised learning algorithms that aim to predict some level of
correspondence or transformation across
frames~\cite{agrawal2015learning,li2019joint}. Cycle consistency is a popular
form of self-supervision that encourages both forward and backward flow on a
sequence of frames to be consistent~\cite{wang2019learning,jabri2020space}.
\cite{dwibedi2019temporal} looked at frame-wise correspondence and encourage
cycle consistency across different videos. Typically these approaches show
competitive performance in terms of video correspondence and label
propagation~\cite{wang2019learning,jabri2020space}, showing a rough
understanding of optical flow. While flow correspondence could be used as
representation for action recognition in the early
literature~\cite{wang2011action,wang2013action}, we would like to decouple the
two tasks between predicting flow correspondence and learning generic visual
representation by providing the flow predictions from off-the-shelf estimators.

\begin{figure*}
\centering
\includegraphics[width=\linewidth]{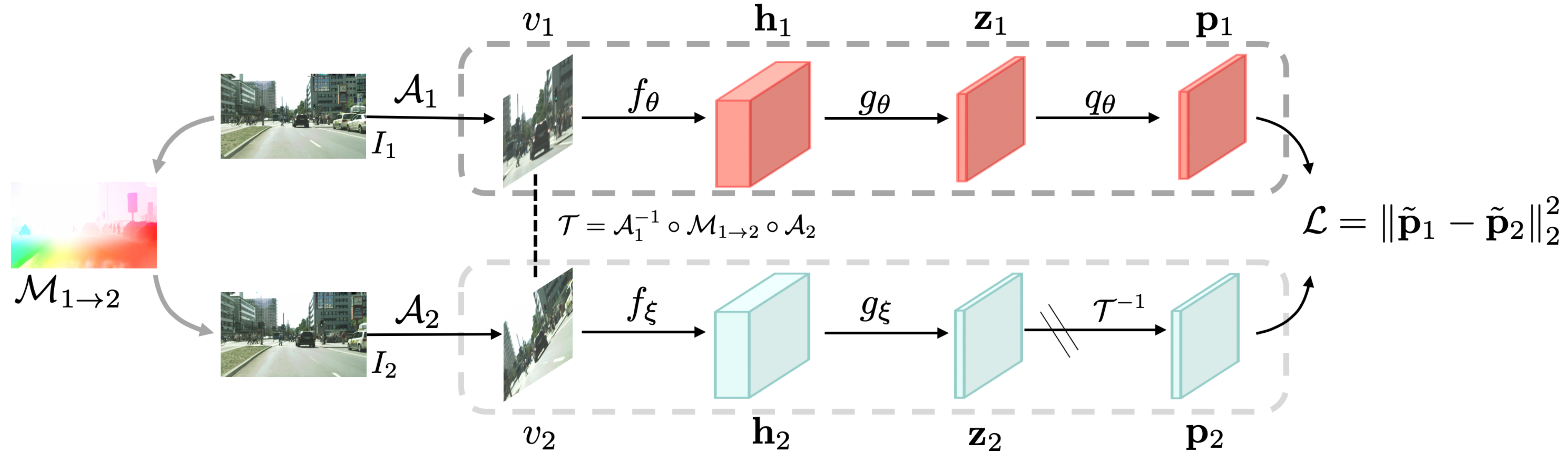}
\caption{
\textbf{The \ourmethod{} learning algorithm.} Given two images of a video $I_1$
and $I_2$, a standalone flow network predicts a dense optical flow field
$\mathcal{M}_{1 \rightarrow 2}$. Two augmented version of images $v = t(I_1)$
and $v' = t'(I_2)$ are fed into the online and target neural network
respectively. The spatial dimension is preserved during the forward pass. The
inverse transformation $\mathcal{T}^{-1}$ is then used to warp the projected
representation $\mathbf{z}_2$ to $\mathbf{p}_2$ to make it align with
$\mathbf{p}_1$.}
\vspace{-0.1in}
\label{fig:model}
\end{figure*}

There has also been a large body of literature on equivariance learning. Just
like how the convolution operator is translational equivariant,
\cite{Nordberg1996EquivarianceAI,cohen2016group,worrall2018cubenet} enforce
strict equivariance over transformation groups. In comparison, we do not
enforce strict equivariance but instead encode it in our training objective to
achieve self-supervision. Our work is most similar to~\cite{denseequi} which
also warps feature maps using optical flow~\cite{zhu2017deep,zhu2017flow}.
Whereas ~\cite{denseequi} tried to directly regress the relative coordinates,
we make use of a simpler distance loss in the feature space. In the end,
\cite{denseequi} produced 3 dimensional image encoding and by contrast we
produce generic high dimensional visual representations.

%% file: section/background.tex
\subsection{Background}
In the background section, we first review BYOL, a state-of-the-art
self-supervised reprensetation learning algorithm, upon which we will build our
\ourmethod{} algorithm on top. We then cover the basics on flow warping.

\vspace{-0.1in}
\paragraph{Bootstrap your own latent (BYOL):}
BYOL performs representation learning by matching two different views of the
same image together. It consists of two neural networks: the online and target
networks. The online network gets updated every iteration and the target
network keeps a momentum averaged copy of the weights. During training, the
online network is going to predict the features produced by the target network,
and the motivation of having a separate target network is to avoid trivial
solution where all images collapse to the same representation.

More specifically, two augmented views $v_1$ and $v_2$ of the same sample are
fed into the encoder $f$ of the online and target network, to get
representation $\mathbf{h}_1,
	\mathbf{h}_2$. To keep the representations general for other readout tasks, BYOL adds a
projector $g$, just like SimCLR~\cite{chen2020simple}. $g$ transforms $\mathbf{h}_1,
	\mathbf{h}_2$ into $\mathbf{z}_1$ and $\mathbf{z}_2$. Finally, the
predictor $q$ takes $\mathbf{z}_1$ and try to produce $\mathbf{p}_1$ that
matches with $\mathbf{z}_2$. Concretely, BYOL minimizes the squared L2 distance
between $\tilde{\mathbf{p}}_1$  and $\tilde{\mathbf{z}}_2$:
\begin{equation}
	\mathcal{L} = \left\lVert \tilde{\mathbf{p}}_1 - \tilde{\mathbf{z}}_2 \right\rVert_2^2,
	\label{eq:byol}
\end{equation}
where $\tilde{\cdot}$ denotes unit-normalization. Note that the target network
is only updated through moving average to avoid trivial solution of collapsed
representation. After the self-supervised training is finished, the projector
and predictor of the online network as well as the target network are discarded
and the encoder of the online network will be preserved for further readout of
downstream tasks such as object classification.

\vspace{-0.1in}
\paragraph{Warping via optical flow:}
Optical flow is widely used in many video processing applications. A flow field
is a two dimensional vector field which defines dense pixel correspondences
between two different video frames. Given a flow field
$\mathcal{M}_{1\rightarrow 2}$, for each pixel on $I_1$ we can find the
corresponding location on $I_2$ and obtain the pixel value via bilinear
interpolation. The warping operation can also be applied to convolutional
feature maps~\cite{zhu2017deep, zhu2017flow}. In our work we use an
off-the-shelf optical flow predictor RAFT~\cite{teed2020raft} because of its
empirical success.

%% file: section/model.tex
\subsection{Learning from Flow Equivariance}
Our method learns dense pixel-level representations based on a flow
equivariance objective, which encourages the features to obey the same flow
transformation as the input image pairs. Our equivariance objective ensures
that a pair of pixels are sampled from the same object across two different
video frames.

Fig.~\ref{fig:model} shows the overview of our framework. Next, we will explain
how it works in detail.

\vspace{-0.1in}
\paragraph{Optical flow \& random affine transformation:}
Given two images $I_1$ and $I_2$ from a video, We use a frozen flow network
$\mathcal{F}$ to predict a dense optical flow field $\mathcal{M}_{1\rightarrow
    2}$ from the two images. We then obtain augmented versions of the two images by
performing random affine transformations $\mathcal{A}_1$ and $\mathcal{A}_2$:
\begin{align} v_1 & =\mathcal{A}_1(I_1) \\ v_2&=\mathcal{A}_2(I_2).
\end{align}
Following~\cite{chen2020simple,grill2020bootstrap}, we further apply random
color distortion and Gaussian blurring on each image. The flow transformation
$\mathcal{T}$ between $v_1$ and $v_2$ is thus defined as the following:
\begin{align}
  \mathcal{T} = \mathcal{A}_1^{-1} \circ \mathcal{M}_{1\rightarrow 2} \circ \mathcal{A}_2.
\end{align}
We then feed the two views into an online network $f_\theta, g_\theta$ and a
momentum updated network $f_\xi, g_\xi$ to obtain the representation
$\mathbf{h}_1,\mathbf{h}_2$, projection $\mathbf{z}_1,\mathbf{z}_2$ and
prediction $\mathbf{p}_1$, as shown in Fig.~\ref{fig:model}.

\vspace{-0.1in}
\paragraph{Equivariance learning:} Our network is fully convolutional and the
spatial dimension of the feature maps are preserved to represent multiple
objects for complex video scenes. We propose to use equivariance as our
training objective. Concretely, we use the inverse flow $\mathcal{T}^{-1}$ to
warp back $\mathbf{z}_2$ to obtain
$\mathbf{p}_2=\mathcal{T}^{-1}(\mathbf{z}_2)$. We use the online network output
$\mathbf{p}_1$ from the predictor to match with $\mathbf{p}_2$. The objective
is simply a squared $\ell_2$ loss averaged over all spatial locations.
\begin{align}
  \mathcal{L} & = \frac{1}{HW} \left\lVert \tilde{\mathbf{p}}_1 -
  \tilde{\mathbf{p}}_2 \right\rVert_2^2,
\end{align}
where $\tilde{\cdot}$ denotes the unit-normalization across the channel
dimension, and $H$ and $W$ denote the spatial resolution of the convolutional
feature map. Similar to the loss in Equation~\ref{eq:byol}, we make sure that
every pixel pair in $\mathbf{p}_1$ and $\mathbf{p}_2$ has a small squared L2
distance.

\begin{figure}
  \centering
  \begin{subfigure}{0.48\linewidth}
    \centering
    \includegraphics[width=\linewidth,keepaspectratio=true]{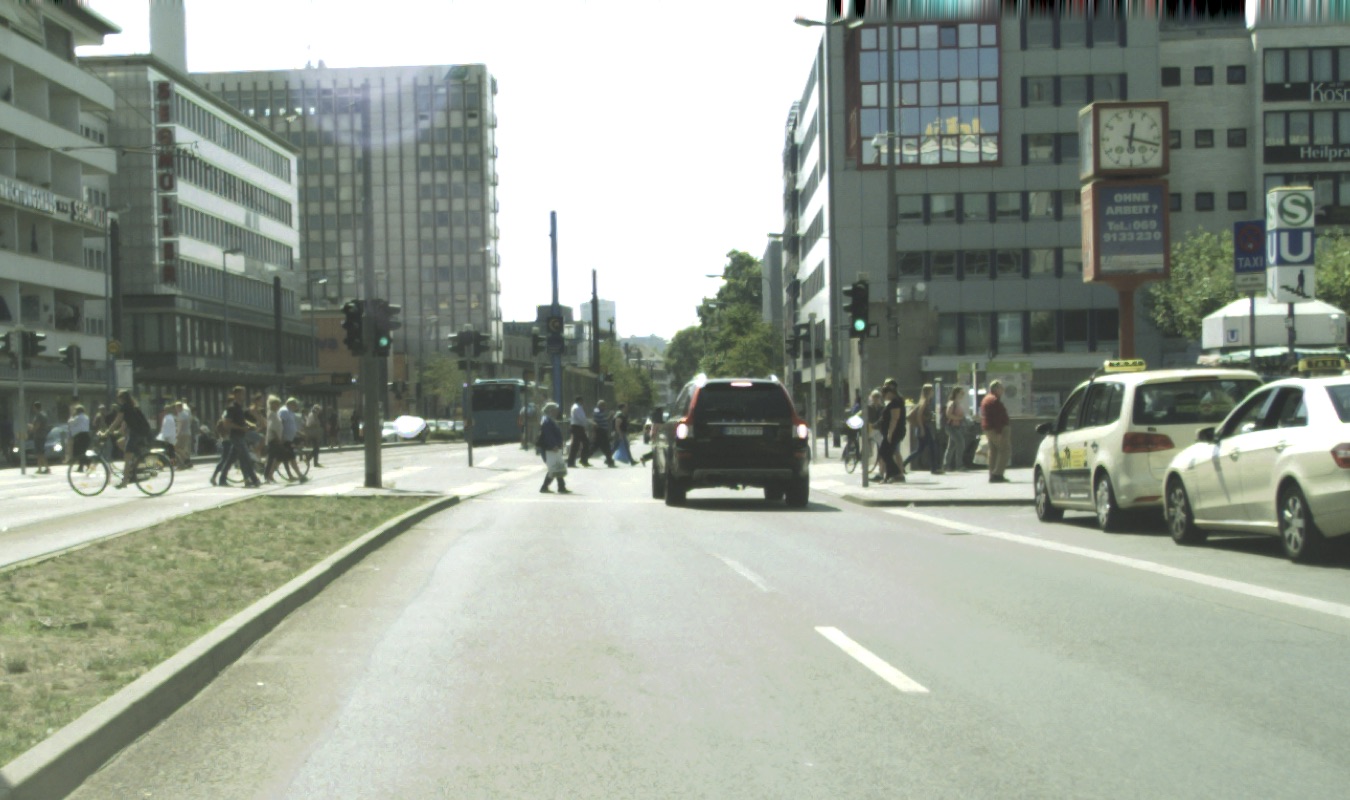}
    \caption{Original}
    \label{fig:intro_illustration_a}
  \end{subfigure}
  \hfill
  \begin{subfigure}{0.48\linewidth}
    \centering
    \includegraphics[width=\linewidth,keepaspectratio=true]{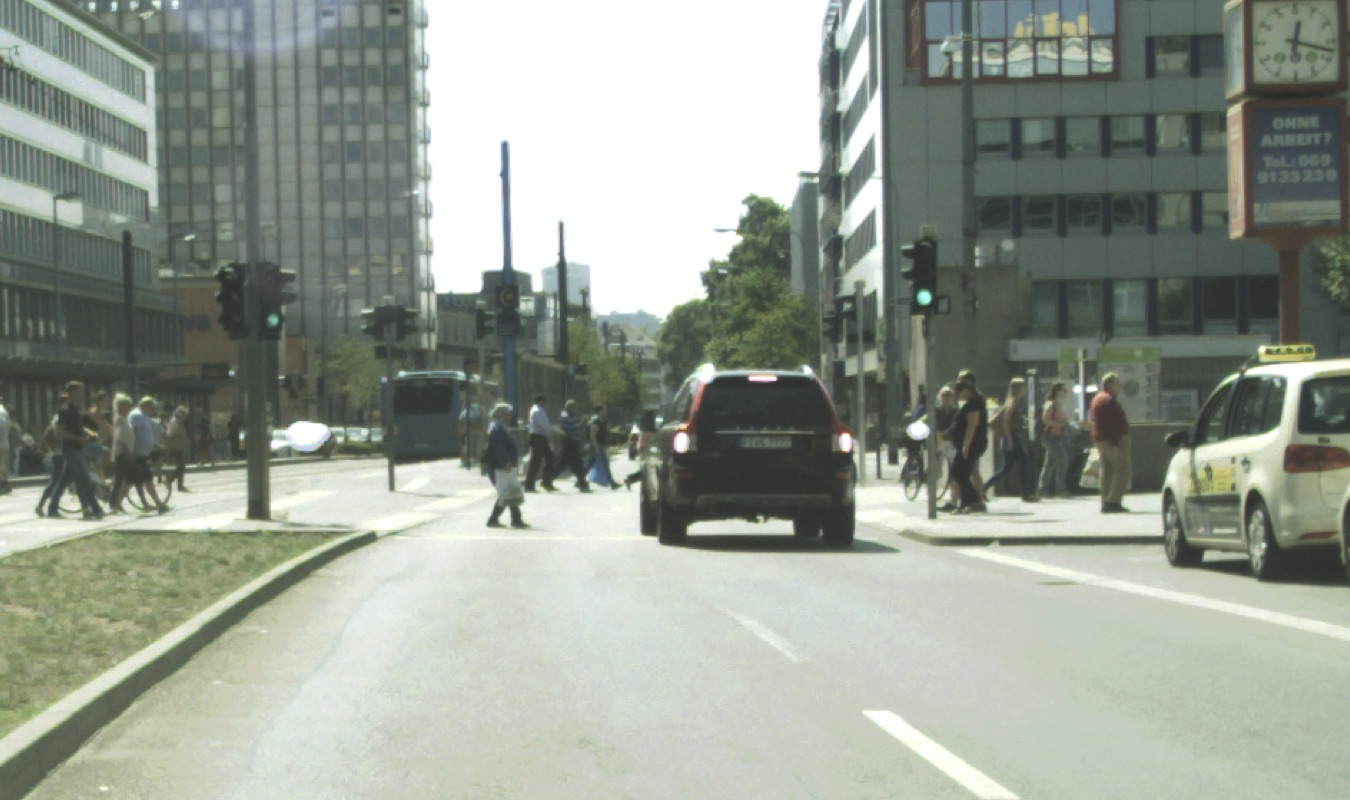}
    \caption{Scale (zoom in)}
    \label{fig:intro_illustration_b}
  \end{subfigure}
  \begin{subfigure}{0.483\linewidth}
    \centering
    \includegraphics[width=\linewidth,keepaspectratio=true]{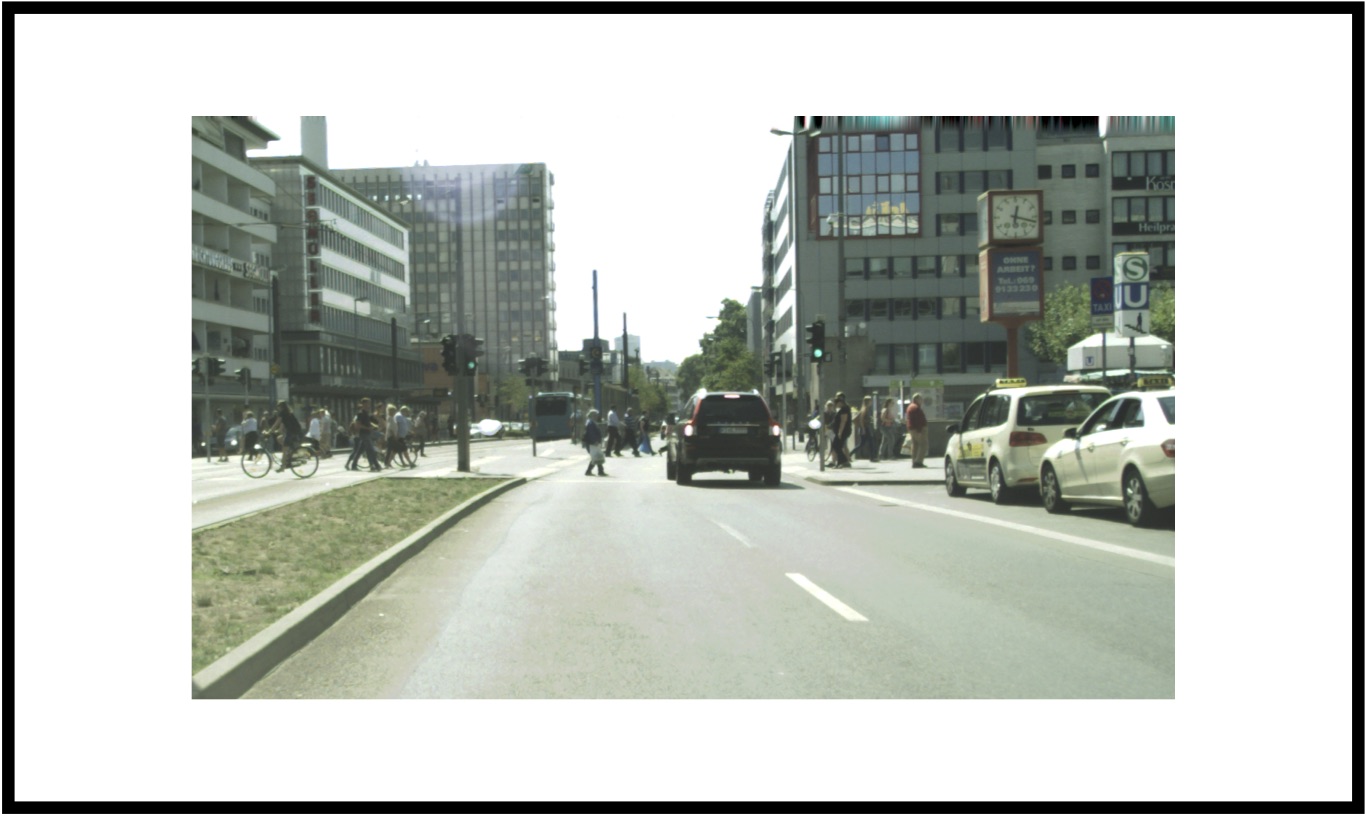}
    \caption{Scale (zoom out)}
    \label{fig:intro_illustration_b}
  \end{subfigure}
  \hfill
  \begin{subfigure}{0.483\linewidth}
    \centering
    \includegraphics[width=\linewidth,keepaspectratio=true]{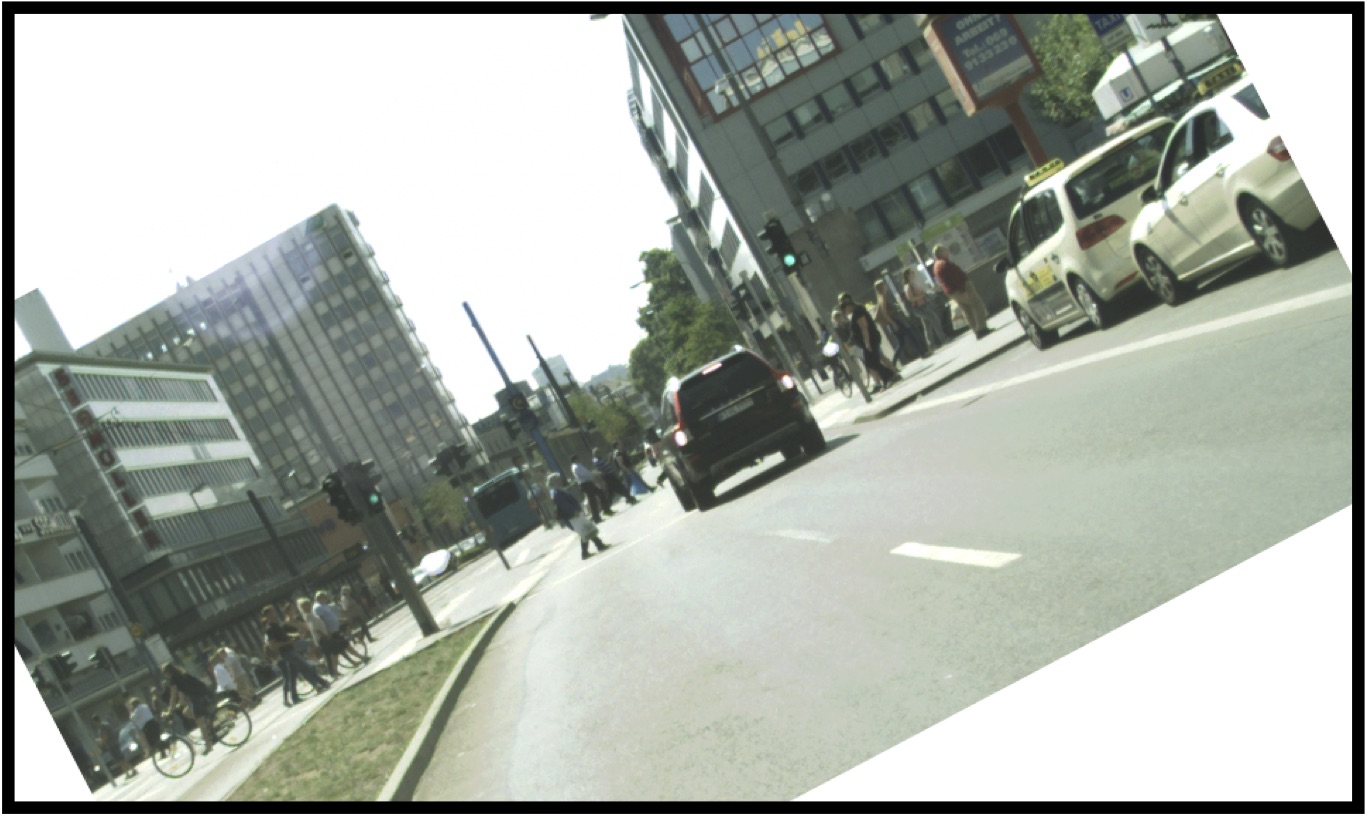}
    \caption{Rotate}
    \label{fig:intro_illustration_b}
  \end{subfigure}
  \caption{\textbf{Random affine transformations.} We consider adding random
    scaling and rotation before sending the images to the network.}
  \vspace{-0.1in}
\end{figure}

\vspace{-0.15in}
\paragraph{Full Algorithm:}
The full learning algorithm is summarized in Algorithm~\ref{alg:code}, in PyTorch style.

\begin{algorithm}[tbp]%
  \caption{\small Pseudocode in a PyTorch-like style. }
  \label{alg:code}
  \definecolor{codeblue}{rgb}{0.25,0.5,0.5}
  \lstset{
    backgroundcolor=\color{white},
    basicstyle=\fontsize{7.2pt}{7.2pt}\ttfamily\selectfont,
    columns=fullflexiblegit,
    breaklines=true,
    captionpos=b,
    commentstyle=\fontsize{7.2pt}{7.2pt}\color{codeblue},
    keywordstyle=\fontsize{7.2pt}{7.2pt},
    escapechar=\&
  }
  \begin{lstlisting}[language=python]
for I1, I2 in data_loader:
  with no_grad():
    flow = flownet(I1, I2) # [B, 2, H, W]

  v1, A1 = data_aug(I1)
  v2, A2 = data_aug(I2)

  h1 = encoder(v1)   # f_theta
  z1 = projector(h1) # g_theta
  p1 = predictor(z1) # q_theta

  with no_grad():
    h2 = target_encoder(v2) # f_xi
    z2 = target_projector(h2) # g_xi

  # upsample to match flow shape
  p1 = upsample(p1) # [B, C, H, W]
  z2 = upsample(z2) # [B, C, H, W]

  # warp the feature map with flow via bilinear interp
  T = apply(apply(inv(A1), flow), A2)
  inv_T = inv(T)
  p2 = transform(z2, inv_T)

  l = loss(normalize(p1), normalize(p2)) # Eq. (5)
  l.backward()

  optimizer.update([encoder, projector, predictor])
  momentum_update([target_encoder, target_projector])
\end{lstlisting}
\end{algorithm}

\subsection{Implementation Details}
\paragraph{Network architecture:} We use ResNet-50 as our base encoder network.
We increase the resolution of the output. Following~\cite{zhao2017pyramid}, we
use dilated convolution~\cite{chen2014semantic} and remove the the downsampling
operation in the last two stages of the encoder, which leads to an encoder with
output stride 8. The number of channels in the projector and predictor are the
same as BYOL~\cite{grill2020bootstrap}. To preserve the spatial dimensions of
the output feature maps, we remove the final global average pooling layer
in the encoder $f$. The linear layers in the following projector $g$ and
predictor $q$ are also replaced with $1\times 1$ convolutional layers to handle
the convolutional feature maps. Note that by using $1\times 1$ convolution, we
do not increase extra parameters comparing to the original BYOL.

\vspace{-0.1in}
\paragraph{Flow network:} For optical flow prediction, we use
RAFT~\cite{teed2020raft} as an off-the-shelf solution. The model is trained on
Flying Chair~\cite{dosovitskiy2015flownet}, Flying Things~\cite{mayer2016large}
and Sintel~\cite{butler2012naturalistic} datasets. All of these datasets
contain only synthetic data with no human labeling. The network is kept frozen
during our experiments.

\vspace{-0.1in}
\paragraph{Data augmentation:}
For color distortion and Gaussian blurring, we use the same parameters as the
ones used in SimCLR~\cite{chen2020simple}. For affine transformation, we do
random scale for $0.5\sim 2.0\times$ and rotate for $-30\sim 30$ degrees.

\vspace{-0.15in}
\paragraph{Flow post-processing:}
Since the affine transformation and flow operation are not strictly bijective
due to cropping and object occlusion, in the loss function we ignore any pixels
that have no correspondence. Occluded pixels then can be found by a
forward-backward flow consistency check~\cite{sundaram2010dense}.

%% file: section/experiment.tex
\section{Experiments}
\label{sec:exp}

\begin{table*}
	\centering
	\begin{tabular}{l|cccc|cccc}
		\toprule
		\multirow{2}{*}{Method}         & \multicolumn{4}{c|}{\ourdataset{}} & \multicolumn{4}{c}{BDD100K}                                                                                \\
		~                               & mIoU                               & mAP                         & mIoU$^\dag$ & mAP$^\dag$ & mIoU       & mAP       & mIoU$^\dag$ & mAP$^\dag$ \\
		\midrule
		Rand Init                       & 9.4                                & 0.0                         & 27.3        & 6.4        & 9.8        & 0.0       & 22.0        & 5.5        \\
		CRW~\cite{jabri2020space}       & 19.0                               & 0.0                         & 31.6        & 15.2       & 19.4       & 1.7       & 34.7        & 22.9       \\
		VINCE~\cite{gordon2020watching} & 30.6                               & 0.9                         & 47.4        & 17.8       & 23.2       & 0.1       & 39.5        & 23.8       \\
		\ourmethod{} (Ours)             & {\bf 49.6}                         & {\bf 5.8}                   & {\bf 61.7}  & {\bf 19.0} & {\bf 37.6} & {\bf 5.8} & {\bf 49.8}  & {\bf 24.9} \\
		\midrule
		End-to-end supervised           & 63.3                               & 2.2                         & 67.0        & 16.5       & 52.0       & 8.0       & 56.6        & 20.0       \\
		\bottomrule
	\end{tabular}
	\vspace{0.1in}
	\caption{
		\textbf{Self-supervised learning results on \ourdataset{} and BDD100K}, in
		comparison with other self-supervised video representation learning methods. All
		readouts are done with a frozen backbone except for the ``end-to-end
		supervised'' entry. Results with $^\dag$ are obtained with heavier readout header.}
	\vspace{0.1in}
	\label{tab:exp_main_results}
\end{table*}

\begin{table*}
	\centering
	\begin{tabular}{l|c|cccc|cccc}
		\toprule
		\multirow{2}{*}{Method}         & \multirow{2}{*}{Train data} & \multicolumn{4}{c|}{\ourdataset{}} & \multicolumn{4}{c}{BDD100K}                                                                                \\
		~                               & ~                           & mIoU                               & mAP                         & mIoU$^\dag$ & AP$^\dag$  & mIoU       & mAP       & mIoU$^\dag$ & mAP$^\dag$ \\
		\midrule
		Supervised                      & ImageNet                    & 39.6                               & 3.3                         & 57.7        & 18.8       & 34.0       & 3.6       & {\bf 52.4}  & 24.9       \\
		SimCLR~\cite{chen2020simple}    & ImageNet                    & 37.0                               & 3.0                         & 58.6        & {\bf 21.0} & 28.1       & 2.7       & 51.0        & {\bf 26.8} \\
		BYOL~\cite{grill2020bootstrap}  & ImageNet                    & 35.4                               & 2.4                         & 59.8        & 19.5       & 28.3       & 2.8       & {\bf 52.4}  & 26.0       \\
		VINCE~\cite{gordon2020watching} & R2V2                        & 23.6                               & 1.2                         & 57.4        & 18.1       & 19.4       & 1.4       & 47.0        & 24.2       \\
		\ourmethod{} (Ours)             & -                           & {\bf 49.6}                         & {\bf 5.8}                   & {\bf 61.7}  & 19.0       & {\bf 37.6} & {\bf 5.8} & 49.8        & 24.9       \\
		\bottomrule
	\end{tabular}
	\caption{\textbf{Readout results on \ourdataset{} and BDD100K}, in comparison
		with competitive representation learning methods trained on other data
		sources.}
	\label{tab:exp_imagenet}
\end{table*}

We first train our model on two self-driving datasets \ourdataset{} and
BDD100K~\cite{Yu_2020_CVPR} to evaluate the quality of the learned
representations, we target semantic and instance segmentation as well as object
detection as the readout tasks using labeled images.
Ablation experiments are conducted on
\ourdataset{} to verify the effectiveness of each component of our model.
We further test the transferability of the learned feature by solely performing readout experiments
on the Cityscapes dataset~\cite{cordts2016cityscapes} with models trained on
\ourdataset{} and BDD100K.

\subsection{Datasets}
We evaluate our method on the following driving video datasets containing
complex visual scenes.
\begin{itemize}[leftmargin=*]
	\item
	      \textbf{\ourdataset{}} is an in-house large-scale self-driving datasets collected by
	      ourselves. It contains around 15,000 video snippets where each is about 25
	      seconds long at 1080p and 10 fps, with a total of 3.5 million frames. Within
	      these, 11,580 and 1,643 images are densely labeled, for training and validation
	      respectively. They contain 7 instance classes and 13 semantic classes. We
	      uniformly sampled 256,000 frames from the videos with a 0.4 second time
	      interval as training frame pairs. In readout setting, we use the annotated
	      train and val split to perform semantic and instance segmentation tasks.

	\item
	      \textbf{BDD100K}~\cite{Yu_2020_CVPR} is a large-scale self-driving datasets
	      which contains 100,000 unlabeled raw video snippets for street scene, where
	      each is about 40 seconds long at 720p and 30 fps. It captures different weather
	      conditions, including sunny, overcast and rainy, as well as different times of
	      day including nighttime. The class definition is the same as
	      Cityscapes~\cite{cordts2016cityscapes} which consists of 8 instance classes for
	      object detection, and 19 classes in total for semantic segmentation. 7,000
	      train and 1,000 val images are densely labeled for semantic segmentation;
	      70,000 train, 10,000 val images are labeled for object detection. We use the
	      70,000 video snippets in the official training split to perform self-supervised
	      learning. At each iteration we will randomly sample two frames with 0.5 second
	      time interval from a video, and no further filtering or frame picking strategy
	      is applied. For evaluation, we use the annotated images to perform readout
	      experiments on semantic segmentation and object detection.

	\item
	      \textbf{Cityscapes}~\cite{cordts2016cityscapes} is another self-driving dataset
	      that contains 5000 images of ego-centric driving scenarios in urban settings
	      which are split into 2975, 500 and 1525 for training, validation and testing
	      respectively. It consists of 8 instance classes and 11 semantic classes. Due to
	      lack of large amount of labeled data, we are interested in investigating
	      whether the representations learned from other source video datasets can be
	      transferred to a new dataset easily. Therefore we perform readout experiments
	      on Cityscapes using models that are pretrained on \ourdataset{} and BDD100K.

\end{itemize}

\subsection{Competitive Methods}
We compare to the following recent competitive methods for representation
learning from video data:
\begin{itemize}[leftmargin=*]
	\item \textbf{CRW}~\cite{jabri2020space} is a self-supervised approach to learn
	      representations for visual correspondence. We use 5 frames with a 0.1 second
	      time interval as inputs.

	\item \textbf{VINCE}~\cite{gordon2020watching} is one of the latest video
	      representation learning method that leverages multi-label contrastive
	      objective. We also train a VINCE model which is a recent proposed approach that
	      extends MoCo~\cite{he2020momentum} and learn representation from videos. We use
	      inputs of 4 frames with a 0.1 second time interval.
\end{itemize}

For fair comparison, we train these methods on our driving video datasets and
we have tried our best to search for their best hyperparameters. Additionally,
we also compare our methods with pretrained SimCLR~\cite{chen2020simple} and
BYOL~\cite{grill2020bootstrap} from ImageNet. Note that we have also tried to
apply SimCLR and BYOL on driving videos but they tend to perform very poorly
since they are designed for clean visual scenes with mostly a single object. We
therefore defer these results to the supplementary materials.

\subsection{Experimental Setup}
\paragraph{FlowE:} We use 64 GPUs with 2 video frame pairs per GPU.
LARS~\cite{you2017large} optimizer is used with a cosine decay learning rate
schedule without restart~\cite{loshchilov2016sgdr} and an initial learning rate
0.1 with weight decay 1e-6. The setting of the exponential moving average
parameter of the target network is kept the same as the original BYOL paper.
For \ourdataset{}, we will randomly scale the image pairs from $0.75 \sim
	1.25\times$, and randomly crop a $512\times 1024$ patch pair at the same
location of the two images; models are run for 160,000 iterations (80 epochs).
For BDD100K, we first upsample the images to be $1080\times 1920$, and follow
the same setting for \ourdataset{}; model are run for 60,000 iterations (110
epochs), and it is worth noting that the performance has not saturated and
longer iteration may yield better performance.

\vspace{-0.05in}
\paragraph{Readout setup:} For the semantic segmentation task, we train models
for 60,000 iterations of SGD with batch size 16, initial learning rate 0.02
and the ``poly'' learning rate decay schedule~\cite{chen2017deeplab} on both
datasets. Patches of size $512\times 1024$ are randomly cropped from images
which are randomly resized with shorter side from 512 to 2048.

For the instance segmentation task on
\ourdataset{}, we train models for 32 epochs of SGD with batch size 8 and
initial learning rate 0.01 with a decay factor 0.1 at epoch 28. Multi-scale
training is used with shorter side from 800 to 1024.

For object detection task on BDD100K we train models for 12 epochs of SGD with
mini batch size 16 and initial learning rate 0.02 with a decay factor 0.1 at
epoch 8 and 11, respectively, we keep the image resolution as it is and do not
apply multi-scale training.

\vspace{-0.05in}
\paragraph{Standard readout header:} In our readout setting, the encoder is
frozen and only the newly added layers are trained. Just like the linear
evaluation protocol on
ImageNet~\cite{he2020momentum,chen2020simple,grill2020bootstrap}, we aim to add
as few parameters as possible. Therefore, we use DeepLab
v1~\cite{chen2014semantic} as our semantic segmentation model as it has no
extra heavy decoder like DeepLab V3~\cite{chen2017rethinking}. Besides dilated
convolutions are used in the encoder, only one convolutional layer is added on
top of the encoder to output per-pixel classification logits.

Similarly, for object detection on BDD100K, we use Faster R-CNN with the
ResNet-C4 architecture which is proposed in \cite{he2016deep}. Only a small
number of parameters are introduced: a small convnet RPN~\cite{ren2015faster}
and two linear layers for bounding box classification and regression.

For instance segmentataion on \ourdataset{}, the same ResNet-C4 architecture is
used with two more convolutional layers added for instance mask prediction as
was done in~\cite{he2017mask}.

\vspace{-0.05in}
\paragraph{Heavier readout header:}
While we believe that standard readout headers should be mainly used to
evaluate the quality of representations since there are less number of extra
parameters, they may not be capable enough to capture the complex output
structure for semantic and instance segmentation. To provide a stronger
comparison, following LoCo~\cite{xiong2020loco}, we also perform readout with a
\textit{heavier header} such DeepLab V3 decoder and FPN-style Faster and Mask
R-CNN, where results obtained with these models are denoted with mIoU$^\dag$
and mAP$^\dag$.

\subsection{Main Results}
\paragraph{Results trained on \ourdataset{} and BDD100K:} The results on
\ourdataset{} and BDD100K are shown in Table~\ref{tab:exp_main_results}. We
compare
\ourmethod{} with various baselines, including Random initialization (readout
from a random projection), VINCE~\cite{gordon2020watching} and
CRW~\cite{jabri2020space}; and our method is able to surpass them by a large
margin. CRW has poor performance on semantic segmentation since it focuses on
video correspondence as its training objective, and the features for different
classes of static objects will not be easily differentiated. For VINCE, we can
see that it can successfully learn some useful features from video data.
However, our method is still significantly better.

\vspace{-0.1in}
\paragraph{Results trained on other data:}
We also compare \ourmethod{} with methods trained on other datasets like ImageNet, including
supervised learning, SimCLR~\cite{chen2020simple} and BYOL~\cite{grill2020bootstrap} on ImageNet,
and VINCE~\cite{gordon2020watching} on R2V2~\cite{gordon2020watching}. The results are shown in
Table~\ref{tab:exp_imagenet}. We simply freeze the pre-trained model weights and perform readout
experiments on \ourdataset{} and BDD100K. For supervised learning baseline, we use the ResNet-50
checkpoint provided by torchvision. For SimCLR, we use our own implementation and train a model with
69.8\% top-1 accuracy on ImageNet. For BYOL and VINCE, we use weights released online by the
authors. Our methods can outperform or stay on par with other strong baselines in most cases,
especially when using a standard readout header. It is worth noting that Supervised/SimCLR/BYOL are
three very strong baselines that pretrained on ImageNet, a large scale and heavily curated dataset.
Although it is not easy to beat these state-of-the-art ImageNet methods, we still manage to surpass
them on three out of the four metrics. Importantly, our framework can directly learn semantically
meaningful representation from raw video data, making it practical for real world applications
where offline curated datasets are not available at hand.

\begin{table}
	\centering
	\resizebox{\linewidth}{!}{
		\begin{tabular}{ccc|cc|cc}
			\toprule
			\begin{tabular}[c]{@{}c@{}} Pixel  \\ based   \end{tabular} & \begin{tabular}[c]{@{}c@{}} Affine  \\ transform   \end{tabular} & \begin{tabular}[c]{@{}c@{}} Optical \\ flow   \end{tabular} & mIoU       & mAP       & mIoU$^\dag$ & mAP$^\dag$ \\
			\midrule
			\checkmark                    &                               &                               & 21.3       & 0.7       & 40.1        & 12.3       \\
			\checkmark                    & \checkmark                    &                               & 28.7       & 2.7       & 45.9        & 15.1       \\
			\checkmark                    &                               & \checkmark                    & 37.3       & 3.3       & 51.9        & 16.2       \\
			                              & \checkmark                    & \checkmark                    & 17.8       & 0.7       & 33.1        & 10.9       \\
			\checkmark                    & \checkmark                    & \checkmark                    & {\bf 37.9} & {\bf 3.8} & {\bf 53.2}  & {\bf 16.5} \\
			\bottomrule
		\end{tabular}
	}
	\caption{\textbf{Ablation studies on different design choices.} Numbers show
		semantic segmentation and instance segmentation readout results on
		\ourdataset{}.}
	\vspace{0.1in}
	\label{tab:exp_ablation}
\end{table}

\subsection{Ablation Studies}
\label{sec:exp_ablation}
We perform ablation studies and show the results in Table~\ref{tab:exp_ablation}. All
entries are trained with 16K iterations for faster experimentation. When
bringing video data with flow matching, we can see a huge performance
improvement, indicating the importance of equivariance objective derived from
videos. For non pixel-based variant, we simply use a global average pooling
after the encoder and get vector representation. Its poor performance on the
readout tasks suggests the necessity of keeping spatial dimension of the
representation. Random affine transformation can also bring some additional
gains. Finally, our full model achieve the best performance.

\label{sec:exp_transferability}
\subsection{Representation Transferability on Cityscapes}
When there is limited labeled data on a new driving dataset, it is often
desirable to learn unsupervised representations from a large scale unlabeled
driving videos of another source. However, standard self-supervised method only
works on static images with few objects. Although ImageNet pretrained
checkpoints are readily available online, there may exist a large domain gap.

In this section, we further test the transferability of the learned
representations of
\ourmethod{} by performing semantic and instance segmentation readout
experiments on the Cityscapes dataset~\cite{cordts2016cityscapes}. Our models
are pretrained on \ourdataset{} and BDD100K. Following the common practice, for
instance segmentation we train 64 epochs with batch size 8, initial learning
rate 0.01 with a decay by a factor of 10 at epoch 56; for semantic
segmentation, we train 40,000 iterations with batch size 8, initial learning
rate 0.01 with ``poly'' learning rate decay schedule. The results are shown in
Table~\ref{tab:exp_transfer}. The results are highly consistent with the
evaluation in Table~\ref{tab:exp_imagenet}, our method can perform better or on
par compared to ImageNet pre-training, which suggests that our method can be
seen as a better alternative way to bootstrap representations from a large
number of unlabeled videos.
We also tried to use the intermediate activation of the RAFT model that is trained on Flying Chair,
Flying Things and Sintel (C+T+S) for semantic readout evaluation. Specifically, we use the RAFT
feature encoder as the backbone to replace ResNet-50, and add a DeepLab v1/v3 decoder as a
standard/heavier header for semantic segmentation readout. The results clearly show that the
representations from the optical flow model does not contain rich semantic information

\begin{table}
	\centering
	\resizebox{\linewidth}{!}{
		\begin{tabular}{l|c|cc|cc}
			\toprule
			Method              & Train data    & mIoU       & mAP       & mIoU$^\dag$ & mAP$^\dag$ \\
			\midrule
			Supervised          & ImageNet      & 43.8       & 6.1       & 59.9        & 25.3       \\
			SimCLR              & ImageNet      & 39.9       & 5.0       & 60.3        & {\bf 28.9} \\
			BYOL                & ImageNet      & 38.2       & 4.1       & 59.8        & 27.4       \\
			VINCE               & R2V2          & 26.7       & 1.1       & 57.5        & 25.6       \\
			RAFT                & C+T+S         & 10.5       & -         & 32.4        & -          \\
			\ourmethod{} (Ours) & BDD100K       & 45.6       & 5.7       & 56.6        & 25.3       \\
			\ourmethod{} (Ours) & \ourdataset{} & {\bf 51.1} & {\bf 7.4} & {\bf 63.7}  & 28.1       \\
			\bottomrule
		\end{tabular}
	}
	\caption{\textbf{Readout results on Cityscapes} with representations learned
		from other datasets. }
	\label{tab:exp_transfer}
\end{table}

\begin{figure*}
	\centering
	\includegraphics[width=0.313\linewidth]{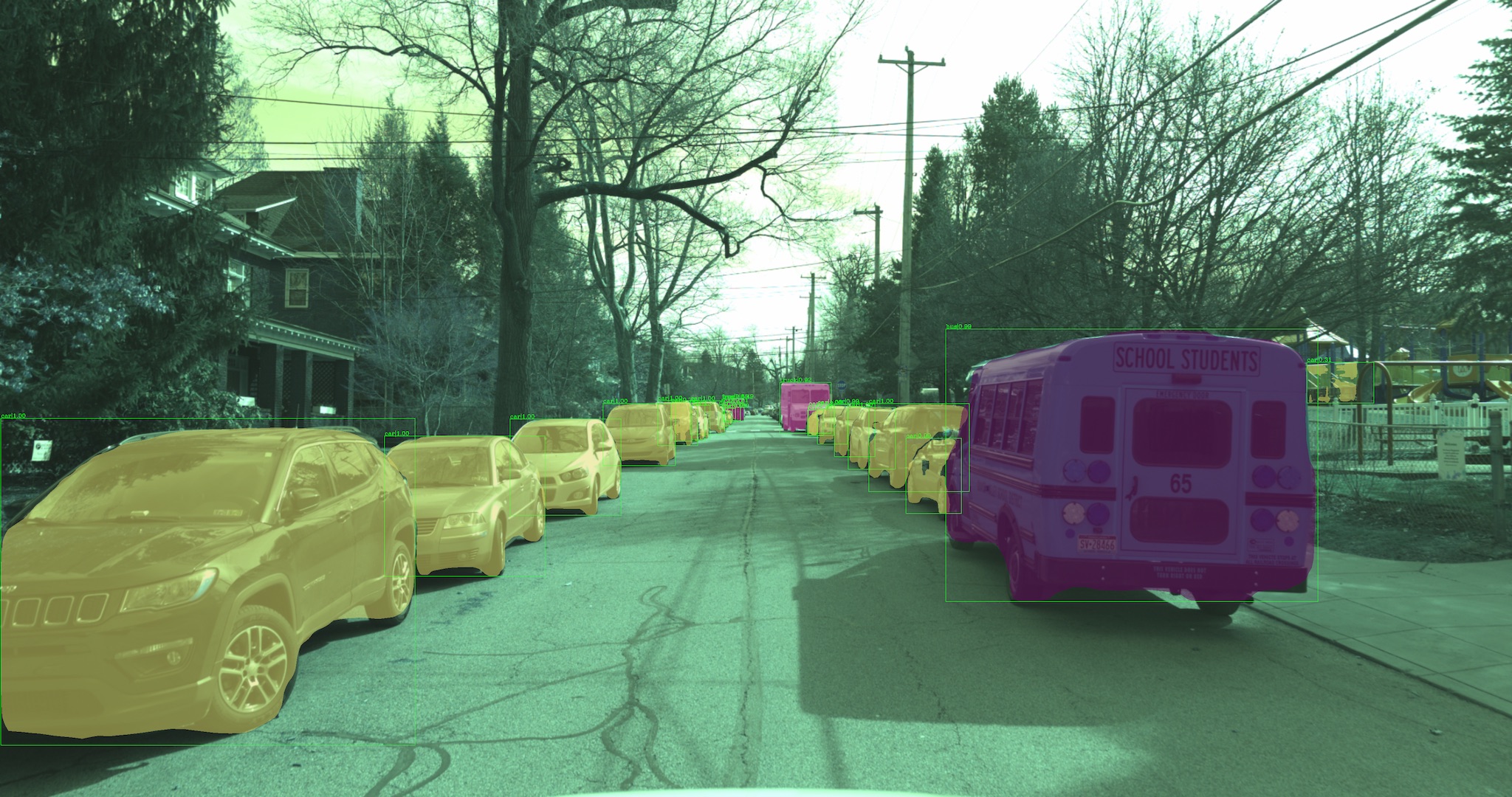}
	\hfill
	\includegraphics[width=0.293\linewidth]{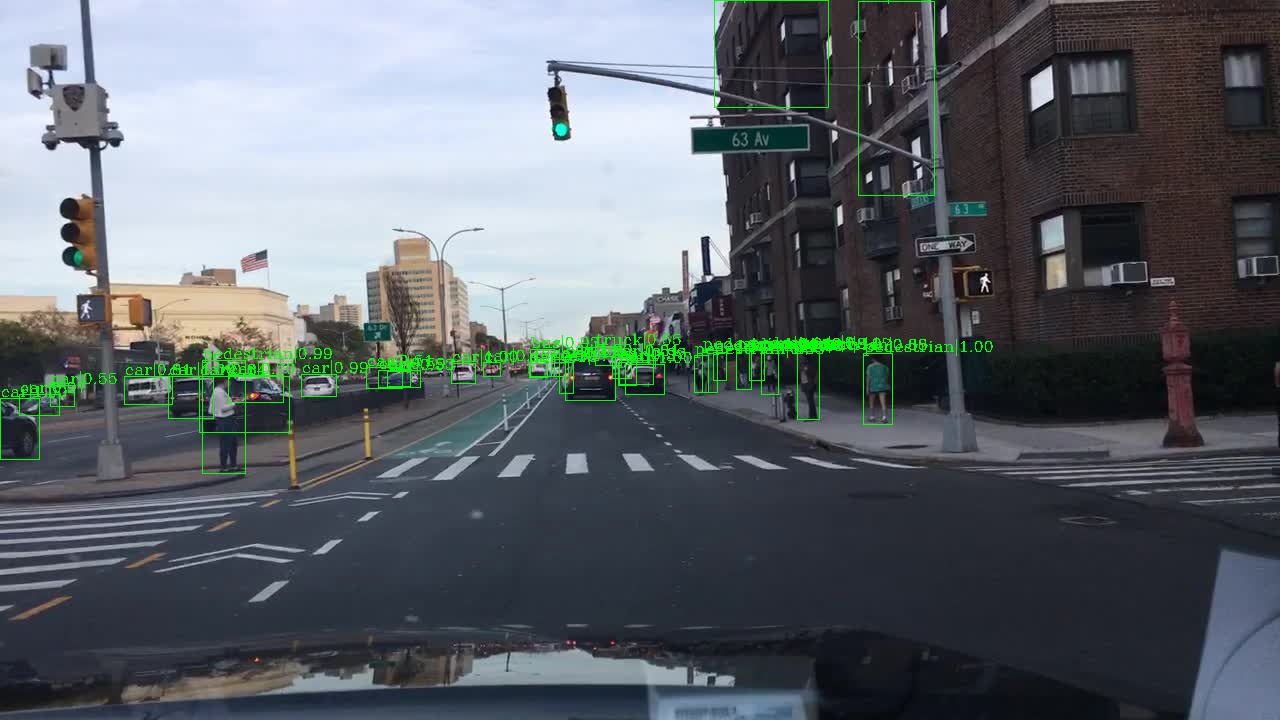}
	\hfill
	\includegraphics[width=0.33\linewidth]{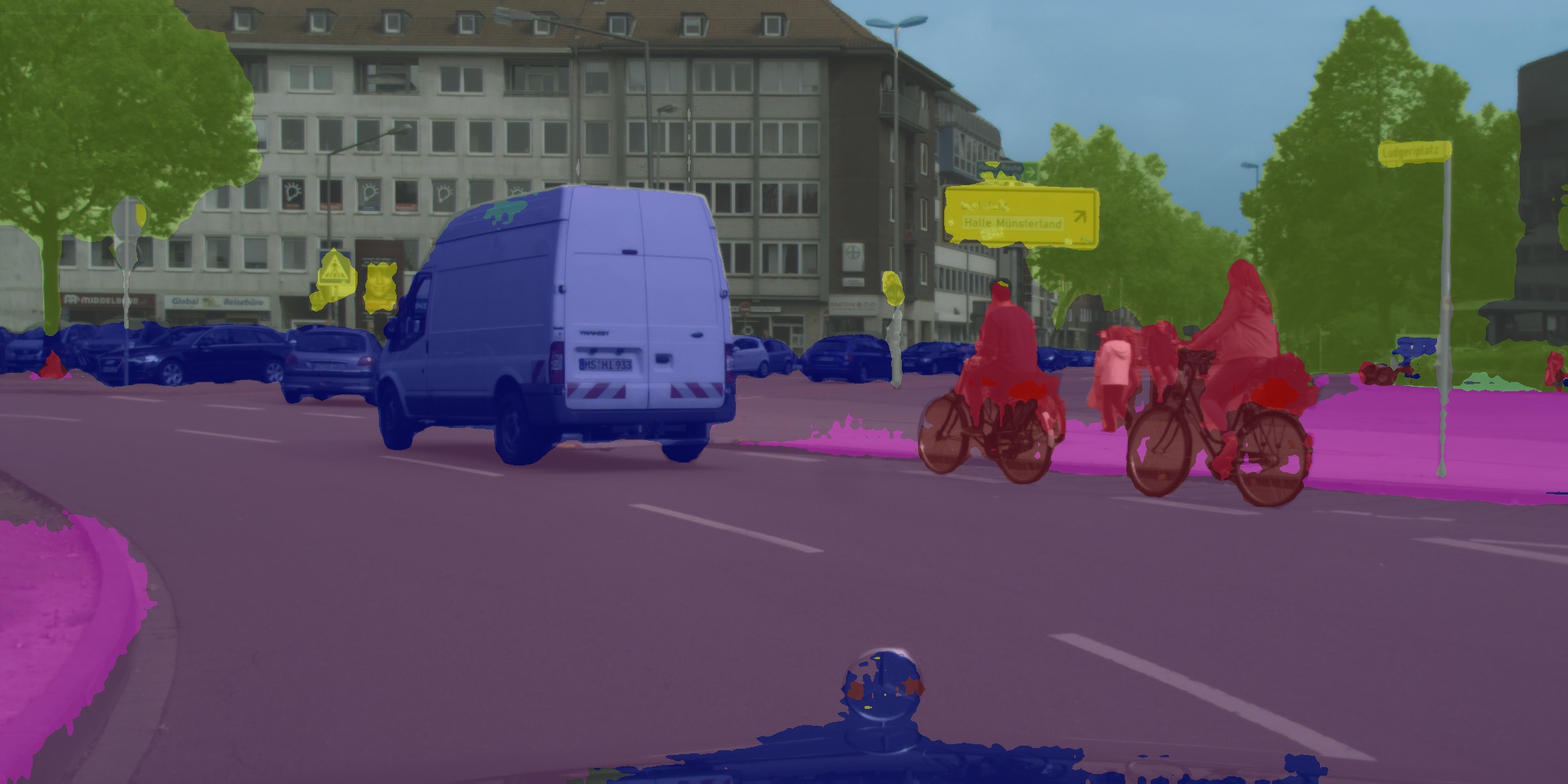}
	\hfill
	\includegraphics[width=0.313\linewidth]{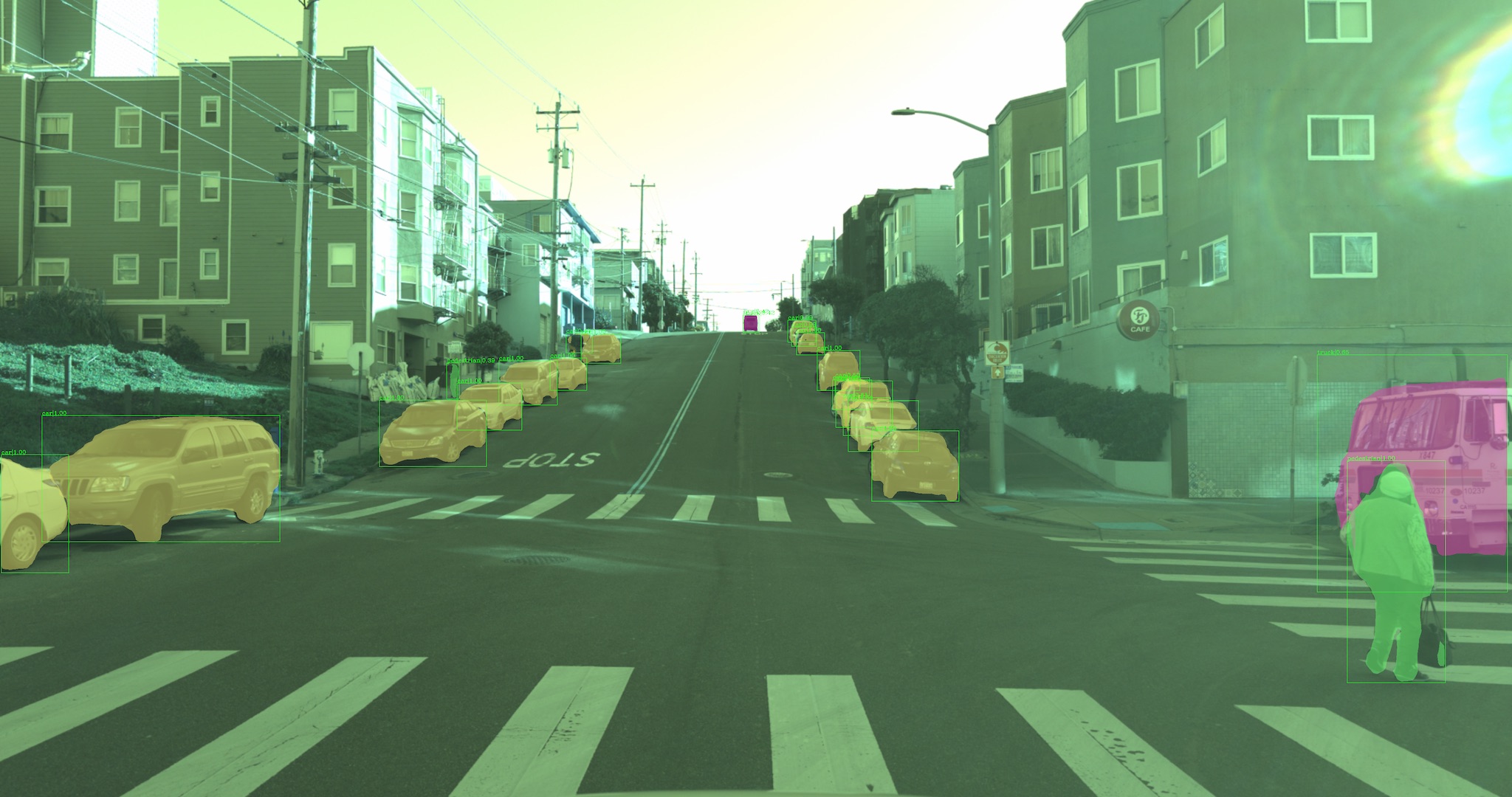}
	\hfill
	\includegraphics[width=0.293\linewidth]{figure/vis_results/faster_rcnn_bdd/b291cfd6-a6275049}
	\hfill
	\includegraphics[width=0.33\linewidth]{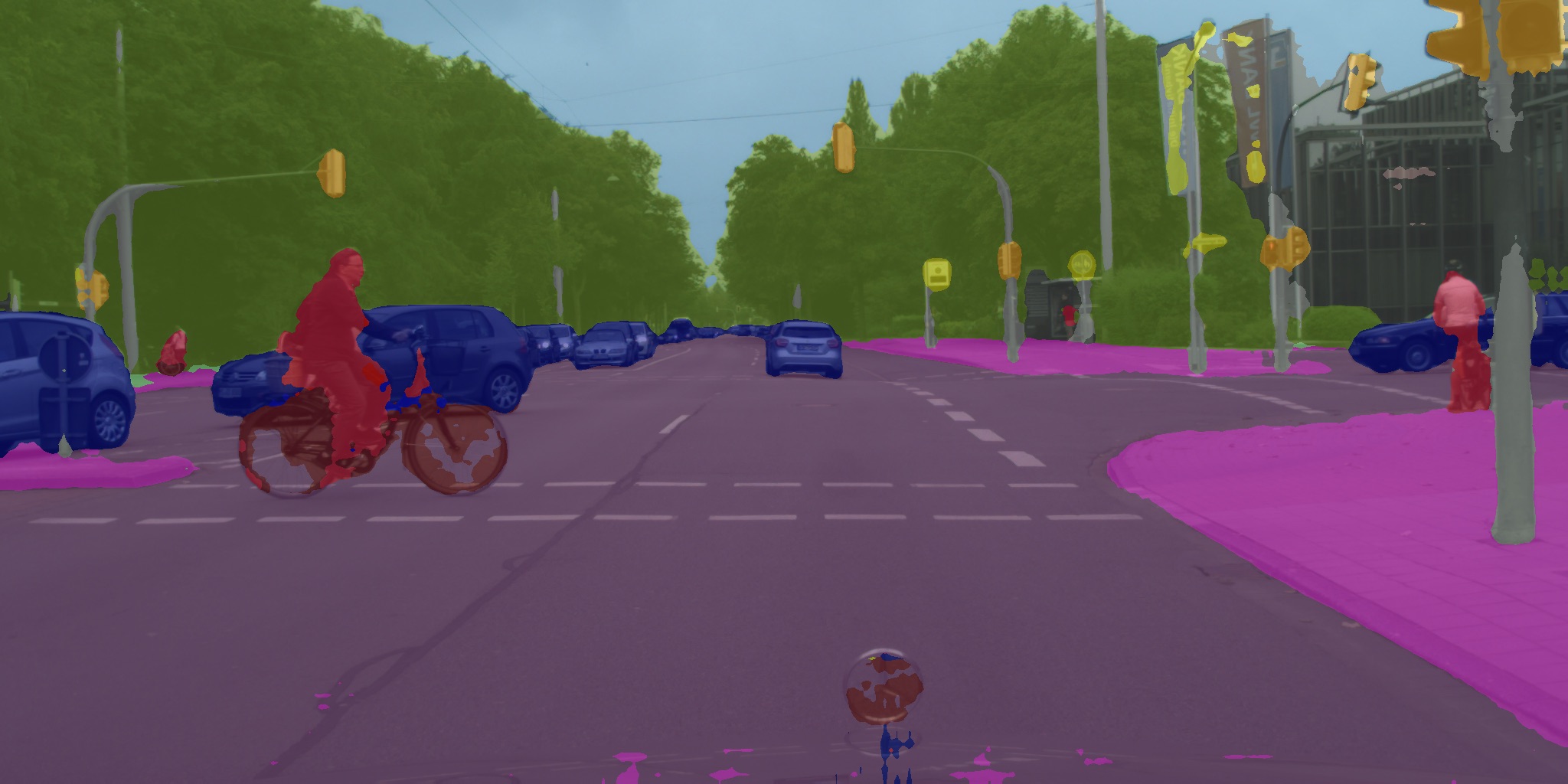}
	\hfill
	\includegraphics[width=0.313\linewidth]{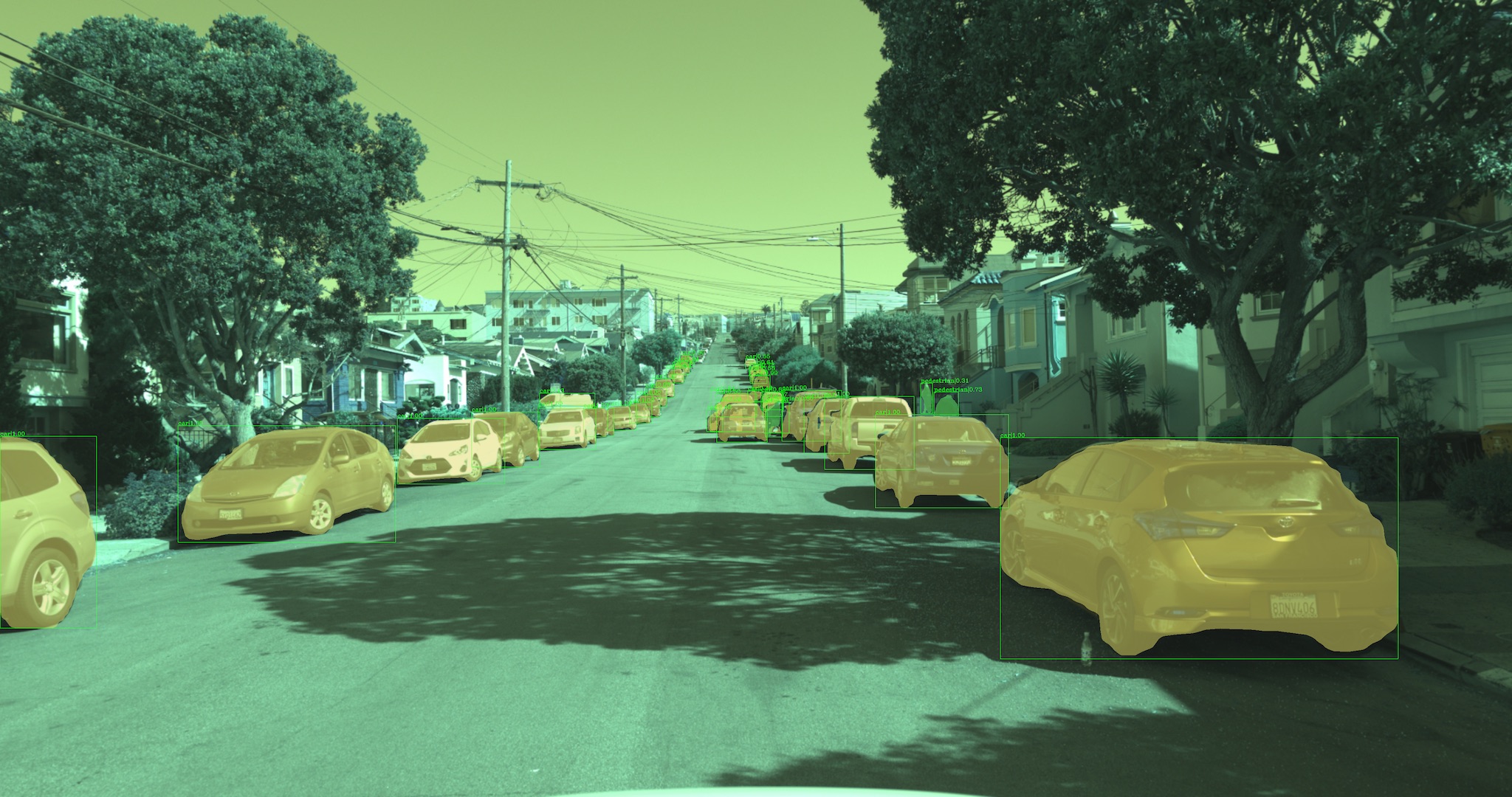}
	\hfill
	\includegraphics[width=0.293\linewidth]{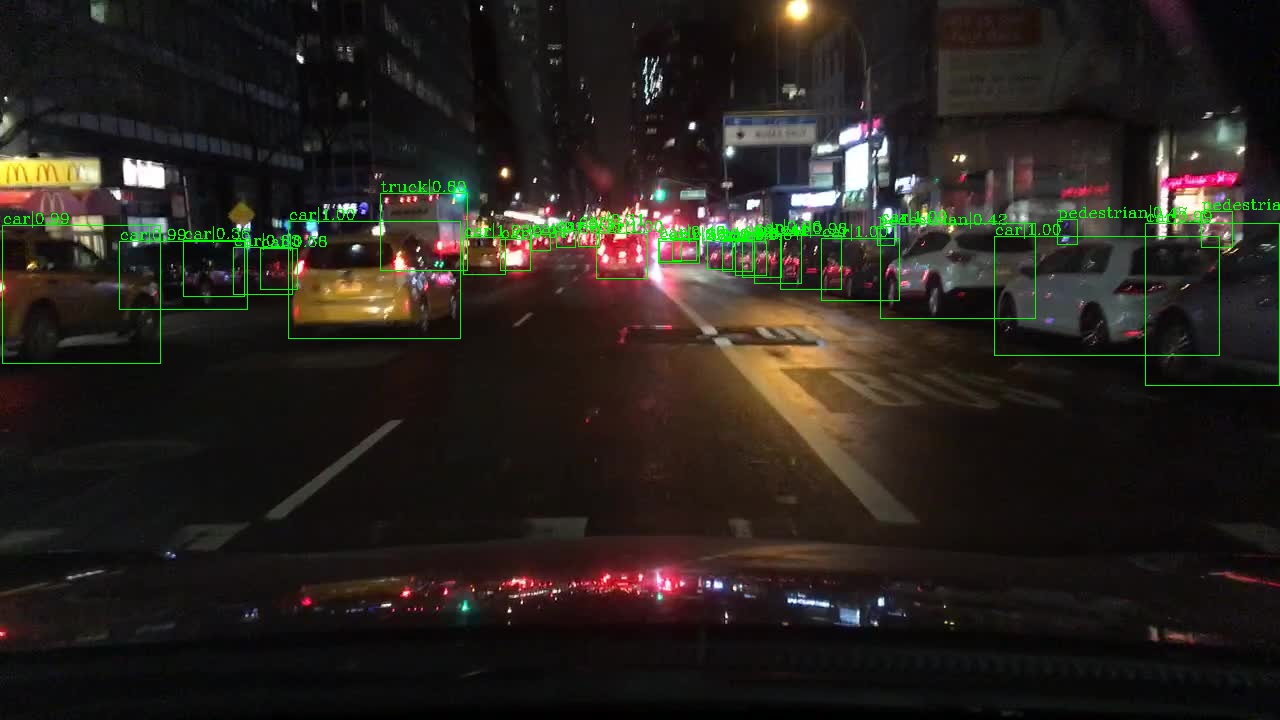}
	\hfill
	\includegraphics[width=0.33\linewidth]{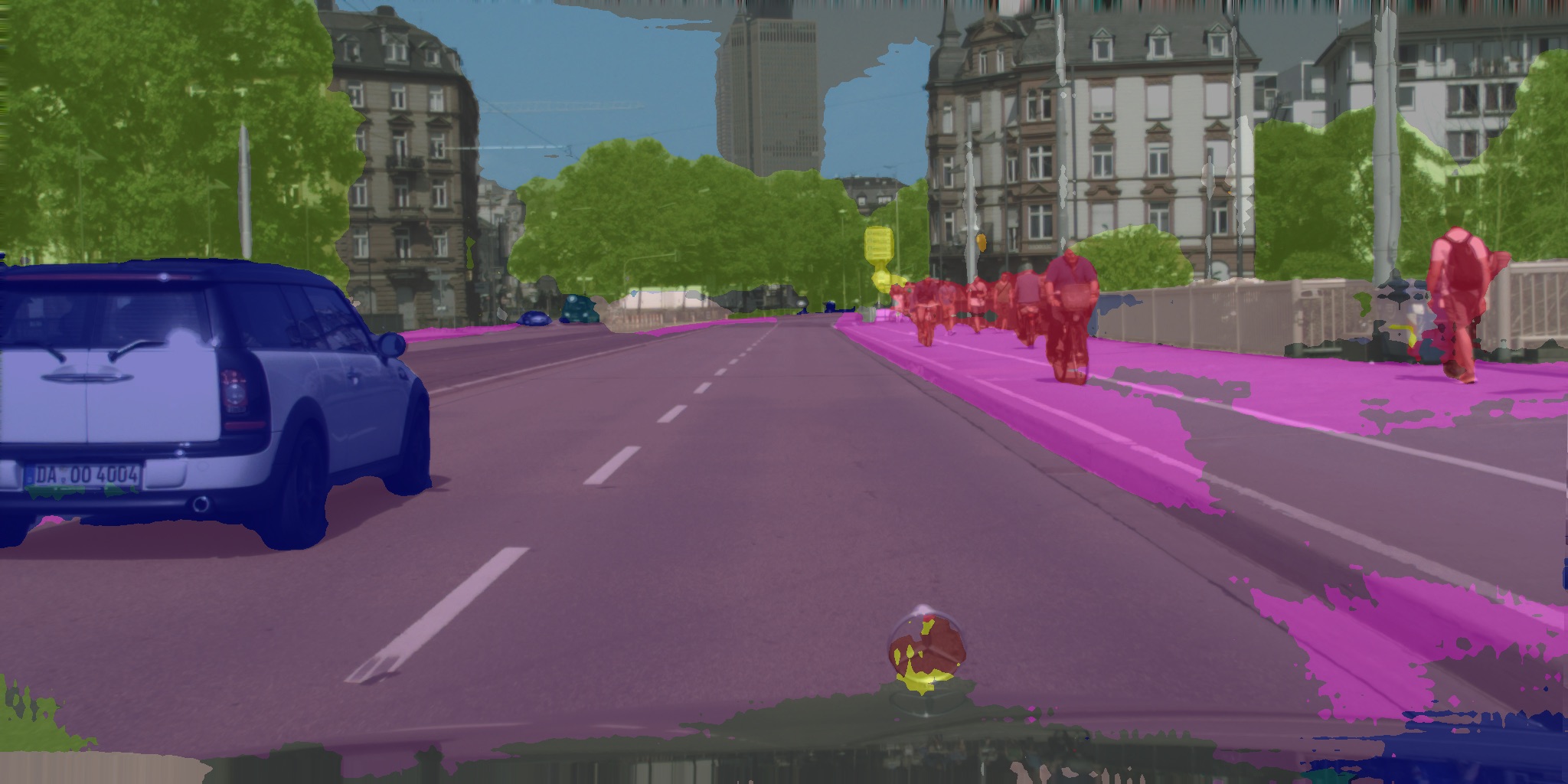}
	\hfill
	\includegraphics[width=0.313\linewidth]{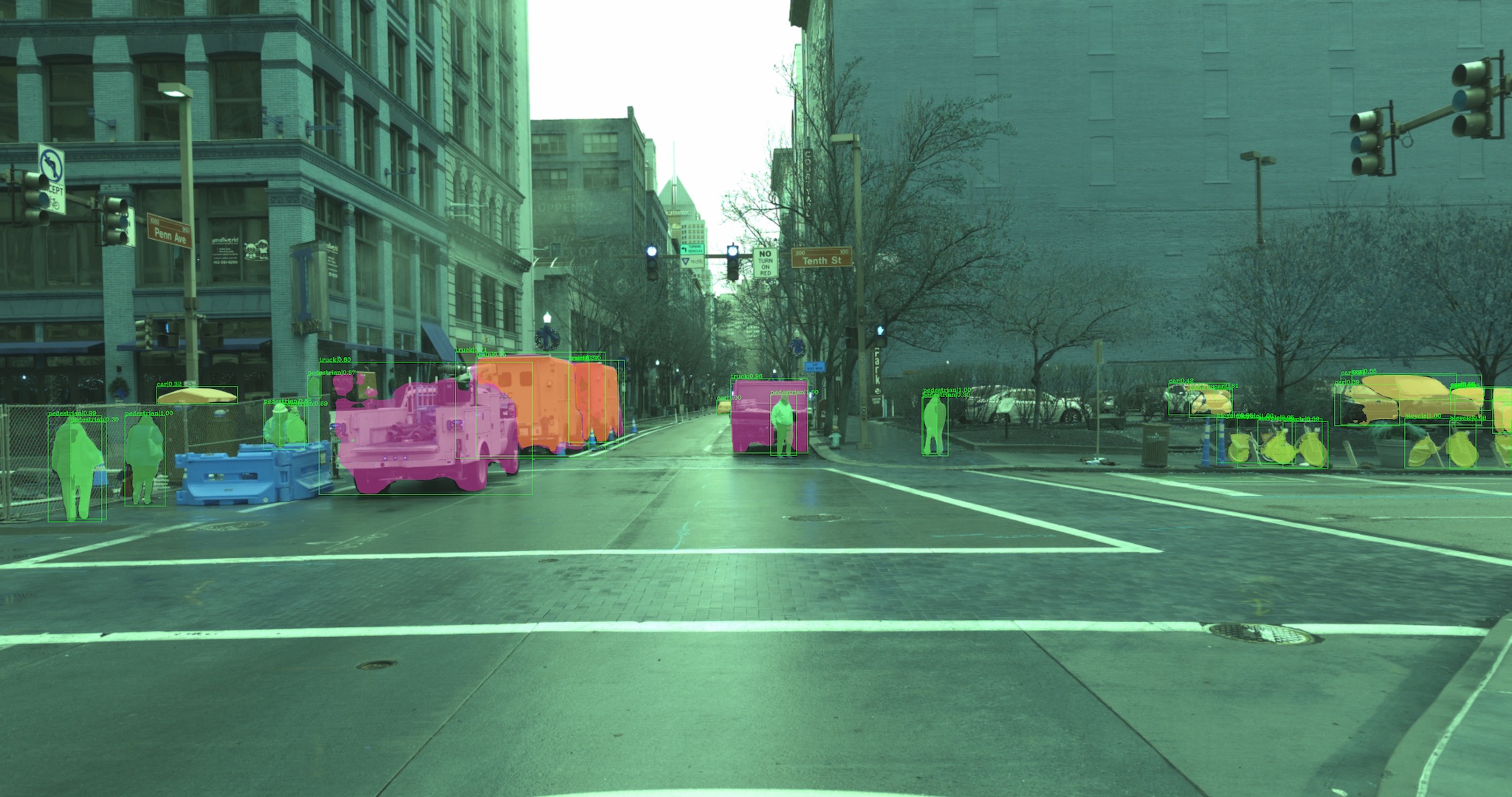}
	\hfill
	\includegraphics[width=0.293\linewidth]{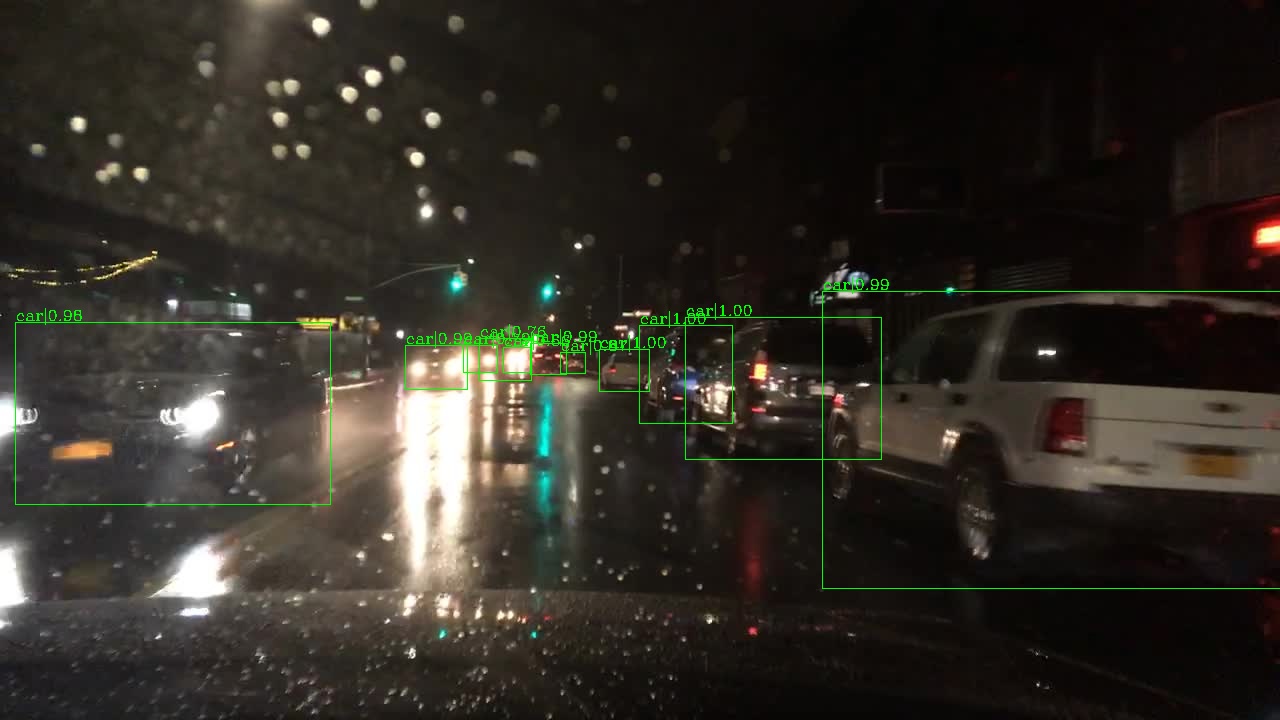}
	\hfill
	\includegraphics[width=0.33\linewidth]{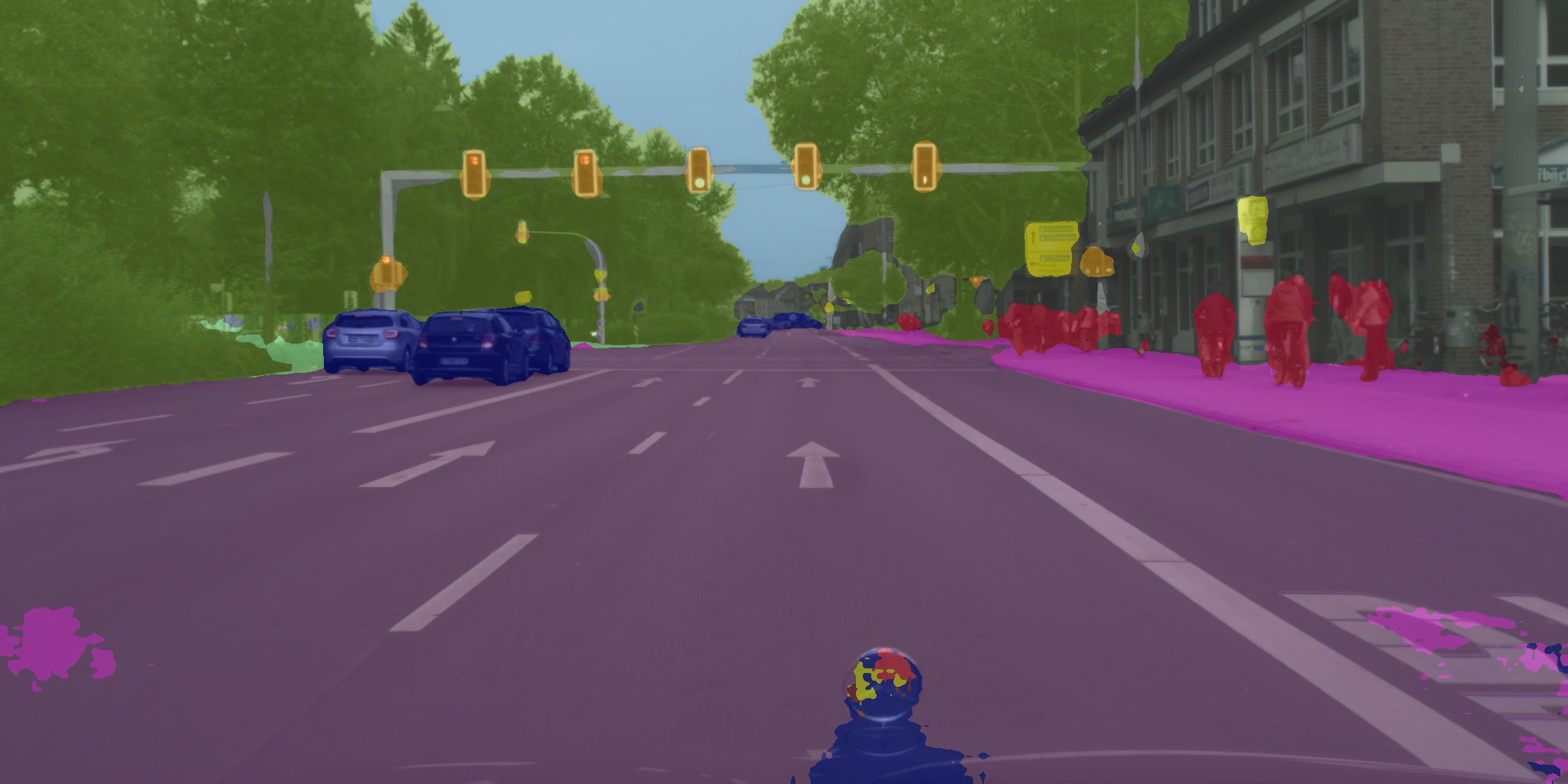}
	\vspace{0.1in}
	\caption{\textbf{Readout visualization.} Left: Instance segmentation on
		\ourdataset{}; middle: object detection on BDD100K; right: semantic
		segmentation on Cityscapes. For
		\ourdataset{} and BDD100K, models are trained on the corresponding dataset with
		heavier readout headers. For Cityscapes, we use a \ourdataset{} pretrained
		model with a standard readout header. }
	\label{fig:vis_results}
	\vspace{0.1in}
\end{figure*}

\begin{table}
	\centering
	\begin{tabular}{c|ccc}
		\toprule
		\% of labels          & 1\%           & 10\%          & 100\%         \\
		\midrule
		End-to-end supervised & 42.0          & 59.5          & 63.3          \\
		FlowE (Ours)          & \textbf{53.9} & \textbf{64.0} & \textbf{68.8} \\
		\bottomrule
	\end{tabular}
	\caption{{\bf Semantic segmentation results with limited labeled data on \ourdataset{}.}
		We compare FlowE with the end-to-end supervised baseline. Our model is
		pretrained on unlabeled videos and then finetuned on the labeled data.	 %
	}
	\vspace{-0.1in}
	\label{tab:semi}
\end{table}
\subsection{Learning with limited labeled data}
Another very practical setting is semi-supervised learning, where a large
video dataset is recorded but only a very small portion of the dataset
is annotated.
To investigate whether our algorithm can reduce the reliance on labeled data,
we randomly subsample 1\%, 10\% labeled data from UrbanCity, and finetune our
pretrained models on the supervised task of semantic segmentation. We compare
it with the end-to-end supervised learning baseline. %
As shown in Table~\ref{tab:semi}, pretraining on unlabeled video data can
significantly boost the performance when the labels are scarce, and pretraining
is still beneficial even when using 100\% of the labeled data.

\subsection{Visualization}
We show the visualization results of instance segmentation on \ourdataset{},
object detection on BDD100K~\cite{Yu_2020_CVPR}, and semantic segmentation on
Cityscapes~\cite{cordts2016cityscapes} in Fig.~\ref{fig:vis_results}. For
\ourdataset{} and BDD100K, models are trained on the corresponding dataset with
heavier readout headers. For Cityscapes, we use a \ourdataset{} pretrained
model with the standard readout header, which only has a simple linear
classification layer on all pixels. Our model can produce impressive results
for these segmentation tasks.

\subsection{Limitations}
We observe that when using heavier readout headers instead of standard readout
headers, the models trained on ImageNet are able to catch up with our model. We
notice that in these cases our method performs much worse on instance classes
like {\it rider} and {\it motorcycle}, which are usually rare in the datasets.
This might caused by the data imbalance when using our pixel-based objective,
whereas ImageNet has a balanced distribution across semantic classes. Solely
relying on equivariance objective and lack of invariance objectives may also
sacrifice some higher level representations, since when using heavier readout
headers, our method does not improve as much as ImageNet pretrained models do.

%% file: section/conclusion.tex
\vspace{-0.1in}
\section{Conclusion}
In this paper, we present a new self-supervised representation learning
framework based on a flow equivariance objective. Our method is able to learn
pixel-level representations from raw high-resolution videos with complex
scenes. Large scale experiments on driving videos suggest that our unsupervised
representations are useful for object detection, semantic and instance
segmentation, and in many cases outperform state-of-the-art representations
obtained from ImageNet.

%% file: section/supplementary.tex
\subsection{Additional ablation experiments}

\subsubsection{Noisier guidance from flow}

To test our framework's generalizability and robustness, here we experiment with a weaker
flow model, which produces worse flow prediction. Specifically, we use PWC-Net~\cite{Sun2018PWC-Net}
which is pretrained on Flying-chair, Flying-things, and Sintel. Following ablation experiments in the
main paper, we train our model on UrbanCity with 16,000 iterations. The results are shown in
Table~\ref{tab:exp_flow}. We can see that using a weaker flow model in our framework only has a
minor impact on the model performance.

\begin{table}[h]
	\centering

	\resizebox{\linewidth}{!}{
		\begin{tabular}{c|cc|cc}
			\toprule
			Flow model        & mIoU          & mAP          & mIoU$^\dag$   & mAP$^\dag$    \\
			\midrule
			PWC-Net           & 37.2          & \textbf{3.9} & 52.4          & 16.4          \\
			RAFT (main paper) & \textbf{37.9} & 3.8          & \textbf{53.2} & \textbf{16.5} \\
			\bottomrule
		\end{tabular}
	}
	\caption{Semantic segmentation and instance segmentation readout results on \ourdataset{} with a
		weaker flow model PWC-Net~\cite{Sun2018PWC-Net}}
	\label{tab:exp_flow}
\end{table}

\subsubsection{Training BYOL on driving videos:}
BYOL~\cite{grill2020bootstrap} shows good performance on ImageNet data which is highly curated and carefully constructed for image-level recognition, and its training protocol is proven to work very well on ImageNet.
It treats the entire image as a single instance and random crops two patches on the images and minimizes the two patches' dissimilarity.
However, images in the wild may be more complex; in this case, the two random crops may not cover the same objects, leading to potential performance degradation.

For reference, we adapt BYOL on the driving video data.
The results are shown in Table~\ref{tab:exp_byol_results}.
For the simplest variant of BYOL, we simply treat video frames like ImageNet images. The random crop
augmentation has no spatial constraint; thus, two crops from the same image may cover different objects on the street.
We can see that the performance is as bad as a randomly initialized encoder, indicating no meaningful
representation is learned in the model.

We then try to slightly modify it by pre-cropping a 480$\times$480 patch on the original image to limit the
movement of the random crop, denoted as ``BYOL (pre-crop)''.
However, the performance is not good either.
We also try to change the pre-cropping size, but it does not help.
We think one reason might be the higher similarity and less diversity of street scene images compared to ImageNet images:
even though the smaller pre-cropped image can reduce the complexity of the image and help the two patches cover the same object,
the street scene images are very similar, and patches from different images may share the same semantic meaning
(e.g., building, road, sky), making the model hard to distinguish them and learn meaningful representations.

Furthermore, we also try BYOL using two neighboring video frames (BYOL (pre-crop) w/ video),
but it is even worse due to extra object movement across frames.
These results indicate that the popular BYOL training protocol on ImageNet is not ideal for raw driving videos.
Instead, our method can utilize the raw driving video and learn meaningful representations effectively.
\begin{table}
	\centering
	\resizebox{\linewidth}{!}{
		\begin{tabular}{l|cc|cc}
			\toprule
			\multirow{2}{*}{Method}        & \multicolumn{2}{c|}{\ourdataset{}} & \multicolumn{2}{c}{BDD100K}                            \\
			~                              & mIoU$^\dag$                        & mAP$^\dag$                  & mIoU$^\dag$ & mAP$^\dag$ \\
			\midrule
			BYOL~\cite{grill2020bootstrap} & 19.6                               & 5.0                         & 21.9        & 4.5        \\
			BYOL (pre-crop)                & 26.6                               & 5.1                         & 18.2        & 4.2        \\
			BYOL (pre-crop) w/ video       & 13.0                               & 2.0                         & 10.7        & 2.1        \\
			\ourmethod{} (Ours)            & {\bf 61.7}                         & {\bf 19.0}                  & {\bf 49.8}  & {\bf 24.9} \\
			\bottomrule
		\end{tabular}
	}
	\caption{BYOL trained on \ourdataset{} and BDD100K.}
	\label{tab:exp_byol_results}
\end{table}

\begin{figure*}
	\centering
	\includegraphics[width=0.342\linewidth]{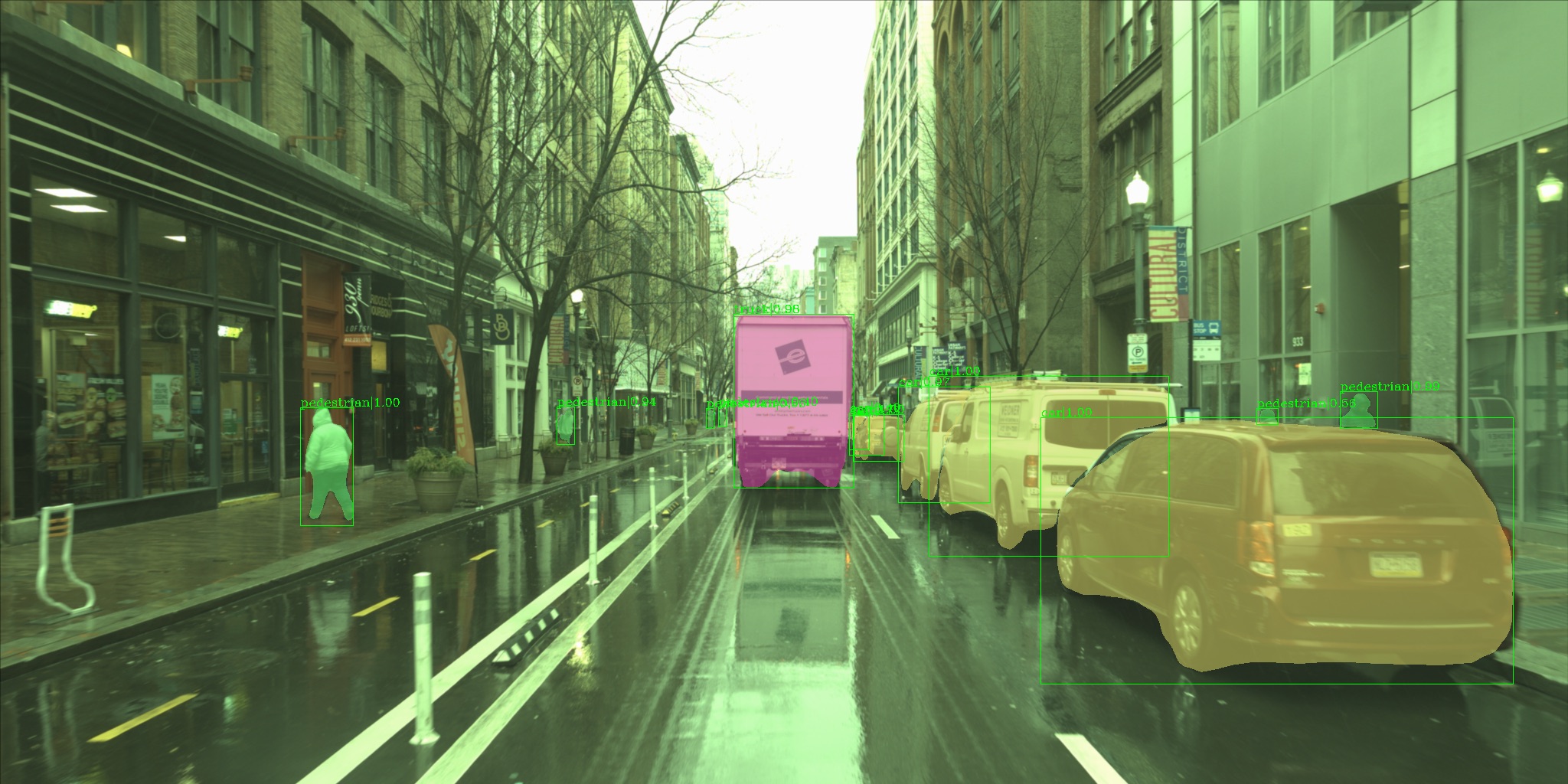}
	\hfill
	\includegraphics[width=0.304\linewidth]{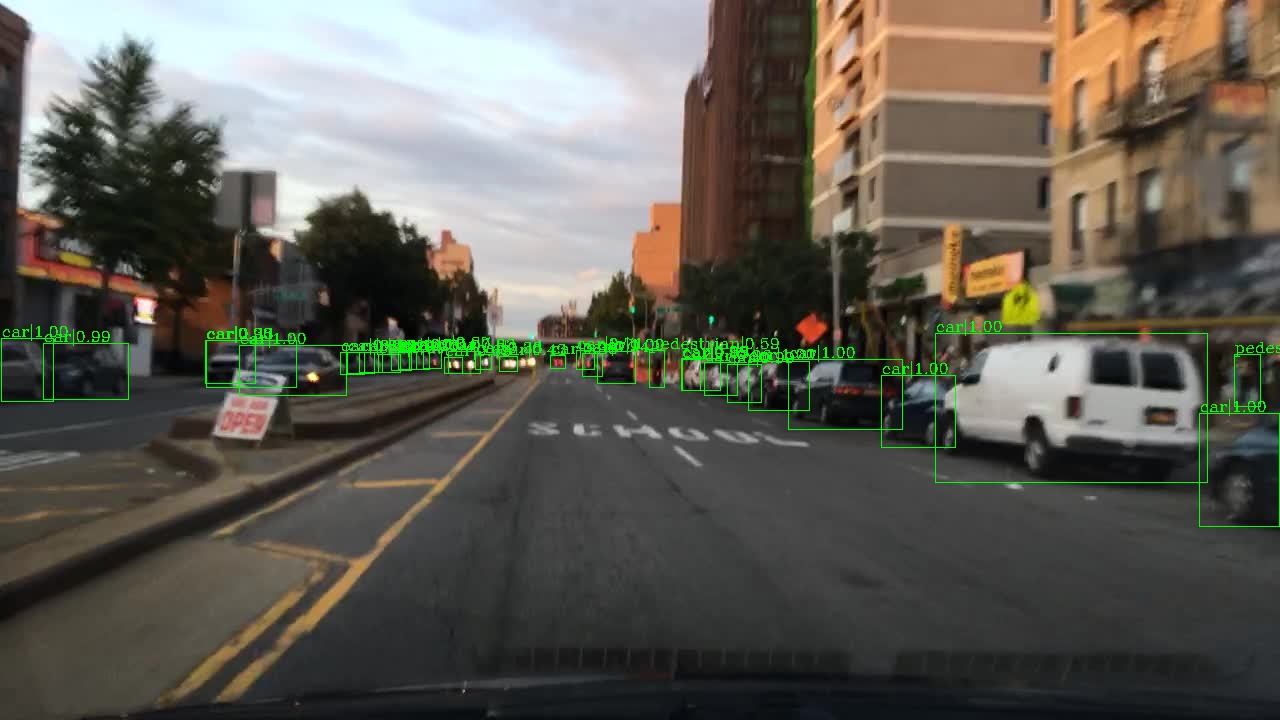}
	\hfill
	\includegraphics[width=0.342\linewidth]{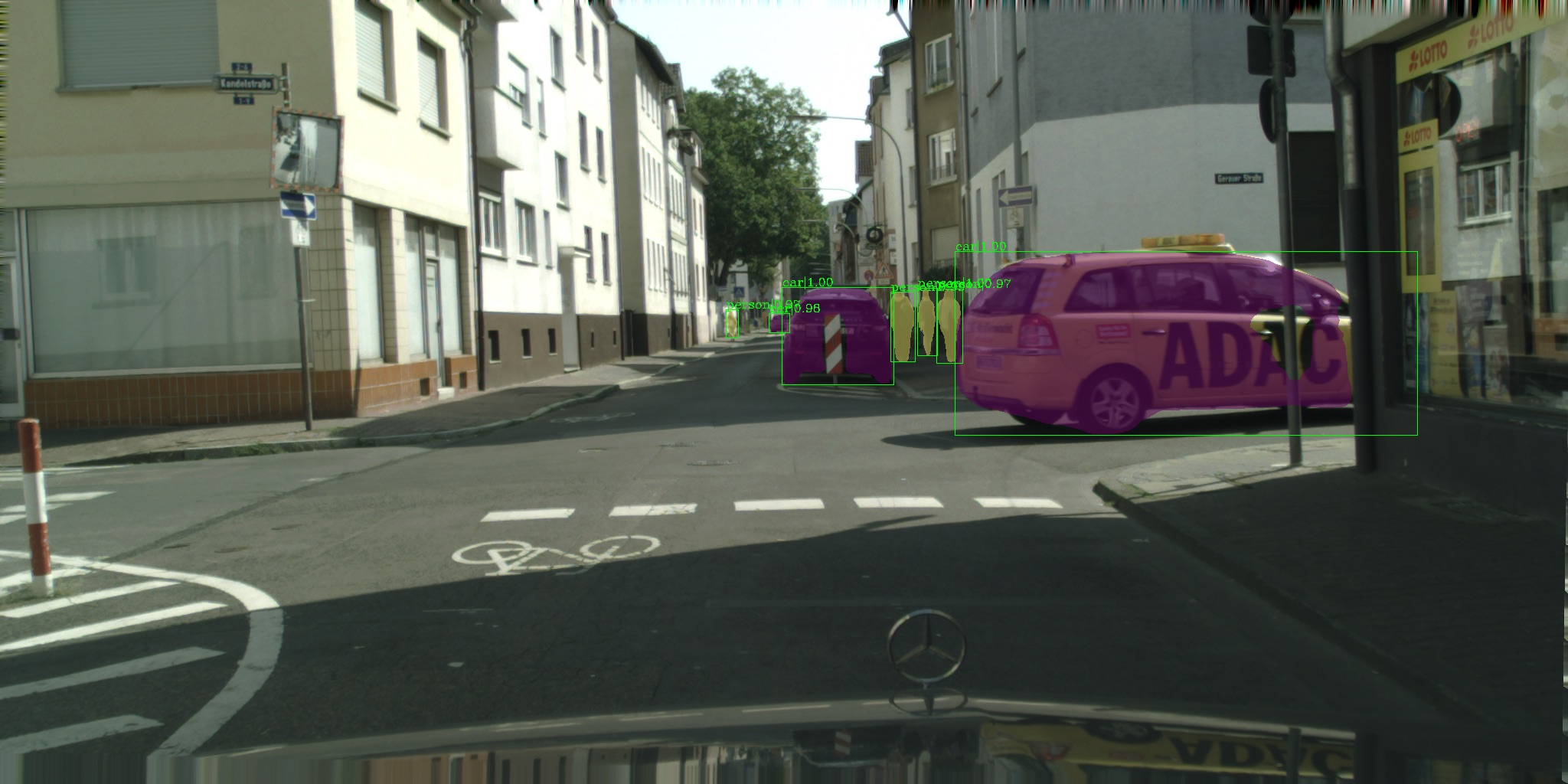}
	\hfill
	\includegraphics[width=0.342\linewidth]{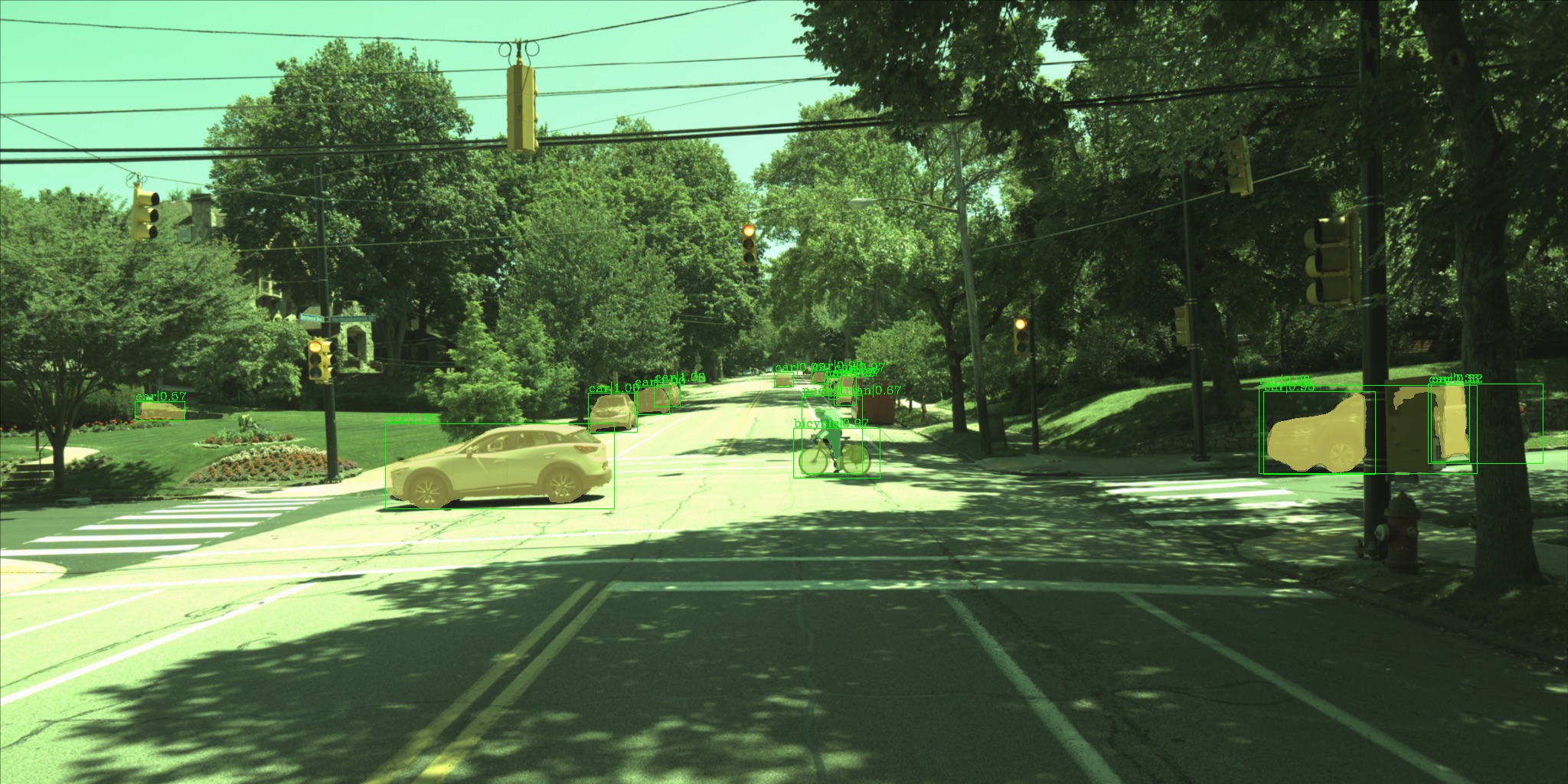}
	\hfill
	\includegraphics[width=0.304\linewidth]{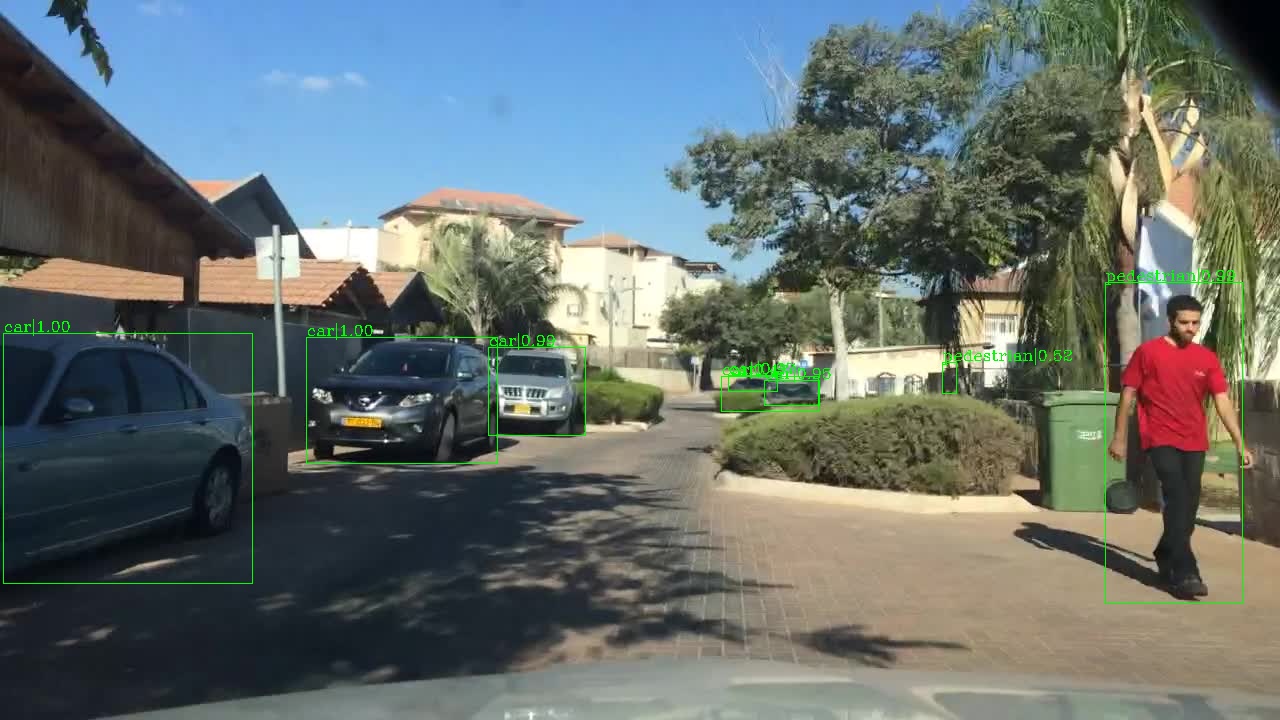}
	\hfill
	\includegraphics[width=0.342\linewidth]{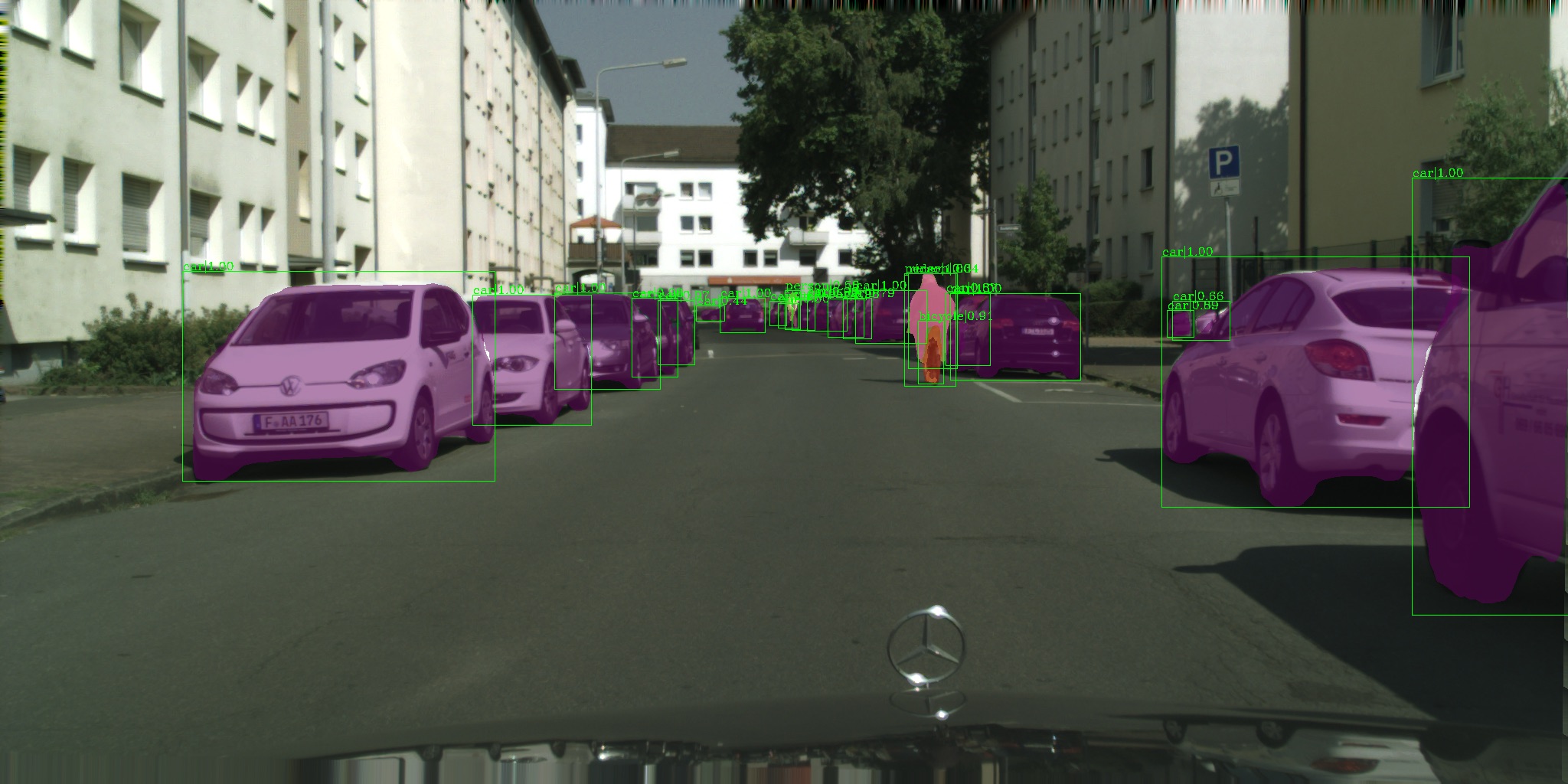}
	\hfill
	\includegraphics[width=0.342\linewidth]{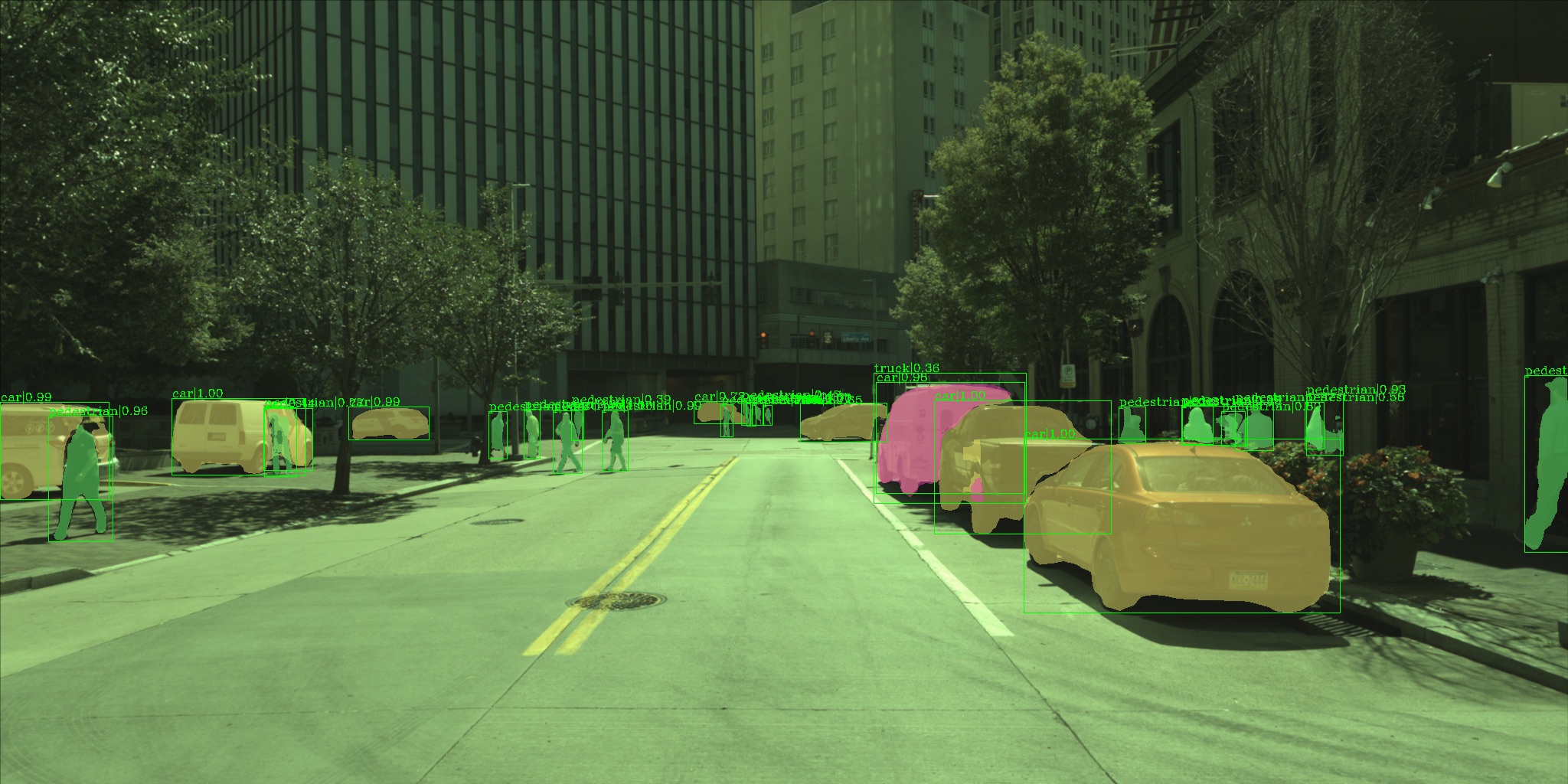}
	\hfill
	\includegraphics[width=0.304\linewidth]{figure/vis_results_supp/bdd_det/b1e3e9f5-92377424.jpg}
	\hfill
	\includegraphics[width=0.342\linewidth]{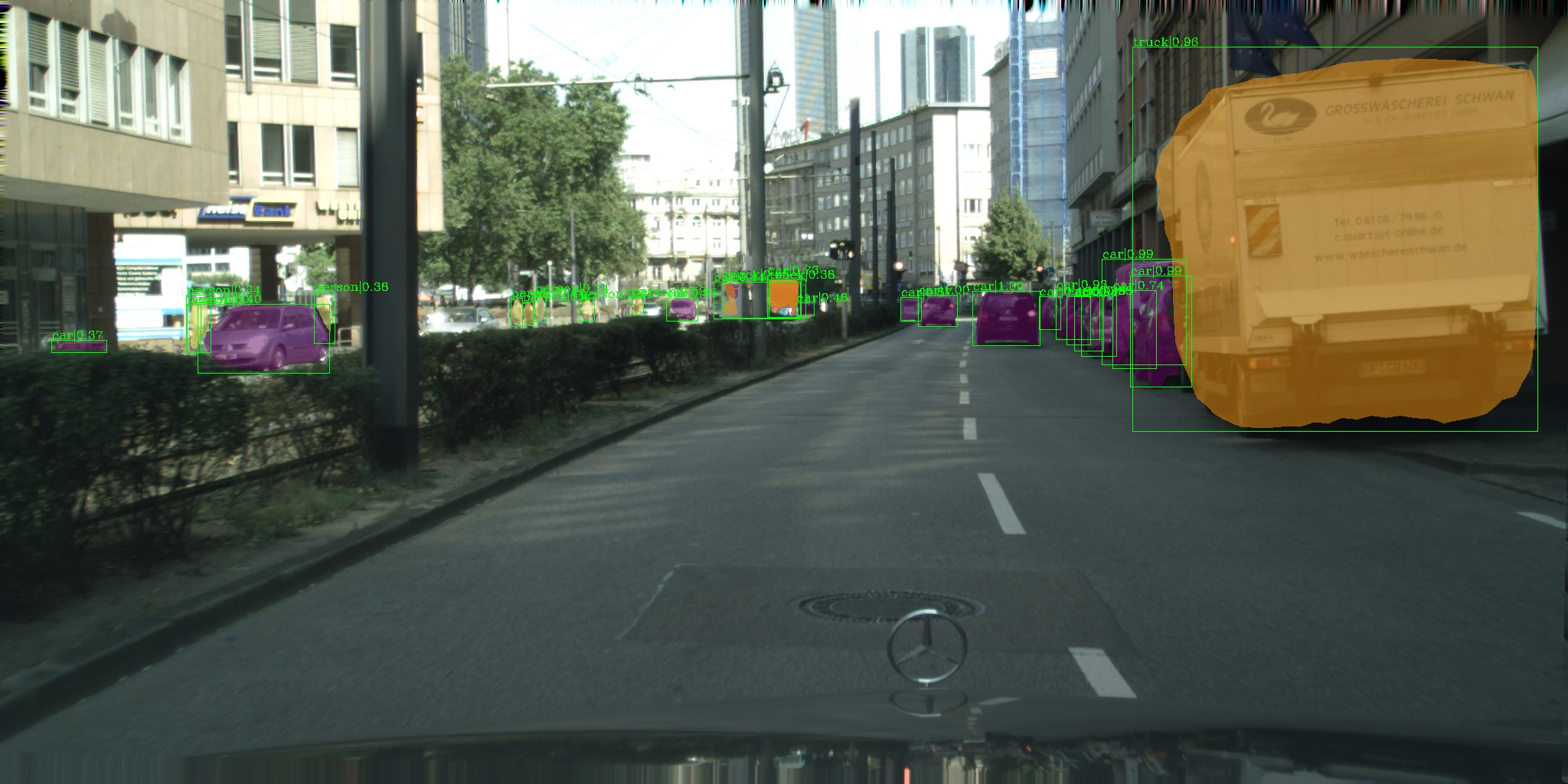}
	\hfill
	\includegraphics[width=0.342\linewidth]{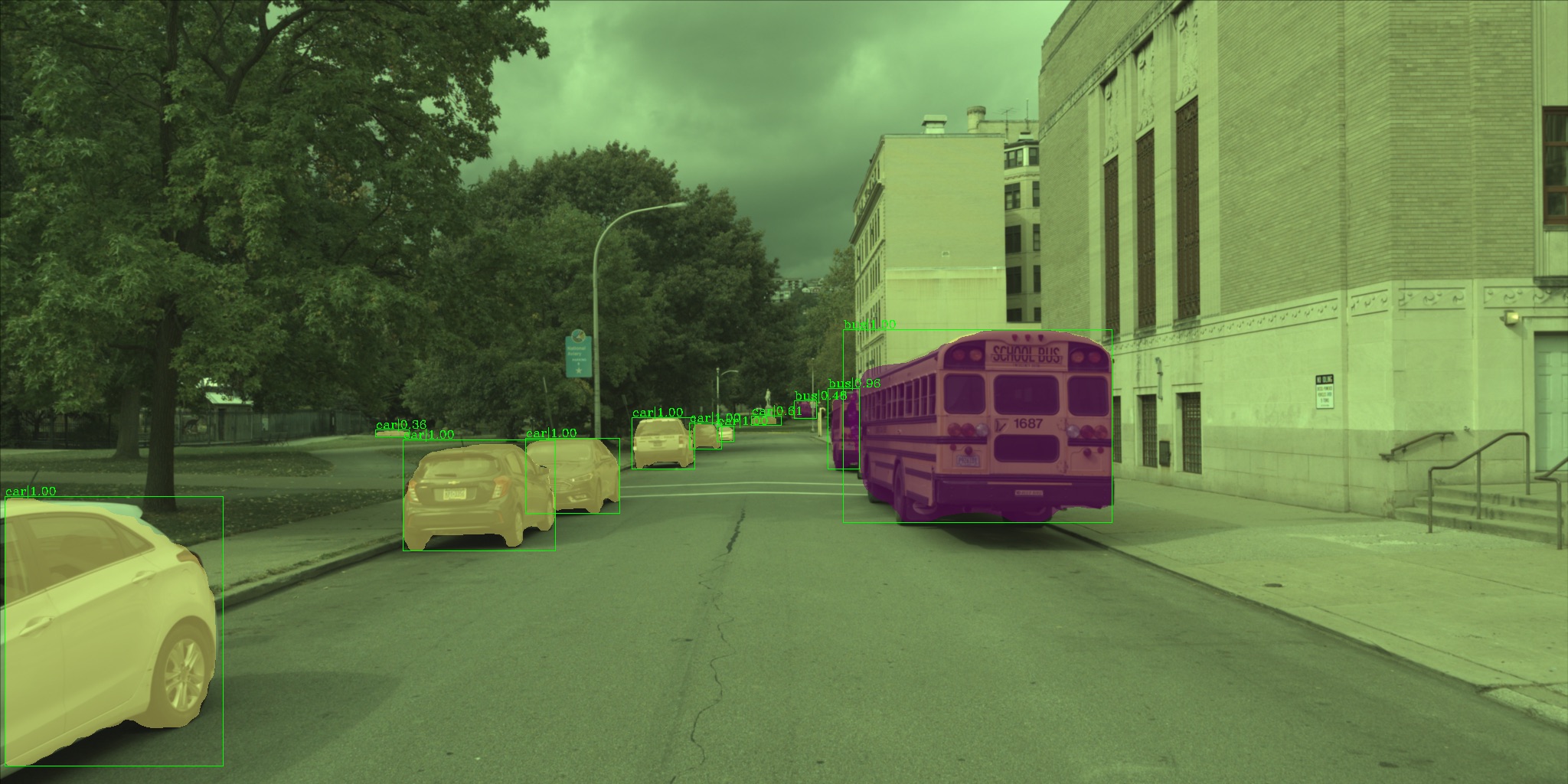}
	\hfill
	\includegraphics[width=0.304\linewidth]{figure/vis_results_supp/bdd_det/b2dbb793-9f0b2bec.jpg}
	\hfill
	\includegraphics[width=0.342\linewidth]{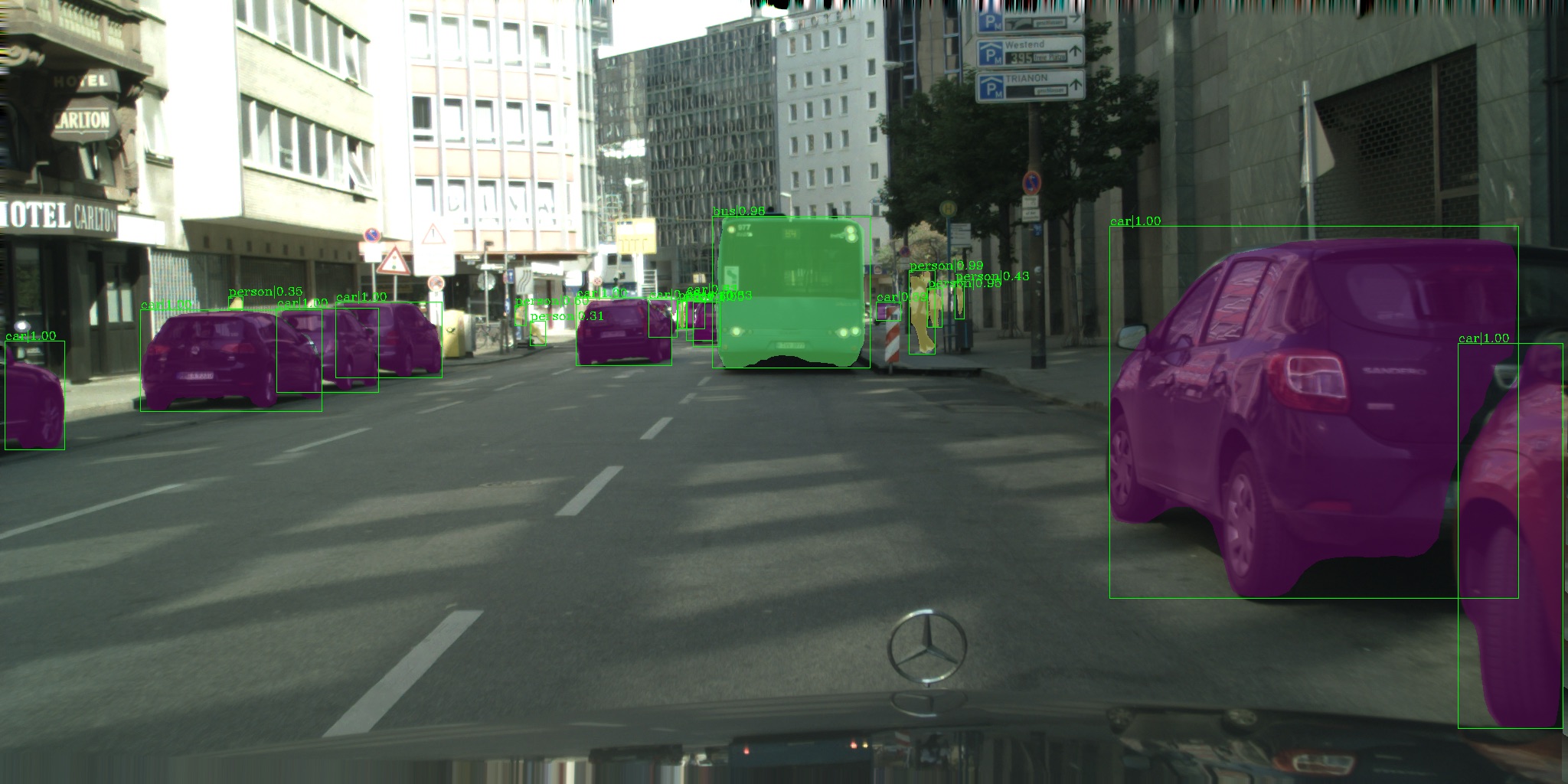}
	\includegraphics[width=0.342\linewidth]{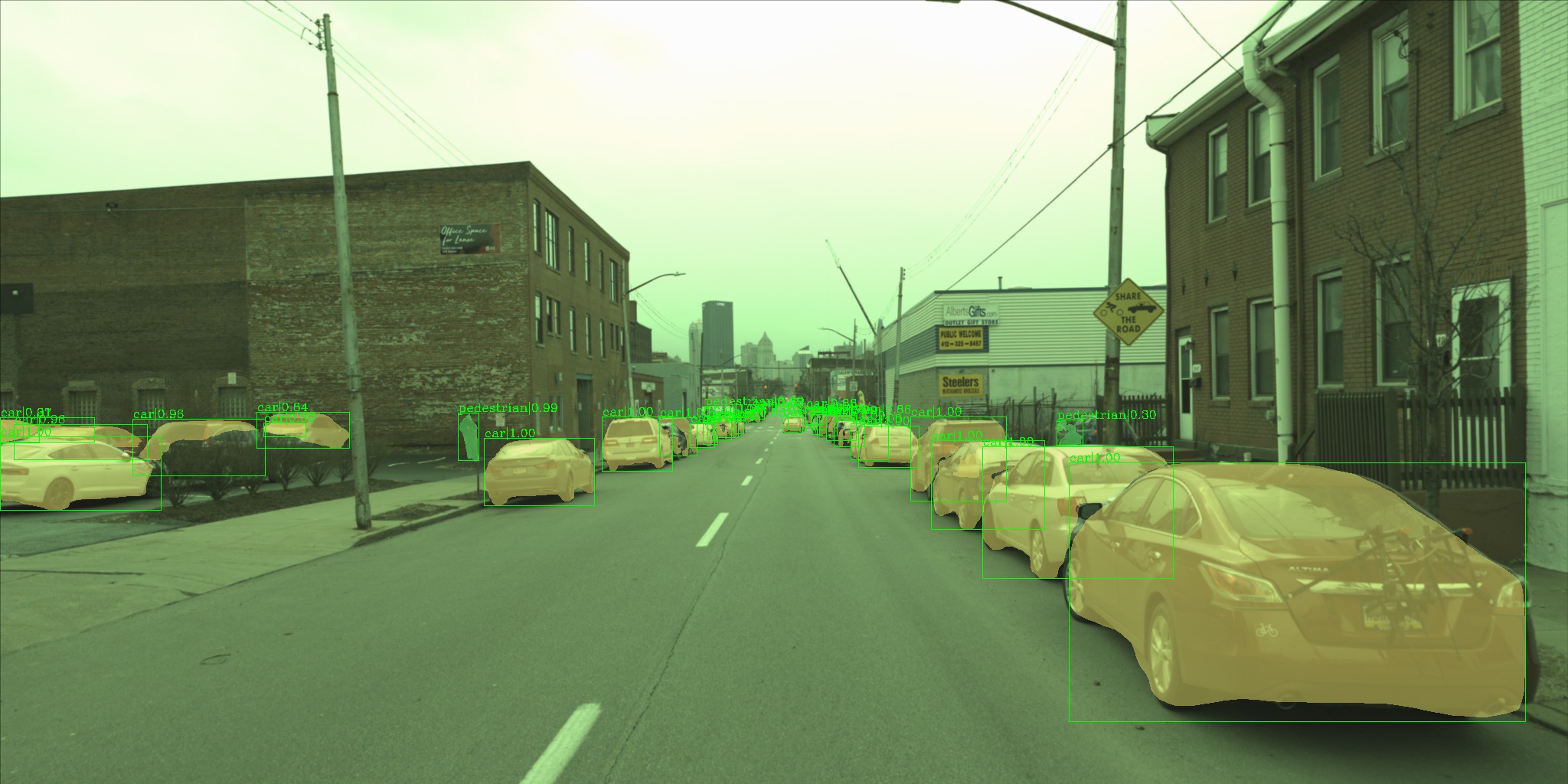}
	\hfill
	\includegraphics[width=0.304\linewidth]{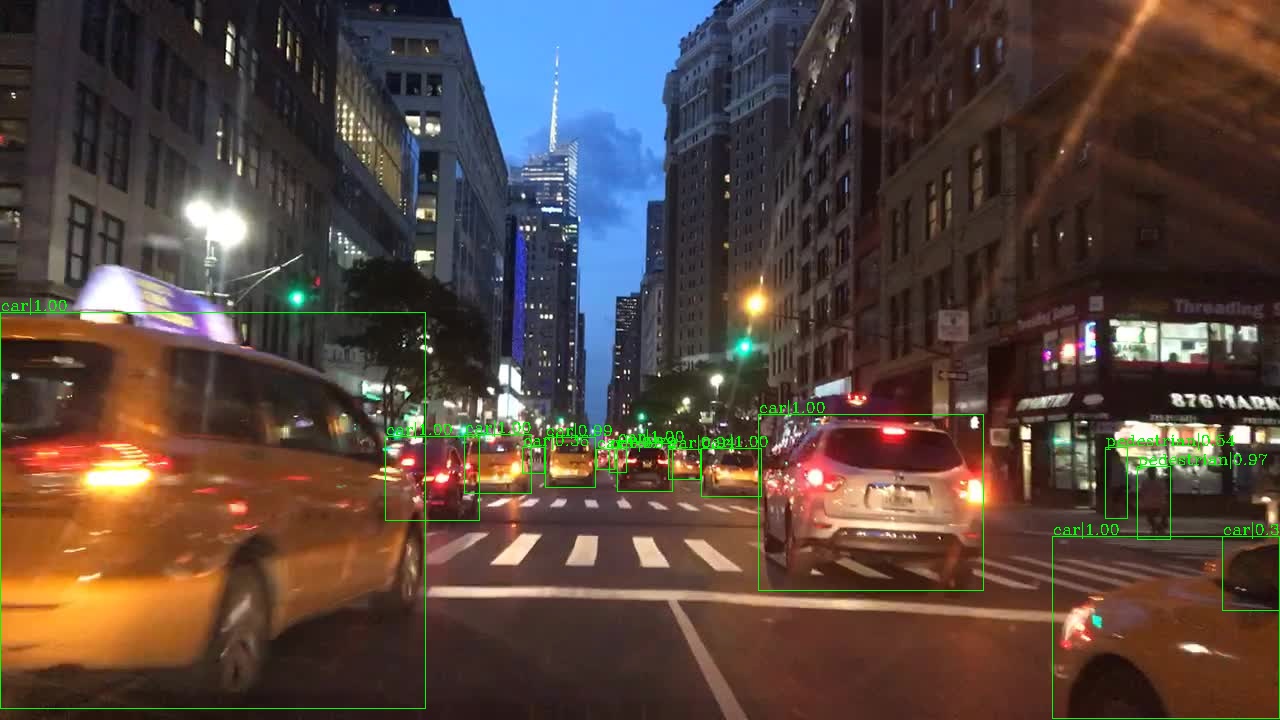}
	\hfill
	\includegraphics[width=0.342\linewidth]{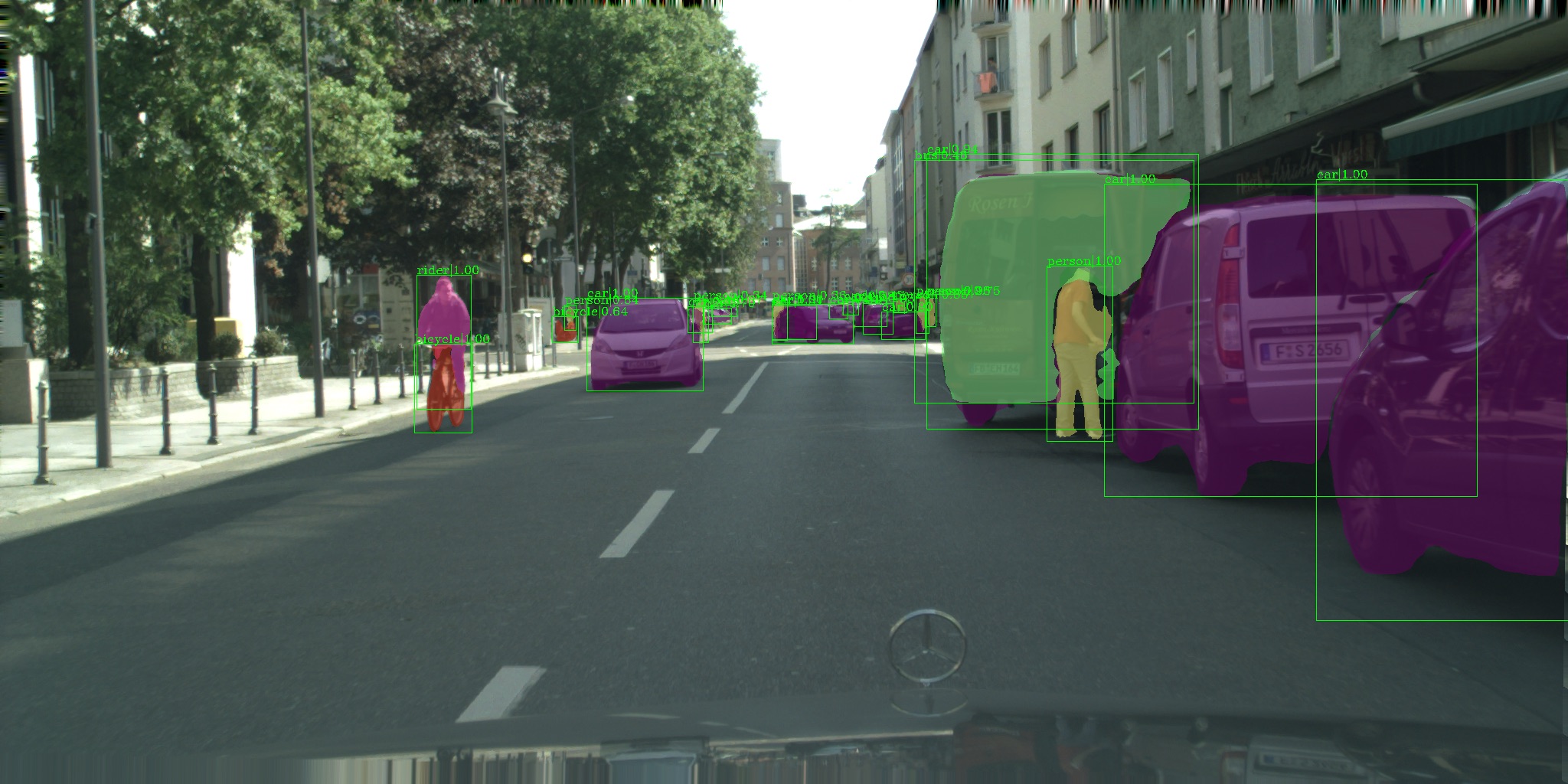}
	\includegraphics[width=0.342\linewidth]{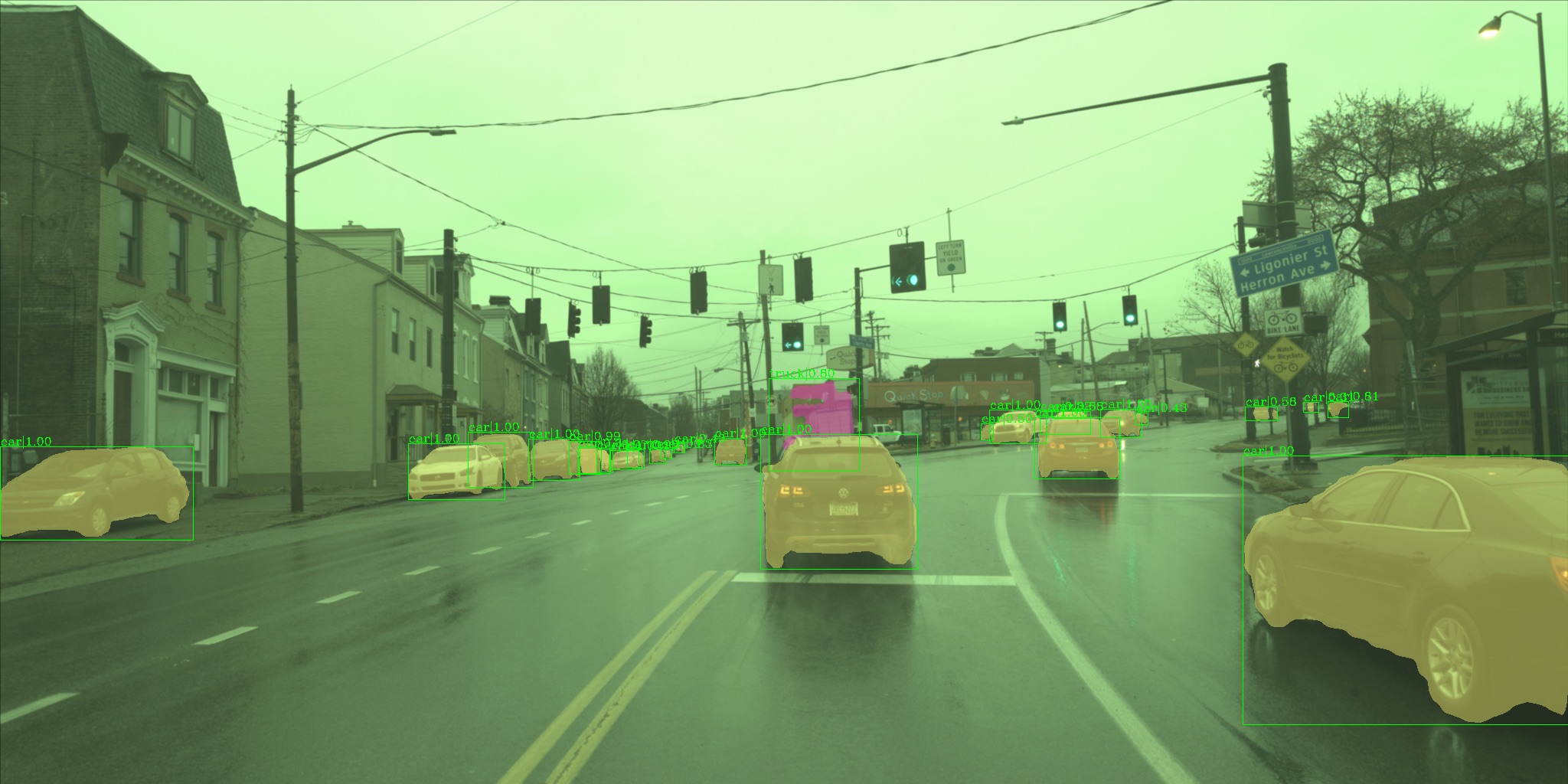}
	\hfill
	\includegraphics[width=0.304\linewidth]{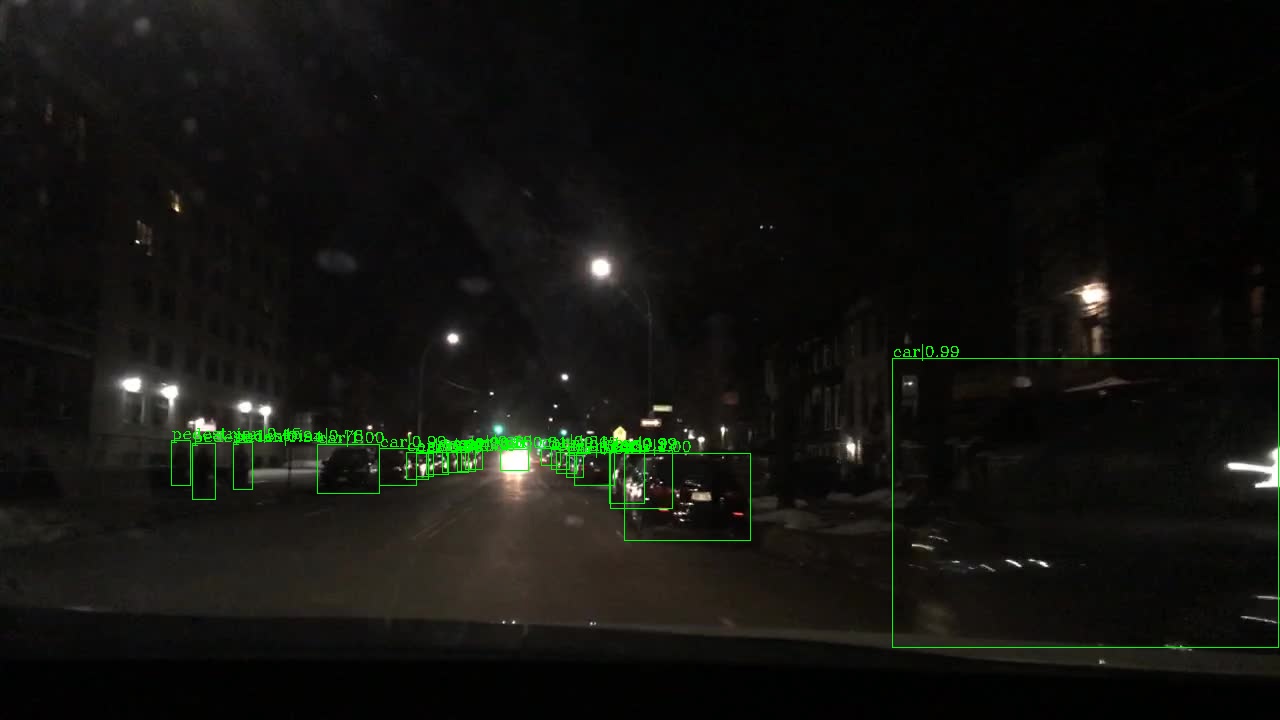}
	\hfill
	\includegraphics[width=0.342\linewidth]{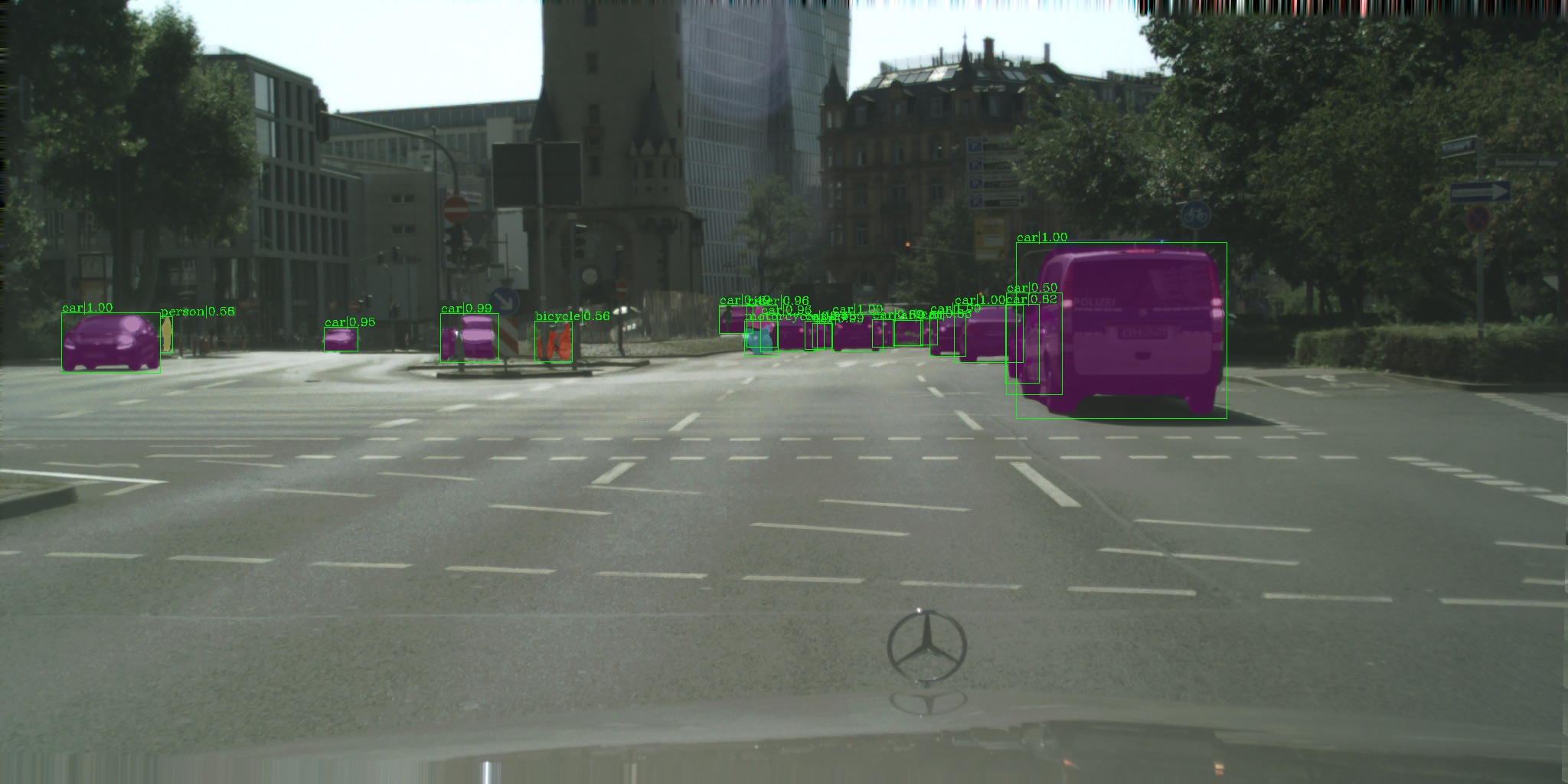}
	\caption{Instance segmentation / object detection visualization results on \ourdataset{} (left),
		BDD100K (middle), and Cityscapes (right). We use the heavier header (ResNet-FPN) for readout. }
	\label{fig:vis_results_1}
	\vspace{-0.2in}
\end{figure*}

\subsection{Additional visualization results}

\begin{figure*}
	\centering
	\includegraphics[width=0.342\linewidth]{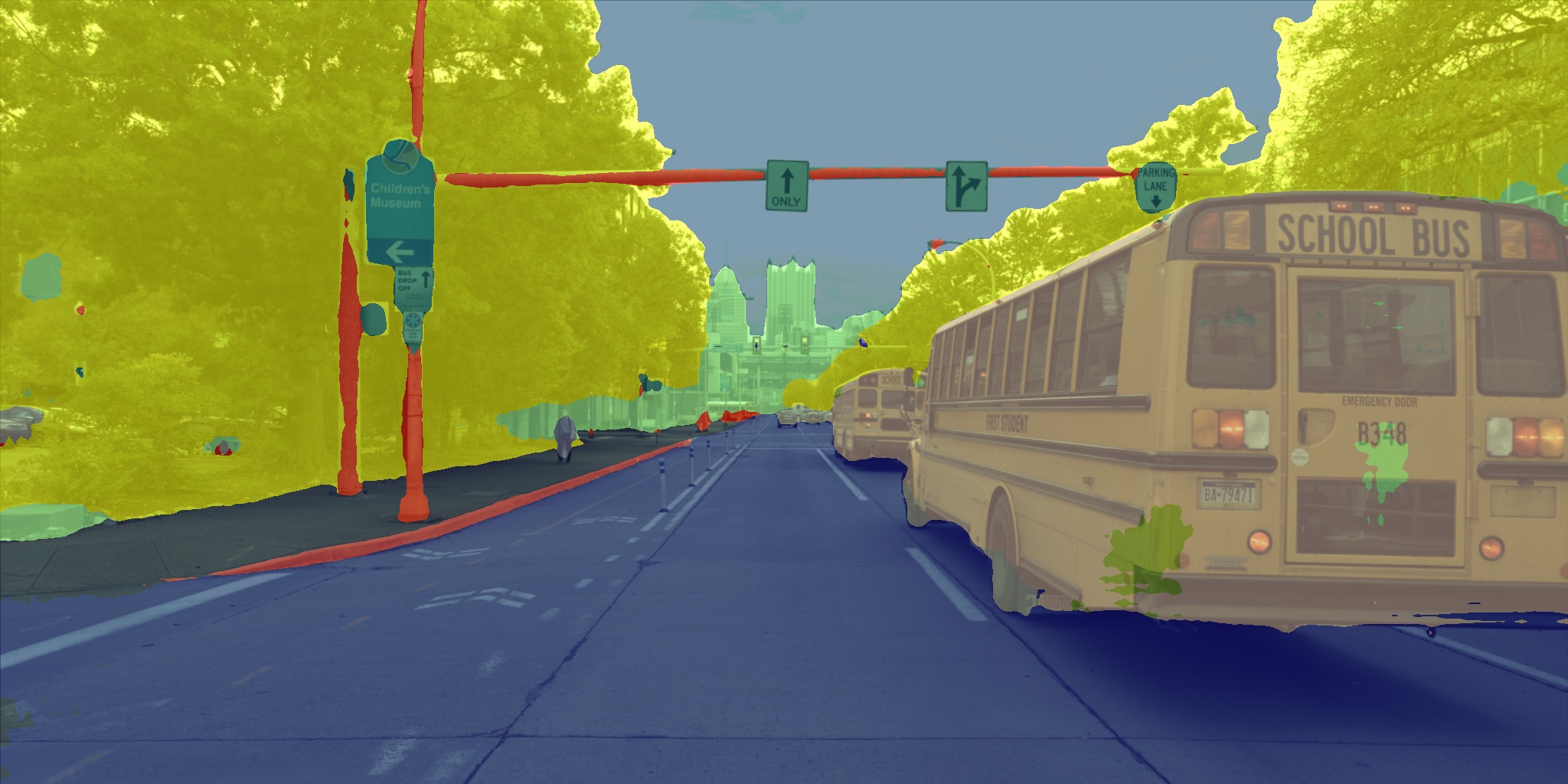}
	\hfill
	\includegraphics[width=0.304\linewidth]{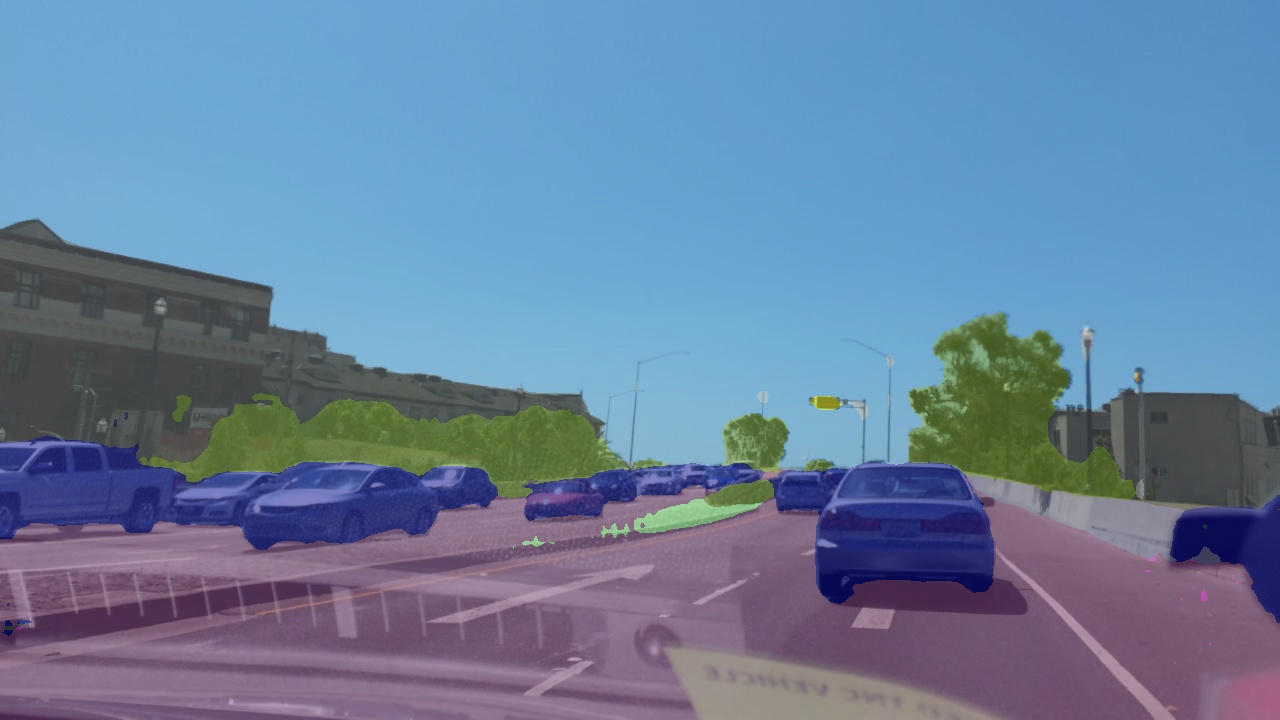}
	\hfill
	\includegraphics[width=0.342\linewidth]{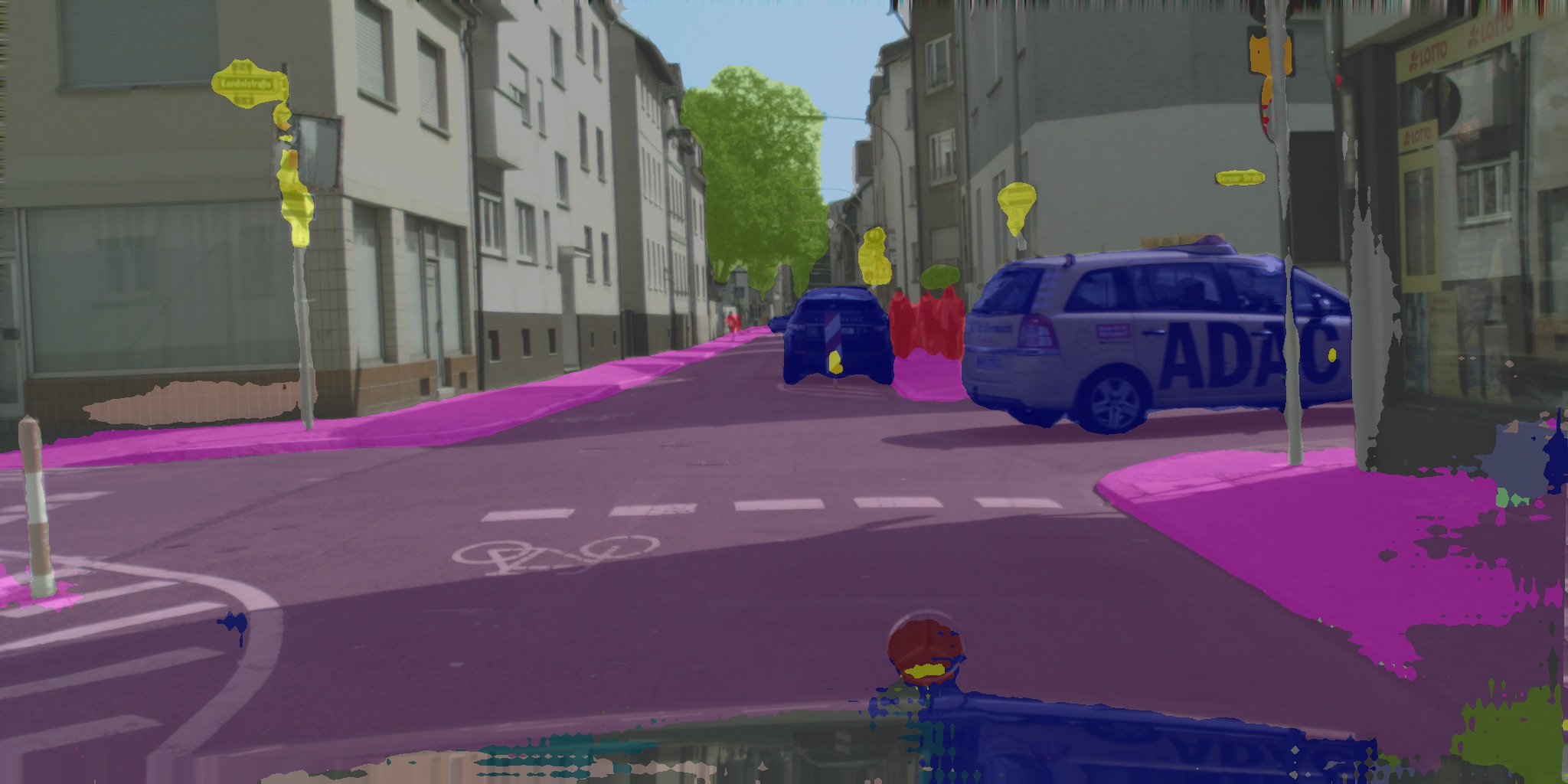}
	\hfill
	\includegraphics[width=0.342\linewidth]{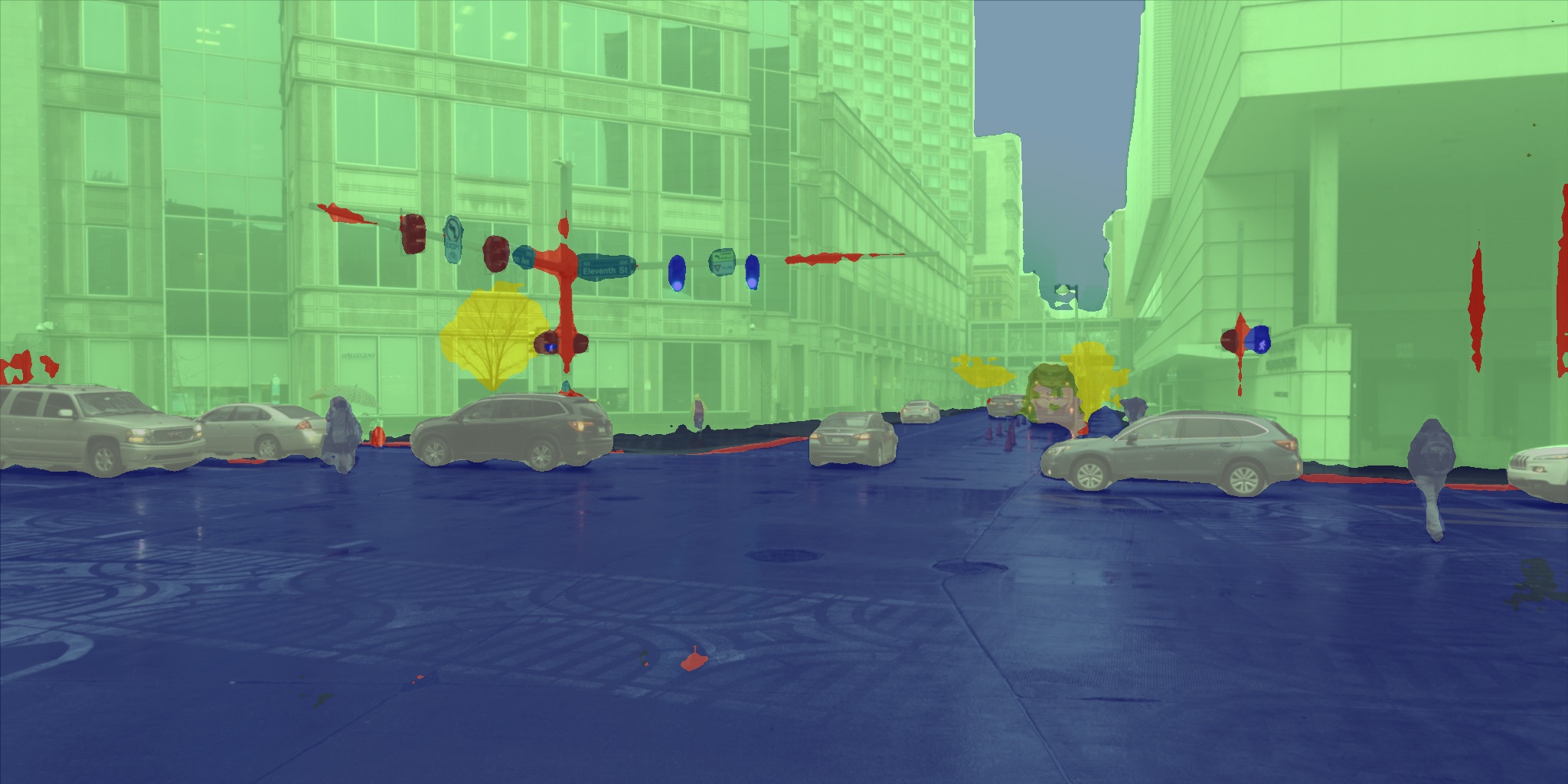}
	\hfill
	\includegraphics[width=0.304\linewidth]{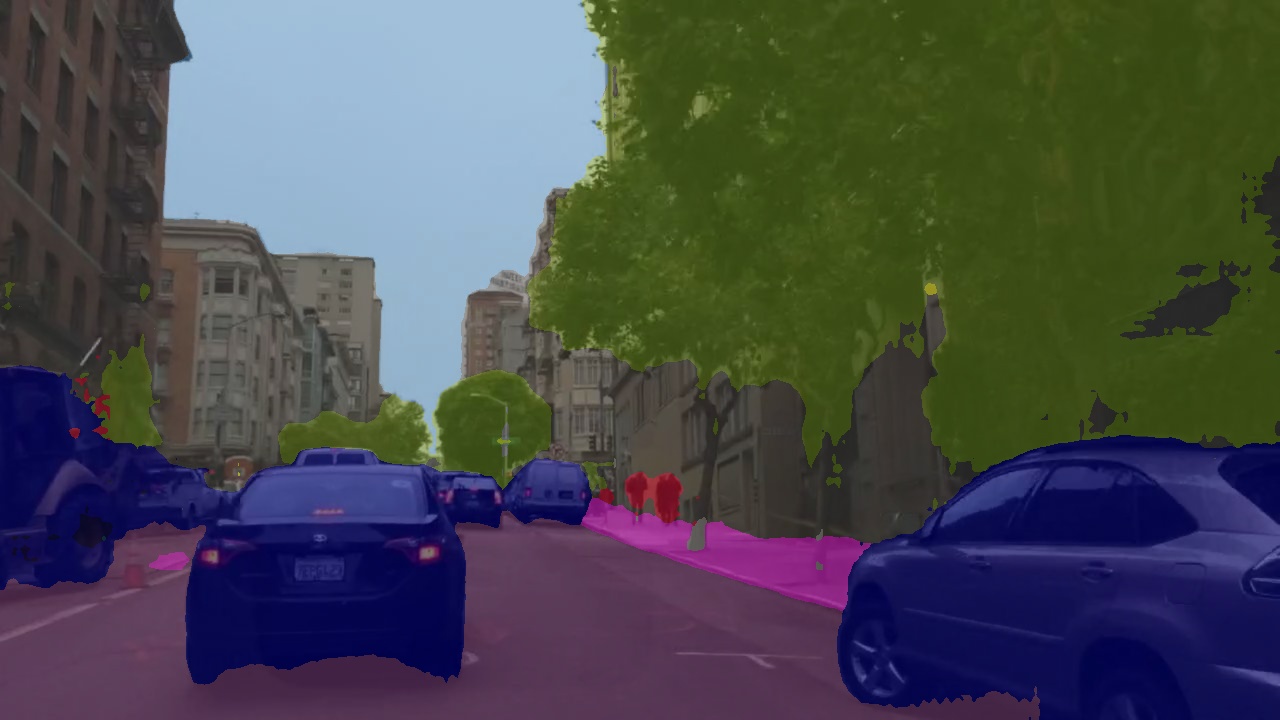}
	\hfill
	\includegraphics[width=0.342\linewidth]{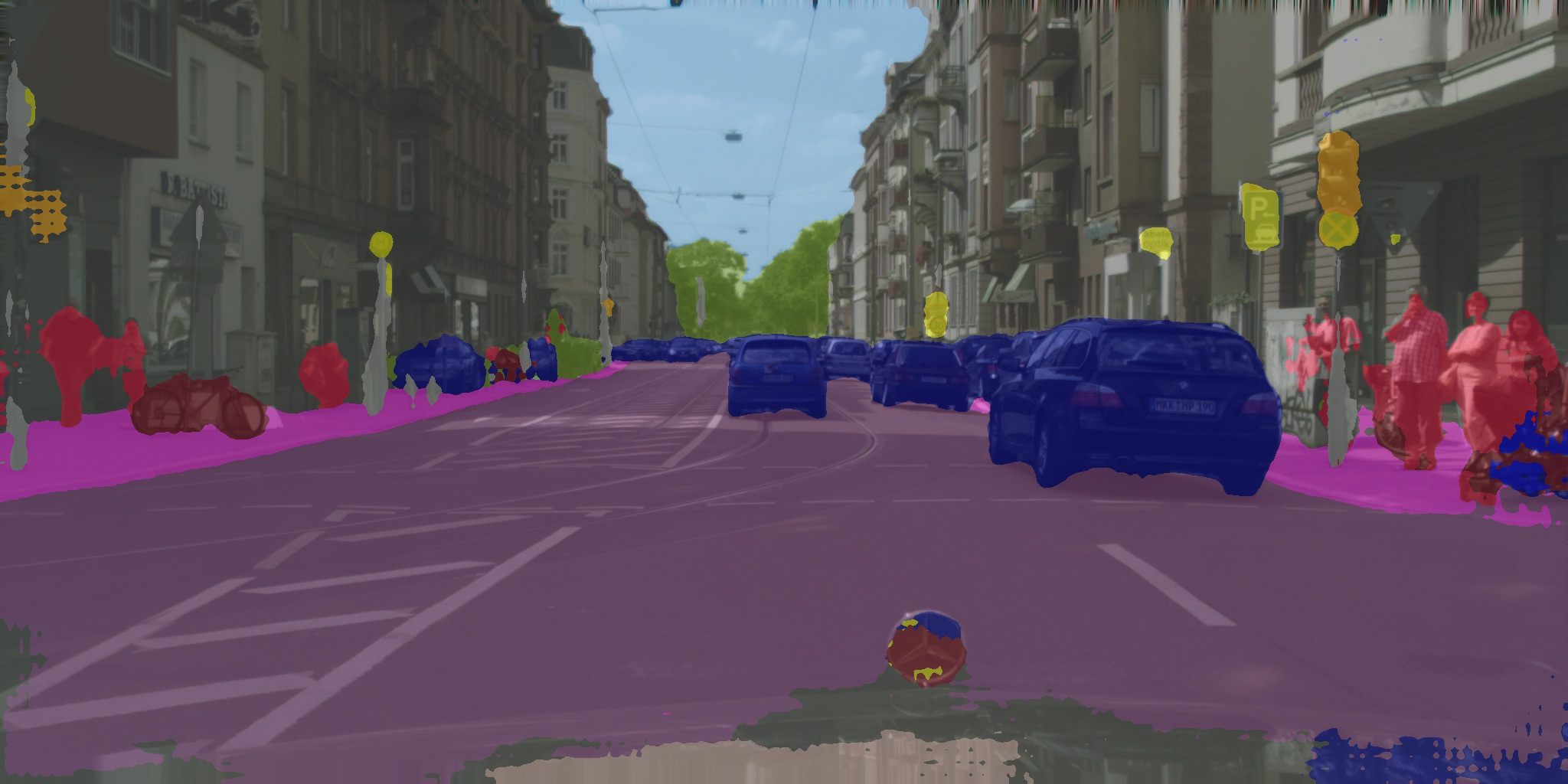}
	\hfill
	\includegraphics[width=0.342\linewidth]{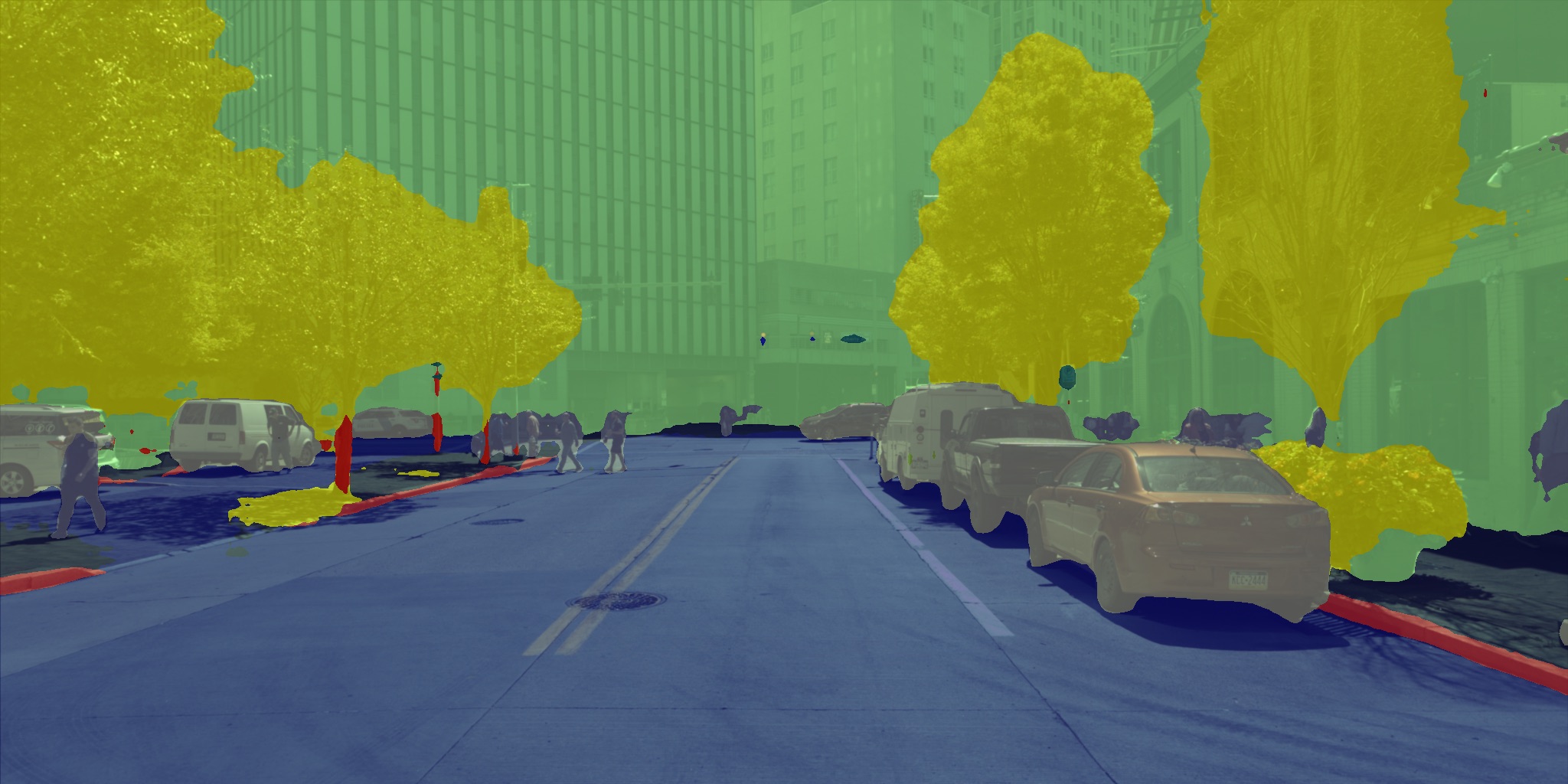}
	\hfill
	\includegraphics[width=0.304\linewidth]{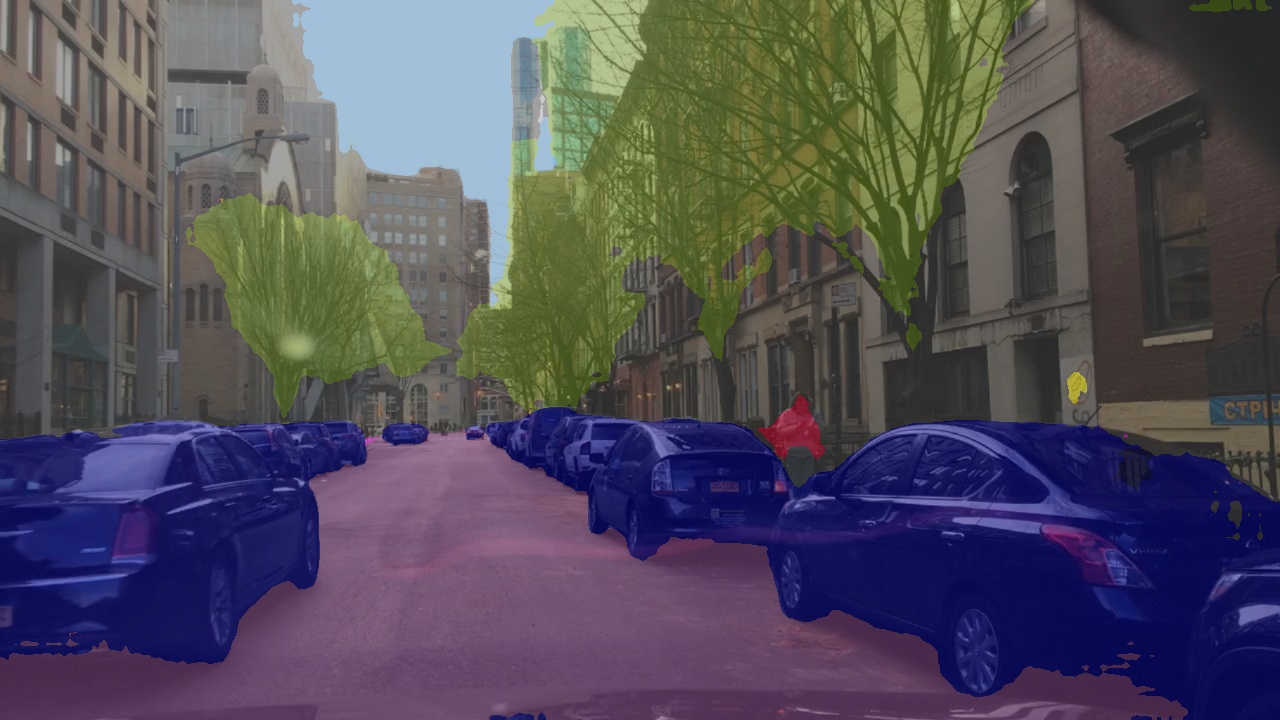}
	\hfill
	\includegraphics[width=0.342\linewidth]{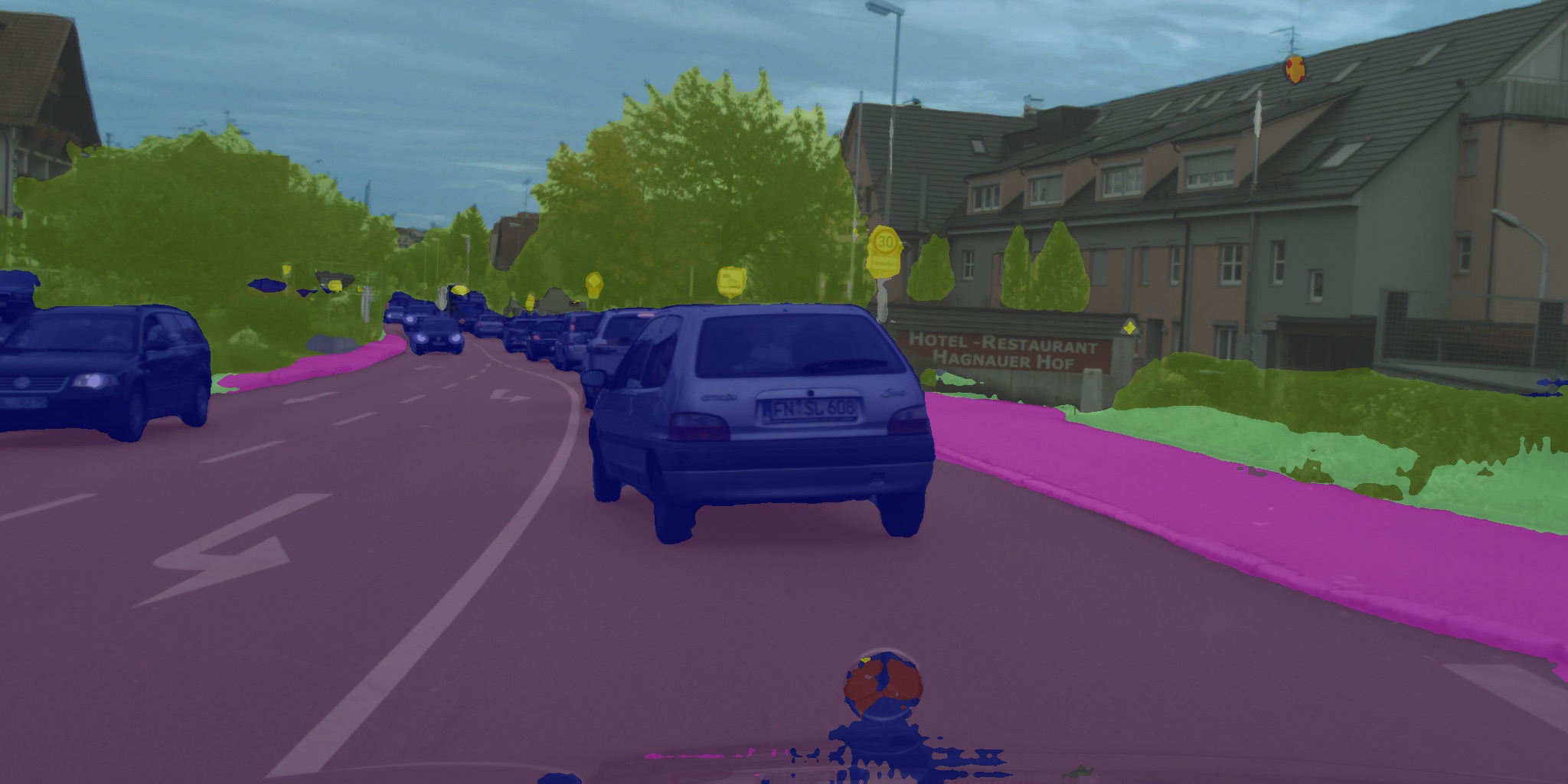}
	\hfill
	\includegraphics[width=0.342\linewidth]{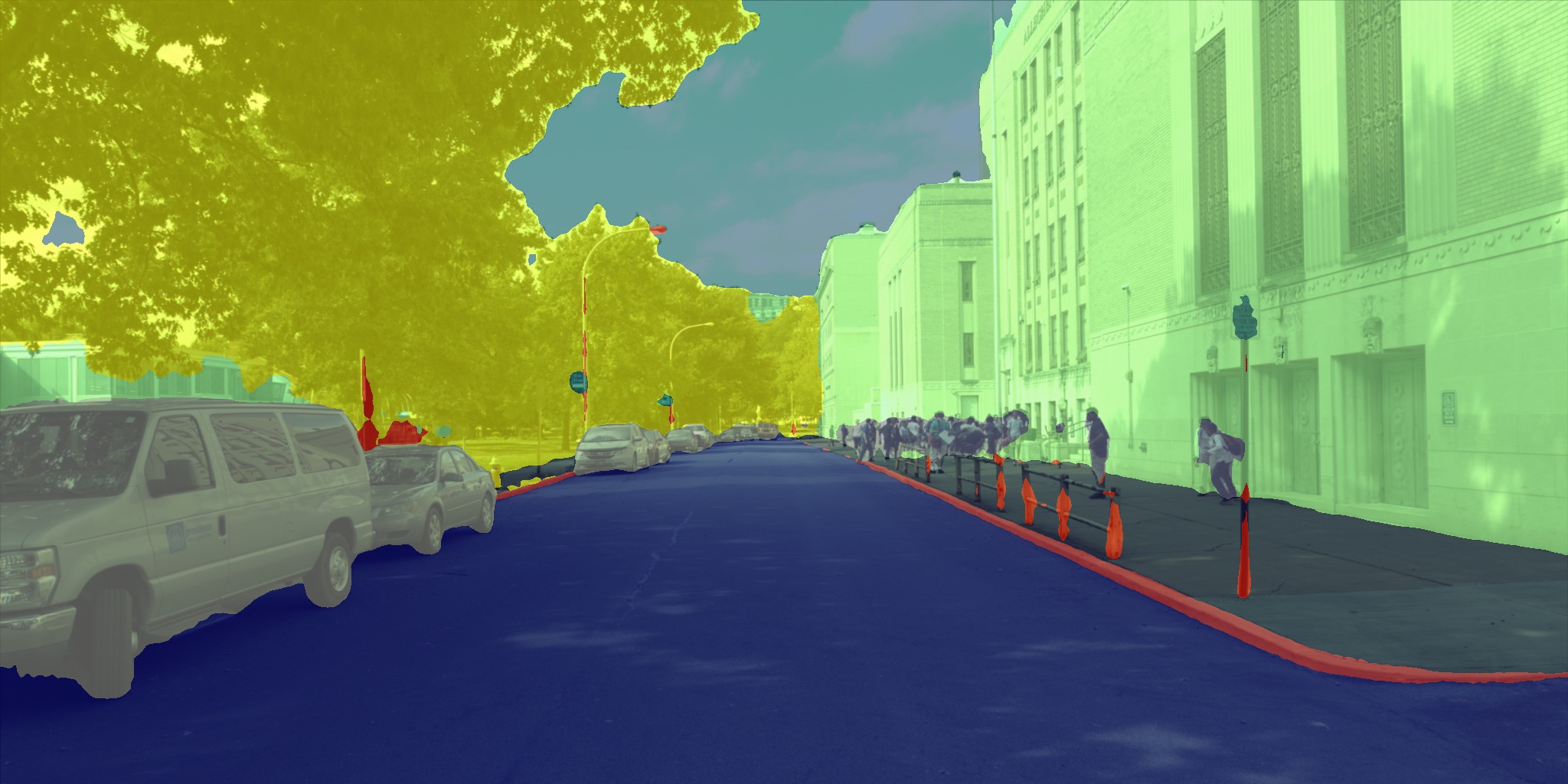}
	\hfill
	\includegraphics[width=0.304\linewidth]{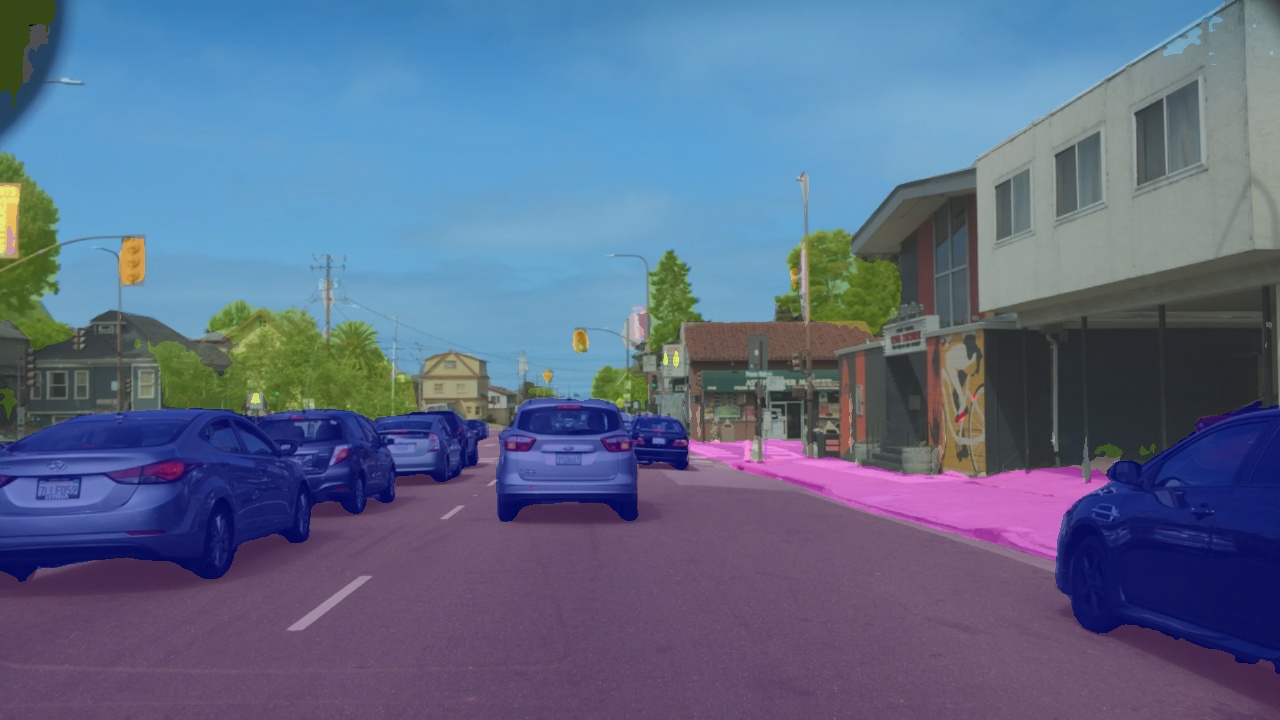}
	\hfill
	\includegraphics[width=0.342\linewidth]{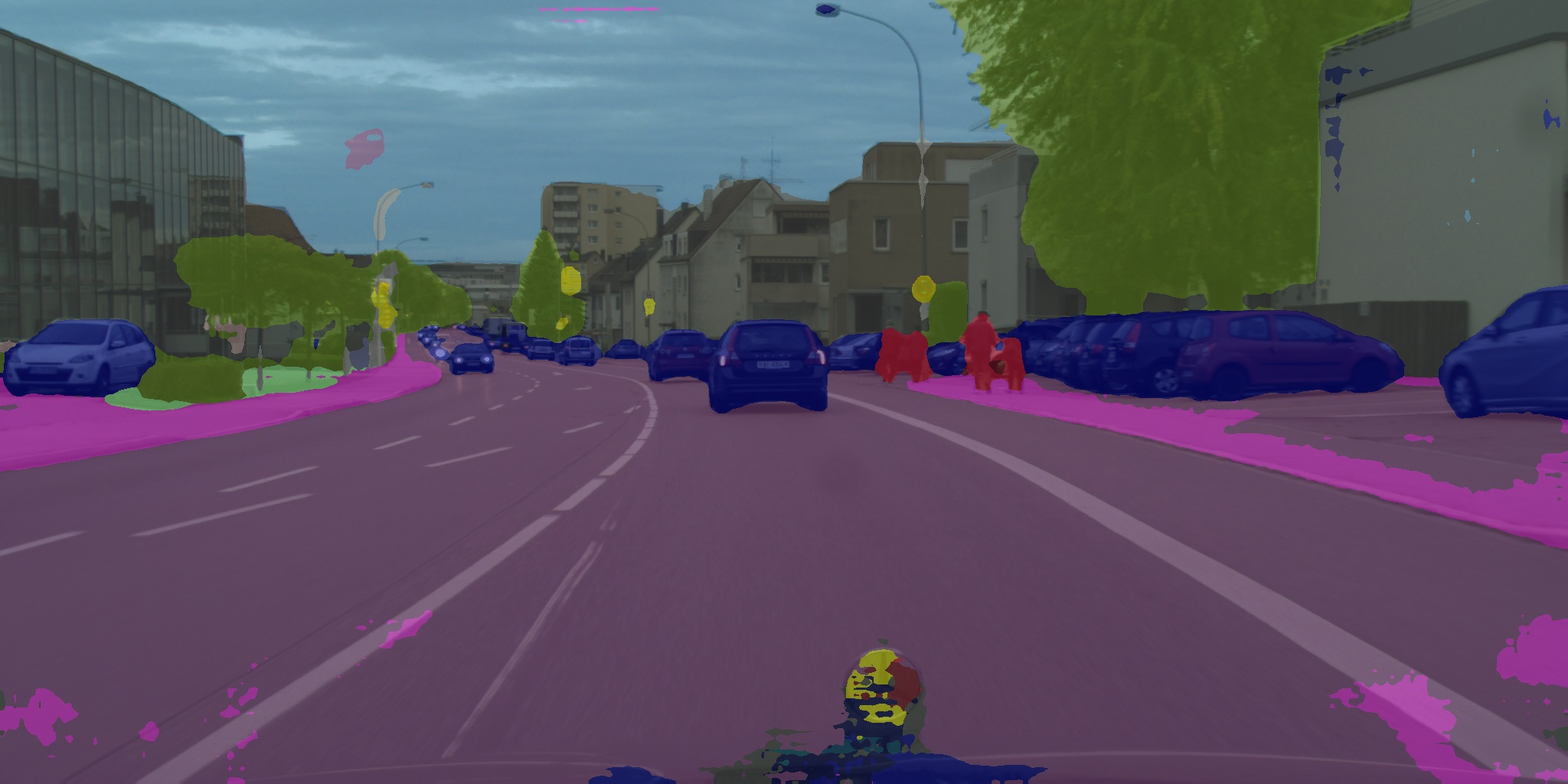}
	\includegraphics[width=0.342\linewidth]{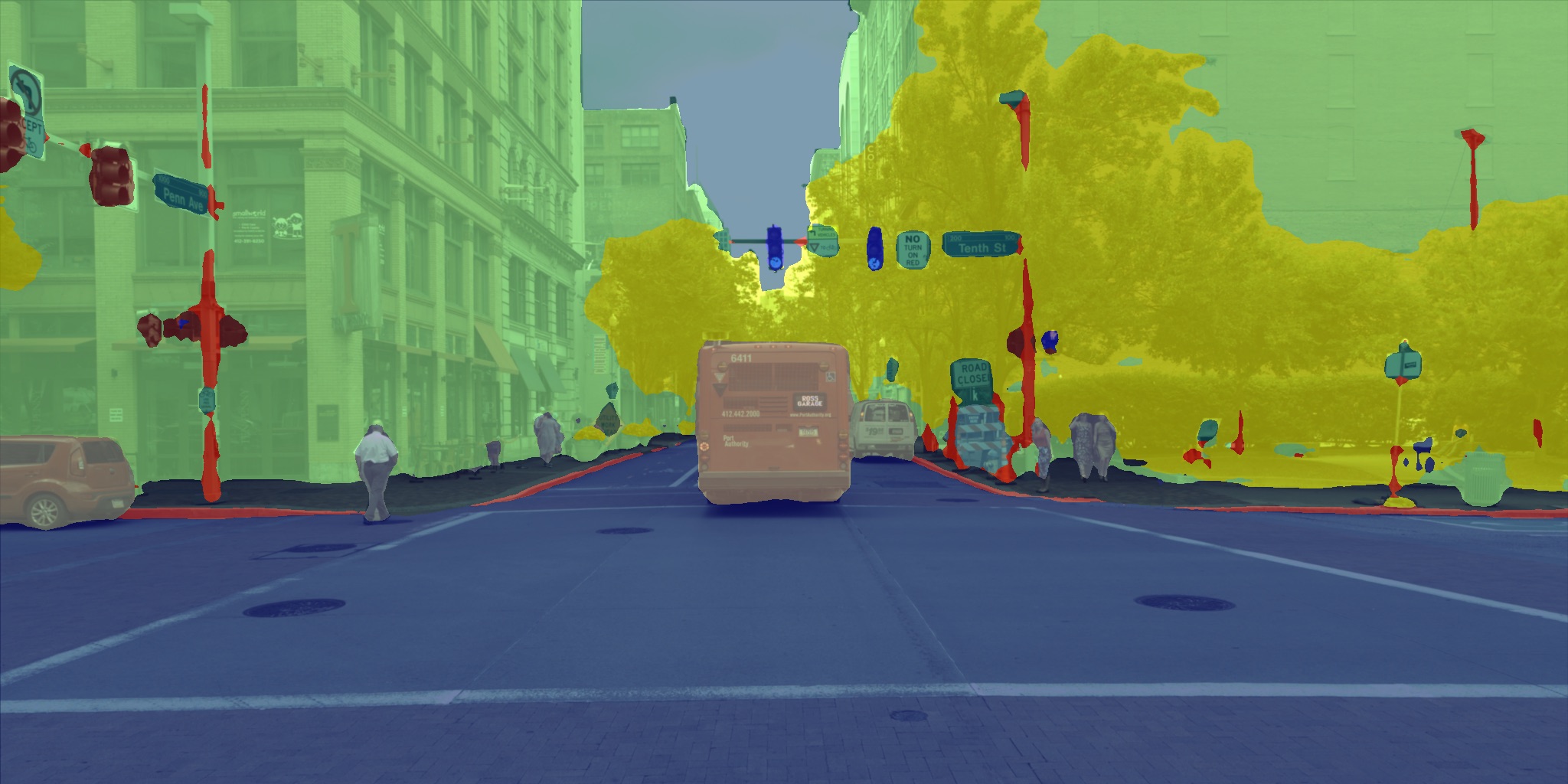}
	\hfill
	\includegraphics[width=0.304\linewidth]{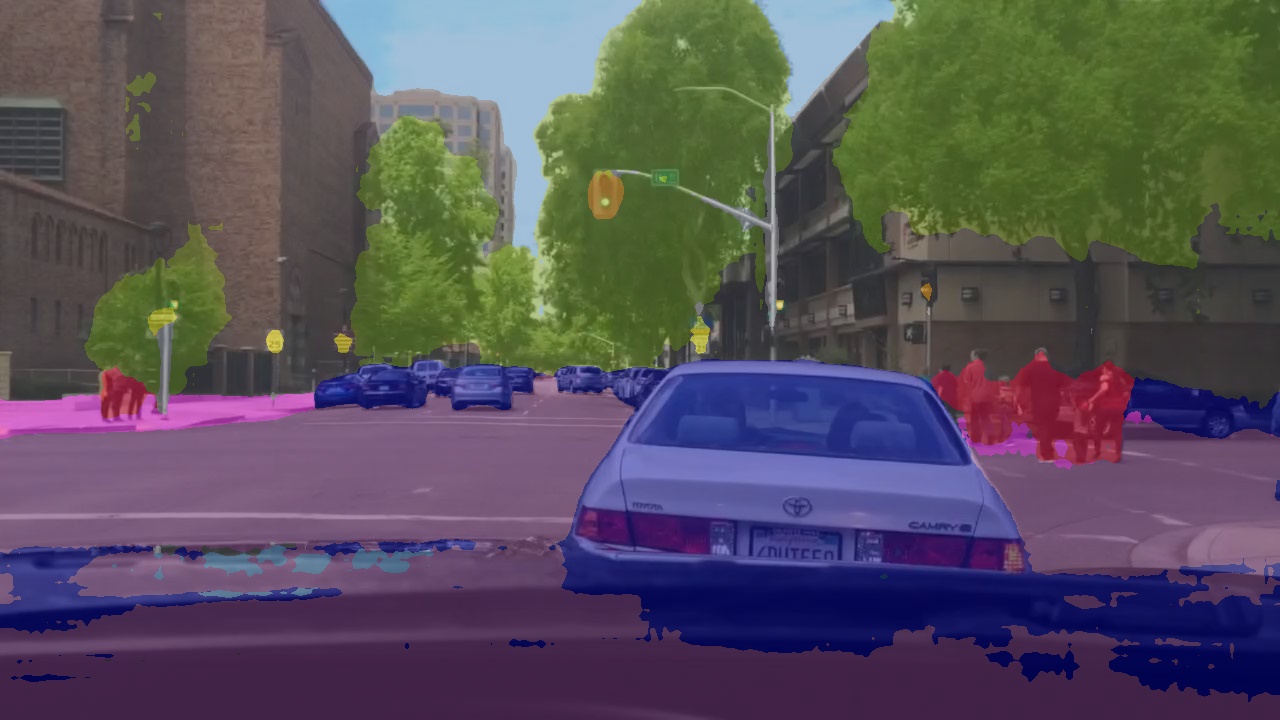}
	\hfill
	\includegraphics[width=0.342\linewidth]{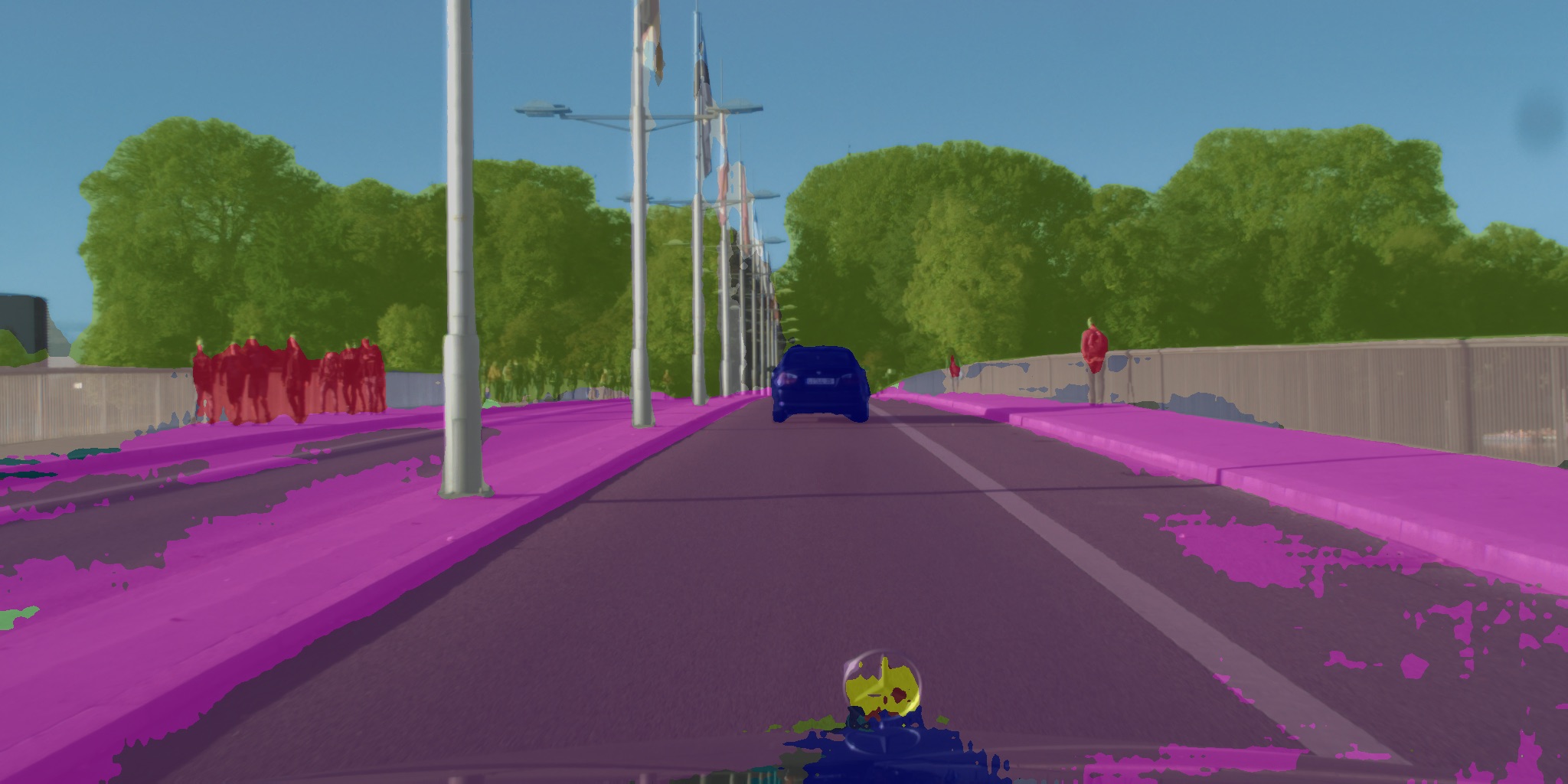}
	\includegraphics[width=0.342\linewidth]{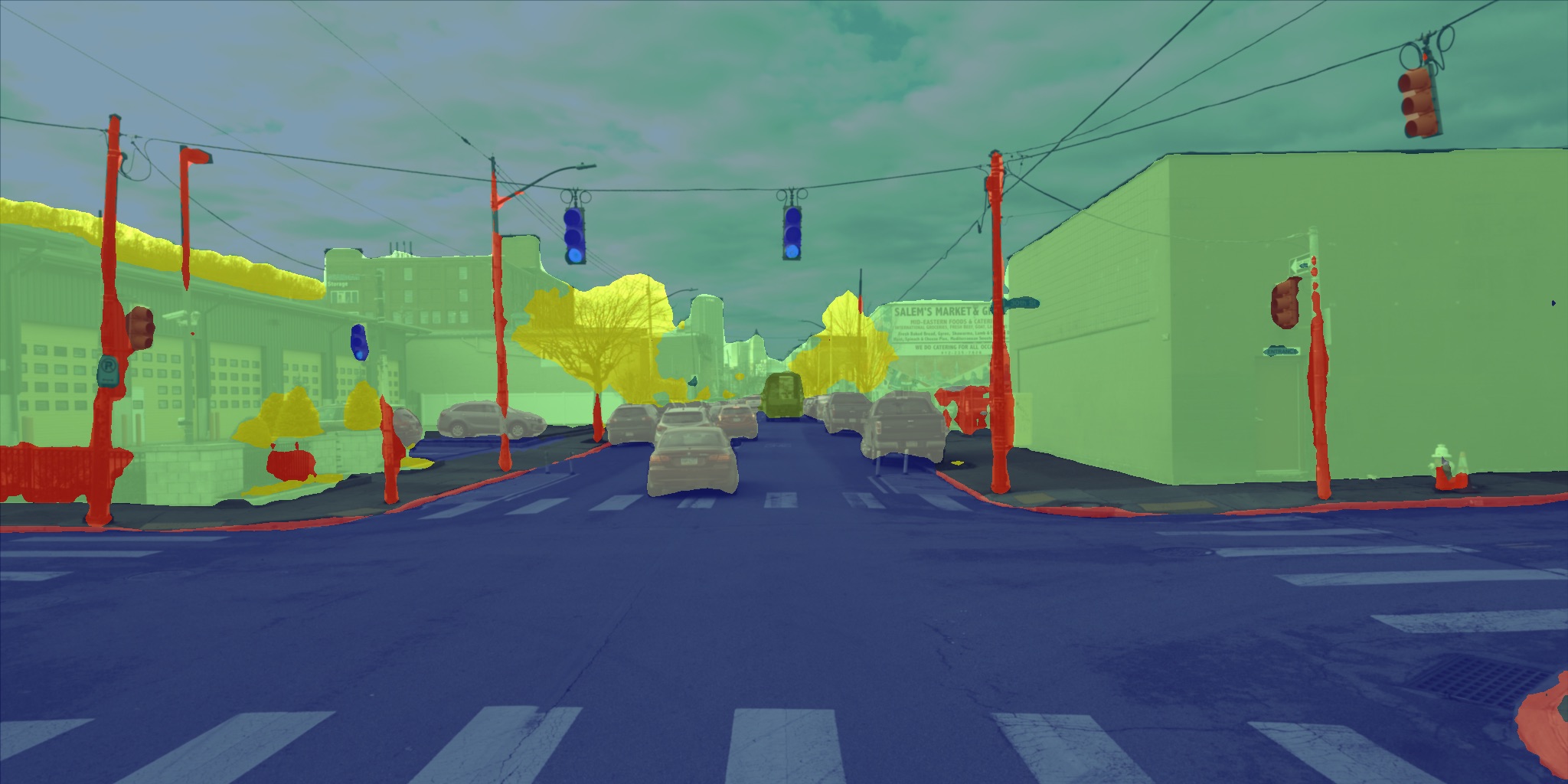}
	\hfill
	\includegraphics[width=0.304\linewidth]{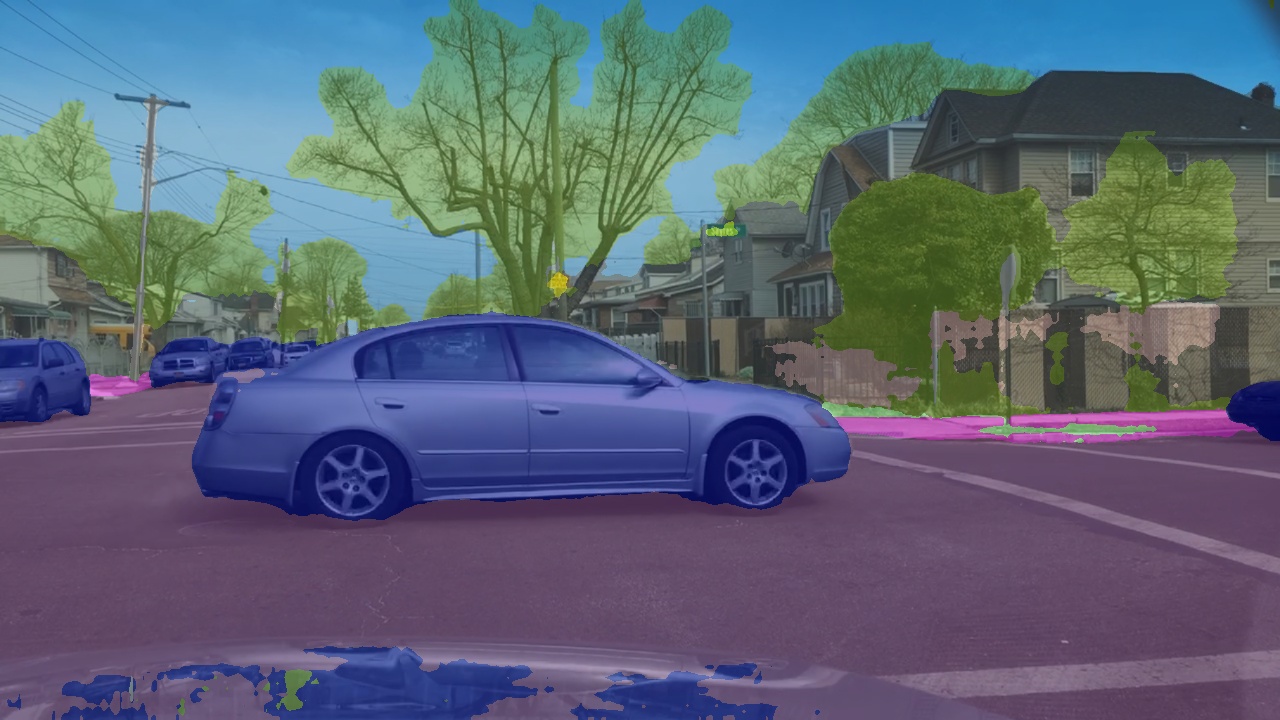}
	\hfill
	\includegraphics[width=0.342\linewidth]{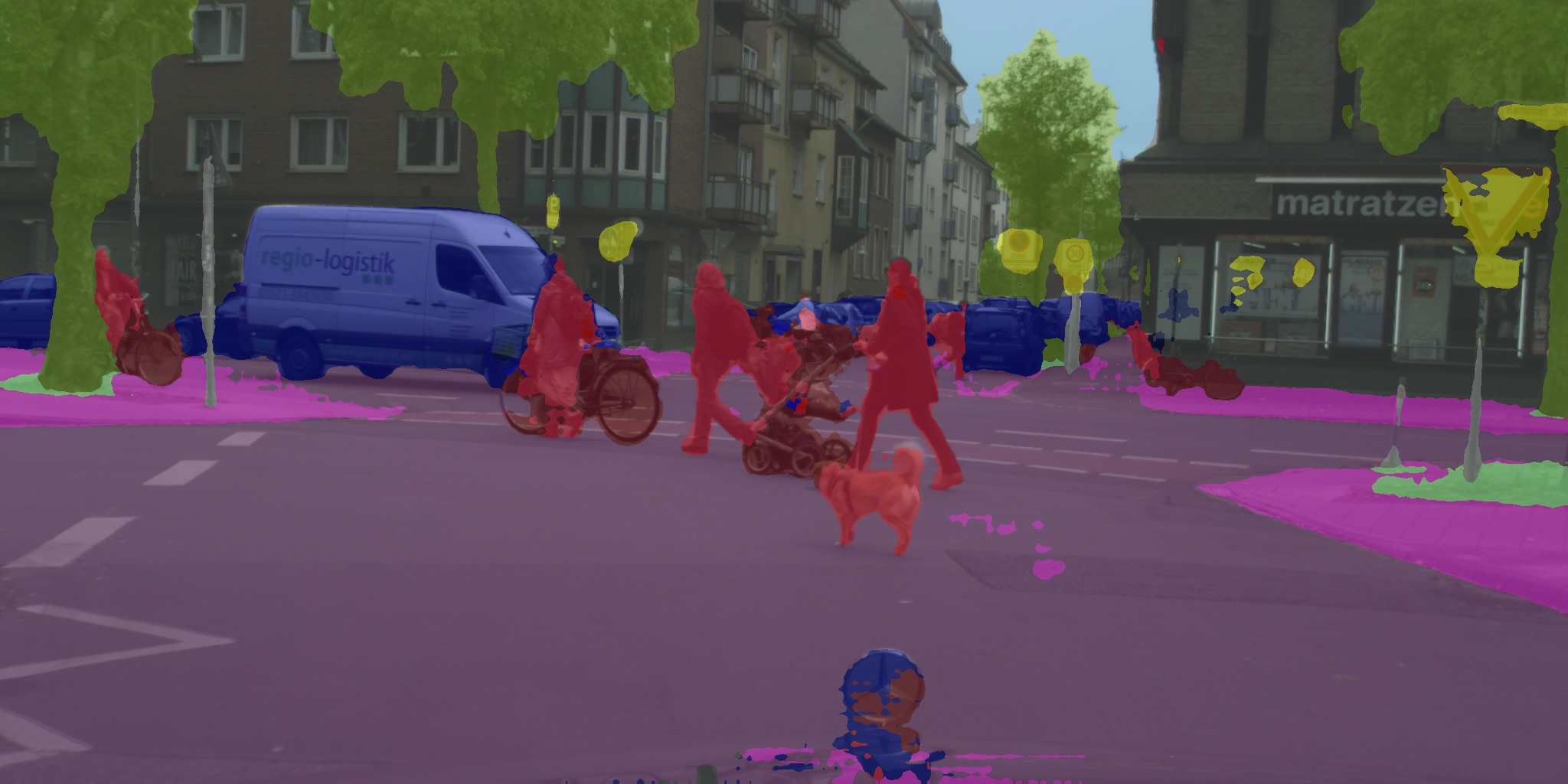}
	\caption{Semantic visualization results on \ourdataset{} (left), BDD100K (middle), as well as Cityscapes (right).
		For \ourdataset{} and BDD100K, models are trained on the corresponding dataset. For Cityscapes, we use a \ourdataset{} pretrained model. We use the standard header (1 convolutional layer) for readout.
	}
	\label{fig:vis_results_2}
	\vspace{-0.1in}
\end{figure*}

We provide additional visualization results of UrbanCity, BDD100K as well as Cityscapes. Instance
segmentation and object detection results are shown in Figure~\ref{fig:vis_results_1}, and semantic
segmentation results are shown in Figure~\ref{fig:vis_results_2}, respectively. For \ourdataset{}
and BDD100K, models are trained on the corresponding datasets. For Cityscapes, the model is trained
on \ourdataset{}. We use a heavier header to perform instance segmentation/object detection
readout tasks and a standard header for semantic segmentation readout task.

We also include a demo video as part of our supplementary material.
The file ``flowe-demo.mp4'' shows semantic segmentation results on the Cityscapes dataset.
In this video, we use \ourmethod{} trained on \ourdataset{}, and train a linear readout (1 $\times$ 1 convolution) layer
on Cityscapes training set to produce classification logits.
We can see that the model can produce impressive results while only have one linear layer learned from the labeled data, indicating that
our method can exploit unlabeled driving videos well and learn semantically meaningful representations from them.

